\PassOptionsToPackage{table}{xcolor}
\documentclass{article} 

\usepackage{iclr2025_conference,times}

\usepackage{amsmath,amsfonts,bm}

\def\eqref#1{equation~\ref{#1}}

\def\1{\bm{1}}

\DeclareMathAlphabet{\mathsfit}{\encodingdefault}{\sfdefault}{m}{sl}
\SetMathAlphabet{\mathsfit}{bold}{\encodingdefault}{\sfdefault}{bx}{n}

\usepackage{sidecap}
\usepackage{tabularx}
\usepackage{hyperref}
\usepackage{url}
\usepackage{longfigure}
\usepackage{longtable}
\usepackage{geometry}
\usepackage{subcaption}
\usepackage{array}
\usepackage{framed}
\usepackage{makecell}
\usepackage{listings}
\usepackage{adjustbox}
\usepackage{wrapfig}
\usepackage{graphicx}
\usepackage{booktabs}
\usepackage{adjustbox}
\usepackage{multirow}
\usepackage{float} 
\usepackage{amsmath}
\usepackage{caption}

\definecolor{lightgray}{rgb}{.9,.9,.9}
\definecolor{darkgray}{rgb}{.4,.4,.4}
\definecolor{purple}{rgb}{0.65, 0.12, 0.82}
\lstdefinelanguage{JavaScript}{
  keywords={break, case, catch, continue, debugger, default, delete, do, else, false, finally, for, function, if, in,cost, instanceof, new, null, return, switch, this, throw, true, try, typeof, var, void, while, with},
  morecomment=[l]{//},
  morecomment=[s]{/*}{*/},
  morestring=[b]',
  morestring=[b]",
  ndkeywords={class, export, boolean, throw, implements, import, this,cost},
  keywordstyle=\color{blue}\bfseries,
  ndkeywordstyle=\color{darkgray}\bfseries,
  identifierstyle=\color{black},
  commentstyle=\color{purple}\ttfamily,
  stringstyle=\color{red}\ttfamily,
  sensitive=true
}
\title{EEE-Bench: A Comprehensive Multimodal Electrical And Electronics Engineering Benchmark}

\author{Ming Li\thanks{Equal contribution.}\ \ $^1$, Jike Zhong\footnotemark[1]\ \ $^2$, Tianle Chen$^3$, Yuxiang Lai$^4$, Konstantinos Psounis$^2$ \\
$^1$The University of Tokyo, $^2$University of Southern California, $^3$Boston University, $^4$Emory University\\
\small \hyperlink{li-ming948@g.ecc.u-tokyo.ac.jp
}{li-ming948@g.ecc.u-tokyo.ac.jp
}
\small \hyperlink{jike.zhong@usc.edu
}{jike.zhong@usc.edu
}
\small \hyperlink{tianle@bu.edu
}{tianle@bu.edu
}
\small \hyperlink{ylai76@emory.edu
}{ylai76@emory.edu
}
\small \hyperlink{kpsounis@usc.edu
}{kpsounis@usc.edu
}
}

\iclrfinalcopy 
\begin{document}

\maketitle

\begin{abstract}
Recent studies on large language models (LLMs) and large multimodal models (LMMs) have demonstrated promising skills in various domains including science and mathematics. However, their capability in more challenging and real-world related scenarios like engineering has not been systematically studied. To bridge this gap, we propose EEE-Bench, a multimodal benchmark aimed at assessing LMMs' capabilities in solving practical engineering tasks, using electrical and electronics engineering (EEE) as the testbed. Our benchmark consists of \textbf{2860} hand-picked and carefully curated multiple-choice and free-form problems spanning \textbf{10} essential subdomains such as analog circuits, control systems, etc. Compared to benchmarks in other domains, engineering problems are intrinsically 1) more visually complex and versatile and 2) less deterministic in solutions. Successful solutions to these problems often demand more-than-usual rigorous integration of visual and textual information as models need to understand intricate images like abstract circuits and system diagrams while taking professional instructions, making them excellent candidates for LMM evaluations. Alongside EEE-Bench, we provide extensive quantitative evaluations and fine-grained analysis of \textbf{17} widely-used open and closed-sourced LLMs and LMMs. Our results demonstrate notable deficiencies of current foundation models in EEE, with an average performance ranging from 19.48\% to 46.78\%. Finally, we reveal and explore a critical shortcoming in LMMs which we term ``laziness": the tendency to take shortcuts by relying on the text while overlooking the visual context when reasoning for technical image problems. In summary, we believe EEE-Bench not only reveals some noteworthy limitations of LMMs but also provides a valuable resource for advancing research on their application in practical engineering tasks, driving future improvements in their capability to handle complex, real-world scenarios.

\end{abstract}

\section{Introduction}
\label{intro}
Electrical and electronics engineering (EEE) embodies a profound synthesis of theoretical knowledge and practical application, serving as the backbone of modern digital and power technological advancements \citep{bose2000energy,bose2020power,irwin2020basic}. From designing intricate circuitry to advancing power systems, this discipline demands rigorous logical reasoning and a deep understanding of physical laws \citep{brophy2014principles,saadat1999power}. Addressing problems in EEE requires that researchers and engineers not only grasp the related theoretical knowledge but also comprehend the associated visual contexts, such as circuit diagrams and signal waveforms. Leveraging machine learning models equipped with robust knowledge and reasoning capabilities, alongside accurate perception of visual contexts, can significantly enhance various aspects of our lives.  These models can aid in intricate hardware design processes \citep{chang2023chipgpt,liu2023chipnemo,wu2024chateda}, optimize operations within the electric energy sector \citep{majumder2024exploring}, drive innovation by proposing novel approaches to traditional scientific challenges \citep{taylor2022galactica}, and assist in solving complex educational problems.

Recent Large Language Models
(LLMs) \citep{brown2020language,jiang2024mixtral,achiam2023gpt,chiang2023vicuna,touvron2023llama} and Large Multimodal
Models (LMMs) \citep{team2023gemini, liu2024visual, zhang2023llama,dai2023instructblipgeneralpurposevisionlanguagemodels} have achieved much progress and demonstrated remarkable
capabilities in various domains, e.g., mathematical reasoning \citep{yue2023mammoth,yu2023metamath} and general science \citep{lu2022learn}. To evaluate the abilities of these foundation models, various specialized or versatile benchmarks \citep{lu2023mathvista,zhang2024mathverse,yue2024mmmu} have been developed. However, the reasoning ability of these foundation models on engineering problems with visual contexts has not been systematically examined. Therefore, to better harness the potential of foundation models, it is crucial to develop a new benchmark specific to engineering problems, and we select EEE as our case study for reasons mentioned before. This will facilitate the evaluation of LLMs and LMMs in their ability to tackle rigorous engineering reasoning tasks, thereby advancing research progress in this field.

\begin{figure*}[t!]
\centering
\includegraphics[width=\textwidth]{ 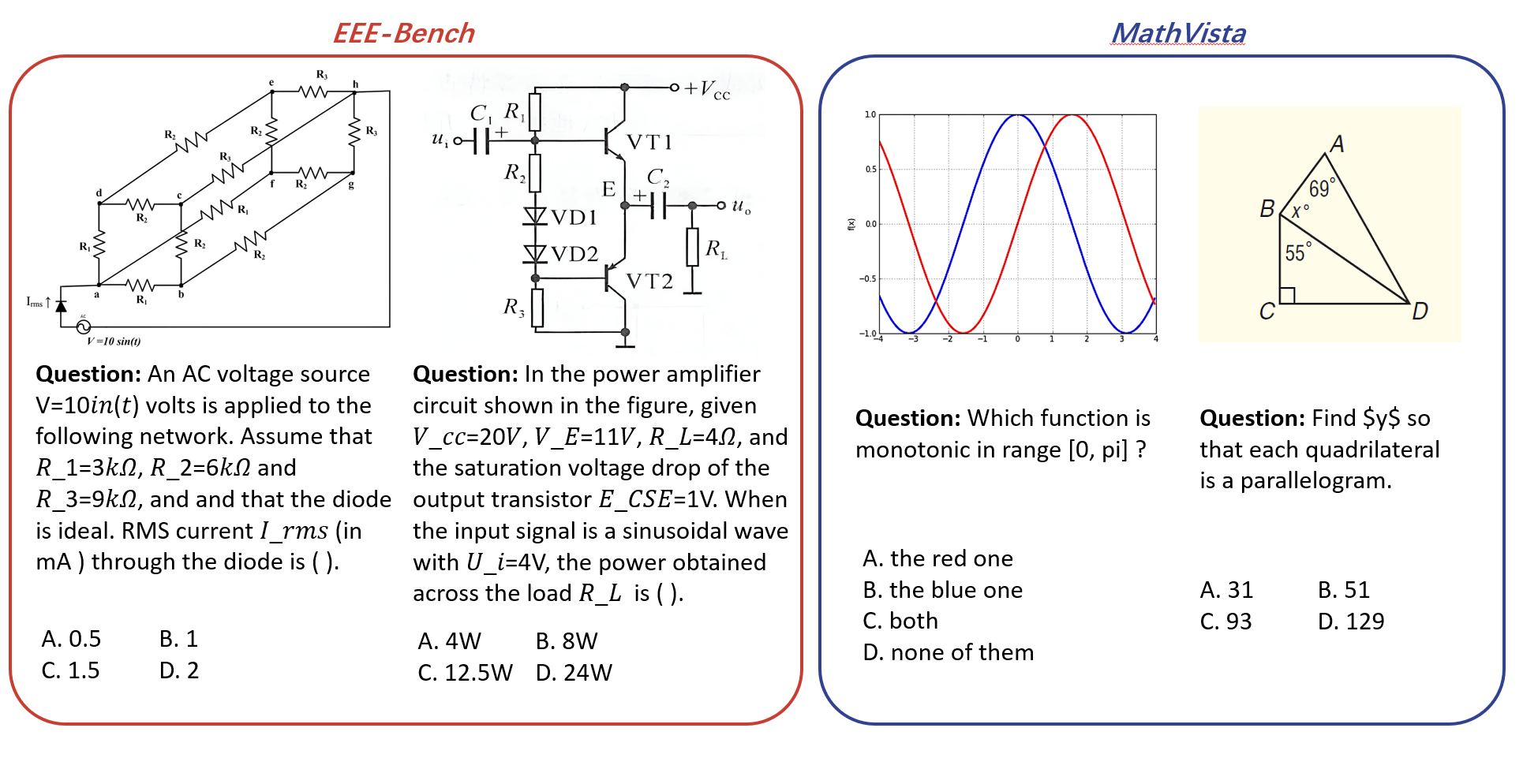}
   \caption{Comparison between EEE-Bench and MathVista problems: EEE problems feature significantly more complex and intricate visual information compared to mathematical problems, requiring specific technical knowledge to interpret diagrams effectively.}
\label{fig1}
\end{figure*}

Compared with other subjects such as mathematics, the visual contexts in EEE are often more abstract and heavily logic-oriented. Solving EEE problems demands specialized technical knowledge, as illustrated in Figure \ref{fig1}. Specifically, unlike mathematical problems \citep{lu2023mathvista, zhang2024mathversedoesmultimodalllm, wang2024measuringmultimodalmathematicalreasoning} which might involve clear numerical data and operations, visual representations in EEE, such as circuit diagrams, are intricate and logical, often containing minimal textual information (more example images can be found in \autoref{appendix: more about eee-bench}). Hence effective problem-solving in this domain requires a deep understanding of the logical relationships between different components of the diagram and the application of specific technical knowledge to interpret these visuals.

In view of the above, the proficiency of existing LMMs in mathematical domains does not guarantee their effectiveness in EEE tasks.  Owing to the same cause, existing benchmarks in other domains \citep{lu2022learnexplainmultimodalreasoning, lu2023mathvista, zhang2024mathversedoesmultimodalllm, wang2024measuringmultimodalmathematicalreasoning, yue2024mmmu, yun2024web2codelargescalewebpagetocodedataset, doris2024designqamultimodalbenchmarkevaluating} also lack some necessary rigor to accurately faithfully inspect LMMs' problem-solving capability in challenging real-world scenarios where the visual context is crucial but involves serious complexity. Consequently, there is a pressing need for a specialized benchmark focused on EEE to thoroughly assess the multimodal reasoning abilities of LMMs within this specific field as well as to provide a broader outlook of their problem-solving capabilities in these practical engineering problems in the real world.

To this end, we present EEE-Bench, a pioneering multimodal electrical and electronics engineering (EEE) reasoning benchmark in visual
contexts. To ensure that EEE-Bench encompasses a thorough range of knowledge in EEE, we initially identified ten pivotal topics for problem collection: (1) \textit{Digital Logic Circuits and Microprocessor Design}, (2) \textit{Circuit Theory and Network Analysis}, (3) \textit{Analog Circuits}, (4) \textit{Power Electronics and Power Systems}, (5) \textit{Signals and Systems}, (6) \textit{Communication Systems}, (7) \textit{Control Systems}, (8) \textit{Electronic Devices}, (9) \textit{Electrical Machines}, and (10) \textit{Electromagnetics}. These topics comprehensively cover both theoretical foundations and practical applications within EEE. When collecting problems for EEE-Bench, we include a broad range of visual contexts such as electric and digital circuits, system diagrams, abstract scenes, electronic component images, and assorted tables, charts, and plots. It's important to note that many figures within EEE-Bench feature combinations of these visual contexts, enhancing their complexity and educational value. Overall, EEE-Bench comprises 2,860 entirely new problems. 

\begin{figure*}[t!]
\centering
\includegraphics[width=\textwidth]{  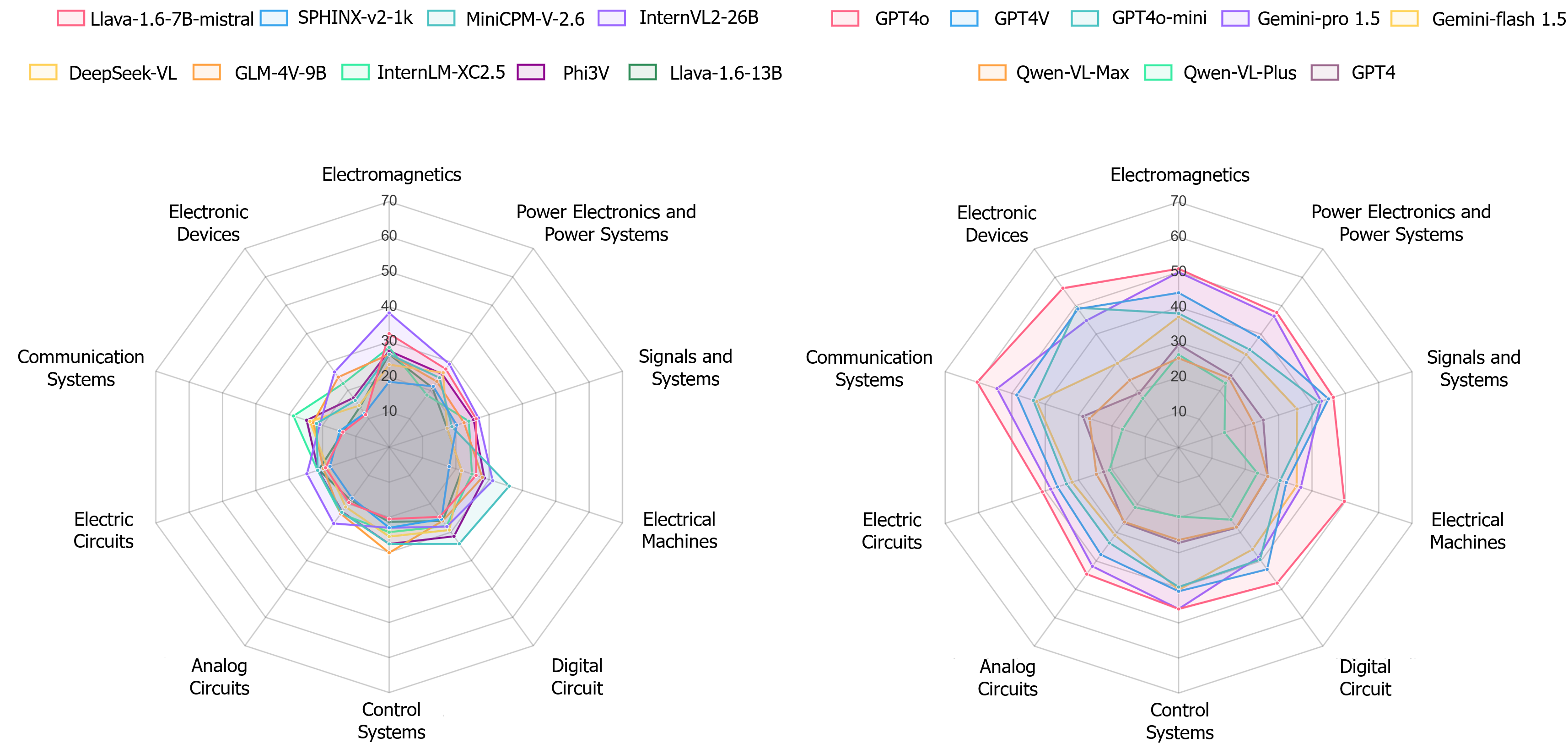}
   \caption{Overview of performance of \textbf{left: open source} models and \textbf{right: close source models}.}
\label{fig2}
\end{figure*}

We conduct extensive experiments using EEE-Bench to assess the reasoning abilities of 17 leading foundation models, including both open-source \citep{chen2024far,liusphinx,liu2024visual} and closed-source \citep{openai2024gpt4,team2023gemini}.  Figure \ref{fig2} illustrates the overall experimental results. We discover that most existing LMMs struggle to understand EEE diagrams and perform poorly on EEE-Bench.  Our results show that GPT-4o \citep{openai2024gpt4} achieves the best overall performance across different 
topics with 46.78\% accuracy. Notably, closed-source models generally outperform open-source models. When evaluating performance across various subjects, we observed that many closed-source models excel in subjects with straightforward visual contexts but require significant computational resources, such as Electronic Devices. However, these models tend to struggle with subjects that involve complex visual diagrams, such as Circuit Theory and Network Analysis. These findings underscore the proficiency of closed-source LMMs in numerical computation, but also reveal their limitations in processing intricate visual information. Through the
fine-grained error analysis, we found that current LMMs struggle to understand EEE diagrams and reason effectively. Taking GPT-4o as an example, we found that over 50\% of the errors stem from reasoning issues, and 26.5\% are due to errors in image perception.
These findings
suggest that better EEE visual understanding capabilities and more robust EEE reasoning for LMMs could be the potential
for future advancement in addressing multi-modal EEE problems. 

Lastly, we analyze which modality—vision or text—LMMs predominantly rely on. Our investigation revealed a 'laziness' phenomenon within existing LMMs.  In contrast to previous studies that concentrated on hallucinations of nonexistent objects with natural images \citep{liu2024phd,qian2024easy,liu2023mitigating}, our focus is on the robustness of the understanding and reasoning process in LMMs dealing with technical images. Specifically, we found that when presented with spurious captions that contradict the accompanying images, LMMs tend to disregard essential visual information, relying instead on the textual content and reasoning, even if it is misleading. This tendency leads to flawed reasoning processes within the models. Consequently, augmenting original question text with such spurious captions results in significant accuracy declines—7.79\% for GPT-4o and 6.78\% for Gemini-Pro 1.5 on the EEE-Bench. These results reveal that existing LMMs mainly rely on text information, and intriguingly, they may overlook visual information when it is explicitly provided in text.

The contributions of this paper can be summarized as follows:
\begin{itemize}
    \item  We introduce EEE-Bench, a novel multimodal benchmark designed for assessing the reasoning abilities of LMMs in electrical and electronics engineering (EEE) problems. EEE-Bench consists of 2860 samples spanning 10 essential subjects in EEE, featuring a diverse range of visual contexts such as electric and digital circuits, system diagrams, and others. This benchmark is specifically crafted to evaluate how well LMMs can handle complex visual and logical challenges within the EEE field.
    
    \item We conduct extensive experiments using EEE-Bench to assess the reasoning abilities of 17 leading open-source and closed-source foundation models. We found that most existing LMMs struggle to deal with EEE problems, especially in subjects with complex visual contexts, yielding an average performance ranging from 19.48\% to 46.78\%. We also demonstrate that reasoning errors and image perception errors are the main problems for current LMMs. These findings provide insights for future improvement.

    \item We further explore which modality—text or vision—current LMMs primarily rely on. Our research identifies a 'laziness' phenomenon in these models: LMMs will not try to read images if relevant visual information for reasoning and solving problems appears in the text, even if this extra-textual information is spurious. Our experiments show that introducing additional misleading captions into the text leads to a 7.79\% drop in accuracy for GPT-4o on the EEE-Bench.
\end{itemize}

\section{The EEE Benchmark}
In this section, we introduce \textbf{EEE-Bench}. First, we introduce our data collection process in \autoref{data collection}. Next, we provide a summary of EEE-Bench in \autoref{benchmark summary}. Lastly, we provide a high-level analysis in \autoref{bench_ana}; we leave detailed analysis to \autoref{exp:results} and \autoref{error analysis}.

\subsection{Data Collection}
\label{data collection}
\paragraph{Guidelines.} As our general principle is to establish a challenging real-world-oriented benchmark capable of assessing both visual understanding and logical reasoning with integrated multimodal knowledge, we adhere to the following guidelines for initial data collection: 1) questions with high-resolution and context-rich images (diverse complex patterns, domain-specific markings, etc); 2) problems that require logical reasoning steps (calculations, cross-domain information integration) to solve in addition to visual comprehension; 3) problems with deterministic answers for robust evaluation (multiple choices questions and free-form questions requiring answer no longer than one word as did in \cite{lu2023mathvista}); 4) balanced collection source, sub-domain representation, and problem type.
\paragraph{Collection and curation.}
We collect all our questions from official EEE exams as well as verified online sources in a multimodal setting (see \autoref{benchmark summary} for summary details). Note that as our work is first in this direction, no existing dataset or benchmark is available, hence all questions are newly collected. Example problems can be found in \autoref{fig1},  \autoref{fid_diff}, and \autoref{appendix: more failure example}. For data curation, we employ a two-stage filtering process. To ensure the solution quality, we first filter for questions appearing in at least two distinct sources having identical solutions. To ensure the significance of visual input, we subsequently adopt methods similar to \cite{yue2024mmmuprorobustmultidisciplinemultimodal} where two trials are run using text-only GPT-4 \citep{openai2024gpt4} in which questions with two correct answers are eliminated. These processes result in 2860 high-quality examples across ten sub-domains. Two final quality checks for typos, image resolution, and prompt completeness were also conducted by graduate students. 

\paragraph{Release.} We release the benchmark as organized in two formats: 1) one complete set and 2) sub-domains. We also release metadata annotations denoting question title (stored in markdown format), image path, answer, question type, image ID, source, source image ID, topic, query, answer type, and choices in JSON format similar to \cite{lu2023mathvista}. For the question title, we also prepend an instructional prompt, see \autoref{app:prompt} for details. We also release the complete evaluation source code.
\begin{figure*}[t]
\centering
\begin{minipage}[c]{0.45\textwidth}
\small
\centering
  \captionof{table}{\textbf{Key Statistics of EEE-Bench.}}
  \label{t1}
  \centering
  \begin{adjustbox}{width=\linewidth}
   \begin{tabular}{lr}
 \toprule
 \textbf{Statistic} & \textbf{Number} \\
 \midrule
  Total questions & 2,860 \\
  ~- Multiple-choice questions & 2,059 (72.0\%) \\
  ~- Free-form questions & 801 (28.0\%) \\
  ~- \textbf{Newly collected questions} & \bf2,860(100\%)\\ 
  \midrule
 Number of unique images & 2813 \\
 Number of unique questions & 2826 \\
 Number of unique answers & 329  \\
 \midrule
 Maximum question length & 466 \\
 Minimum question length & 7\\
 Maximum answer length & 8 \\
 Average question length & 57.4\\
 Average answer length & 1.4 \\
 \bottomrule
 \end{tabular}
 \end{adjustbox}
\end{minipage}
\qquad
\begin{minipage}[c]{0.50\textwidth}
\centering
\vspace{-0.2cm}
\caption{\textbf{Subject Distribution of EEE-Bench.} DLC\&MPD: Digital Logic Circuits and Microprocessor Design, PS: power systems. NA: network analysis}
\label{fig3.5}
\vspace{0.15cm}
\includegraphics[width=0.99\linewidth]{  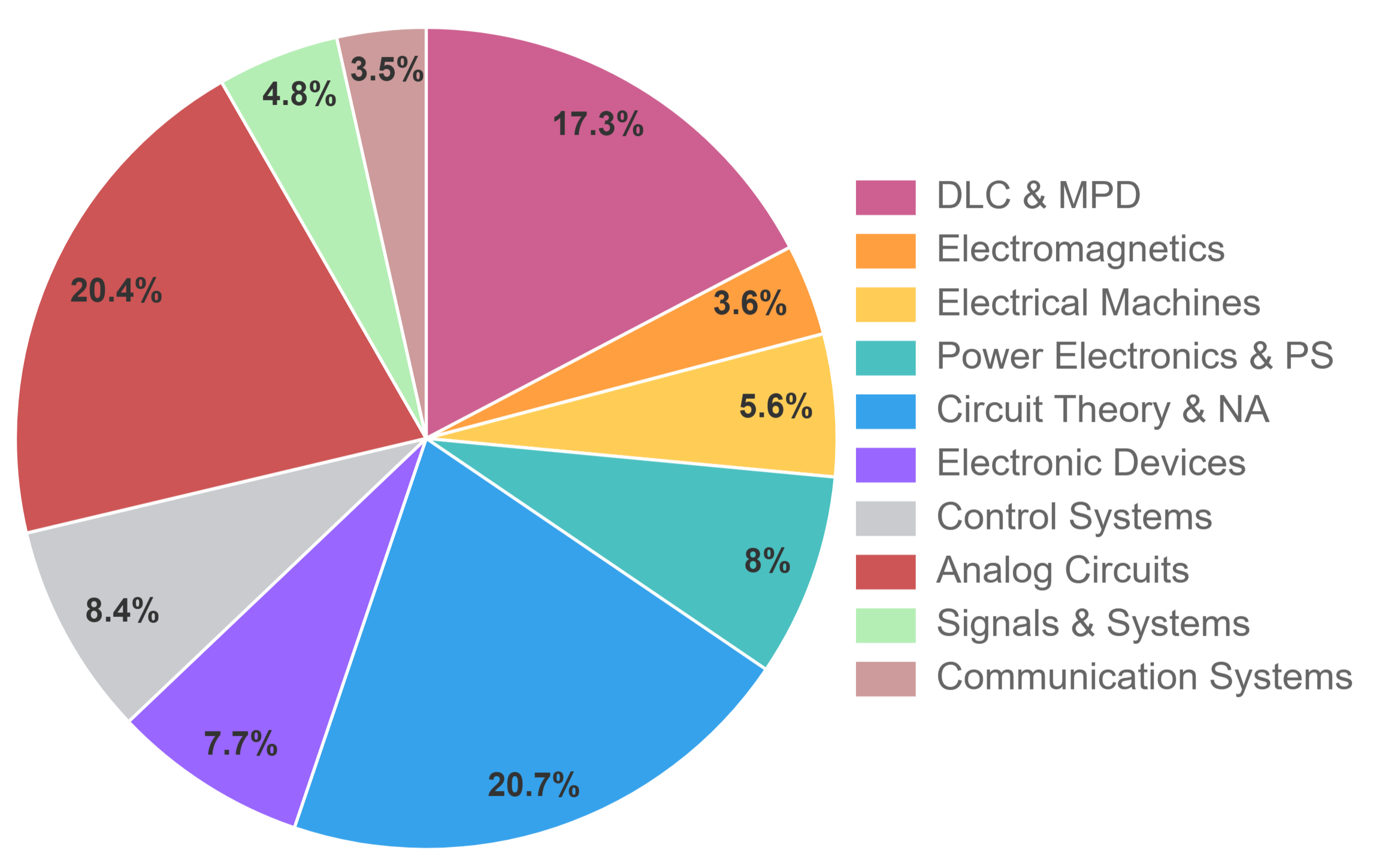}
\end{minipage}
\end{figure*}

\subsection{Benchmark Summary}
\label{benchmark summary}
The key statistics of EEE-Bench are shown in \autoref{t1} and subject distribution is shown in \autoref{fig3.5}. Our benchmark consists of a total of 2860 examples divided into 10 diverse but essential sub-domains. We first summarize general statistics. Questions in our benchmark are categorized into two types similar to \cite{lu2023mathvista}: multiple choice and free-form, counting 2059 and 801, respectively. Regarding answer type, four forms are present: text (single word) represents the largest group of 2059 questions, 595 are float, 204 are integer, and 2 are list. The average token length for question text is 306. Next, we provide a breakdown of the 10 sub-domains. \textit{Digital Logic Circuits and Microprocessor Design} (496 examples) assess the design and functionality of circuits central to computing systems like processors and memory. \textit{Electronic Devices} (219 examples) evaluate the understanding of components such as diodes and transistors, foundational to all electronics. \textit{Communication Systems} (101 examples) assess signal transmission and processing, crucial for wireless networks and satellite communications. \textit{Signals and Systems} (138 examples) evaluate the mathematical representation and manipulation of signals, fundamental for control systems and digital signal processing. \textit{Electrical Machines} (161 examples) assess knowledge of devices like motors and generators, essential for power generation and industrial applications. \textit{Electromagnetics} (102 examples) evaluate the principles of electromagnetic fields and waves, vital for RF communication and antenna design. \textit{Control Systems} (239 examples) assess feedback mechanisms and system stability, critical for automation and robotics. \textit{Power Electronics and Power Systems} (229 examples) evaluate the conversion and control of electrical power, key to renewable energy systems and power grids. \textit{ Circuit Theory and Network Analysis} (592 examples) assess both AC and DC circuit analysis, forming the foundation of all electrical applications. Lastly, \textit{Analog Circuits} (583 examples) evaluate continuous-time circuit design, essential for applications like audio, RF, and sensor interfaces. We leave more detailed descriptions of these 10 subdomains to \autoref{appendix:detail_intro}. 


\subsection{Benchmark Analysis}
\label{bench_ana}
\paragraph{Overall quality.} Benchmarks catered toward LMM evaluations should provide unbiased, fair, and exhaustive evaluations. EEE-Bench achieves these qualities by ensuring diversity in terms of both type and sub-domain. The dynamic question settings reduce the potential for bias, and the wide coverage of sub-domains captures the diverse nature of engineering problems while ensuring a comprehensive and fair representation of the EEE domain. More importantly, the diverse nature of EEE-bench also enables granular evaluation of model performance, thus providing a framework for more nuanced by-type and by-sub-domain analysis (\autoref{exp:results}), readily unveiling the underlying strengths and weaknesses of different models. Varied difficulty levels are also introduced to best mimic real-world situations and enhance the interpretability of the assessment. We note that the best-performing proprietary model GPT-4o \citep{openai2024gpt4o} and open-source model InternVL2-26B \citep{chen2024internvl} set at 46.78\% and 26.89\% overall accuracy respectively, while the averages of all proprietary models and open-sourced models are 35.28\% and 23.19\% respectively. For more detailed results and analysis, please refer to \autoref{exp:results}.
  \begin{figure*}[t]
\label{intro:perform}
    \centering
    \begin{minipage}{0.95\linewidth}
            \centering
            \vskip-5pt
            \includegraphics[width=\linewidth]{  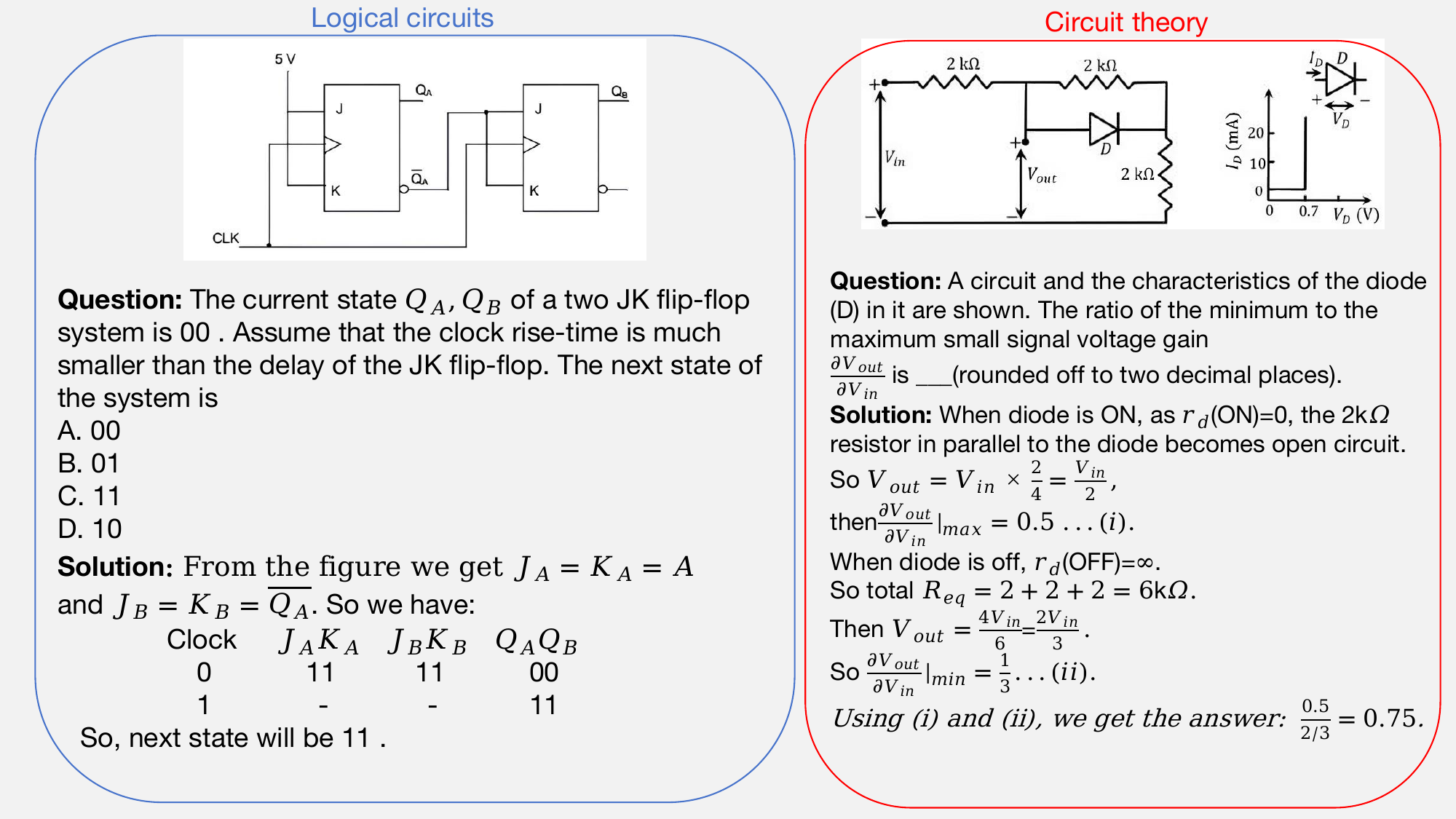}
    \end{minipage}
    \caption{Comparison between solutions of logical circuits and circuit theory. Logic Circuits demand a thorough comprehension of logical structuring and timing analysis, yet they do not heavily rely on computation. In contrast, problems of circuit theory involve solving complex equations and applying numerical methods to analyze circuits.}
    \label{fid_diff}
    \vskip-15pt
\end{figure*}
\paragraph{Comprehensive evaluation of LMM's various abilities.} EEE-Bench encompasses 10 pivotal subjects within the field of electrical and electronics engineering, extending from circuit analysis to control systems. Each subject demands specialized capabilities for addressing unique challenges. For instance, as shown in \autoref{fid_diff}, tackling problems within the Digital Logic Circuits and Microprocessor Design discipline requires models to deeply understand intricate logic relationships in digital circuits and reason effectively. These problems demand a thorough comprehension of logical structuring and timing analysis, yet they do not heavily rely on computational power or advanced mathematical computations. In contrast, many challenges from Circuit Theory and Network Analysis may present less complexity regarding logic understanding but require extensive computational processing and the use of sophisticated mathematical tools. This involves solving complex equations and applying numerical methods to analyze and interpret circuit behaviors and responses under various conditions. A detailed introduction about the characteristics of each subject's problems and the required ability for LMMs to solve them are discussed in \autoref{appendix:lmm_abilities}. The juxtaposition of these disciplines within EEE-Bench highlights the diverse skill sets needed to address the spectrum of issues prevalent in the field of electrical and electronics engineering, therefore effectively examining various abilities of existing LMMs.
\paragraph{Significance of vision in EEE-Bench.}
Meaningful multimodal benchmarks should place largely equal weights on both visual and language inputs. Recent studies have unveiled that some previous multimodal benchmarks lack visual depth and complexity thus making their visual input ad-hoc \citep{tong2024cambrian1, yue2024mmmuprorobustmultidisciplinemultimodal}. In EEE-Bench, we ensure that vision plays an indispensable role. \textbf{First}, as shown in \autoref{main_result}, the text-only baseline with GPT-4 results in a 21.26\% and 14.69\% drop in accuracy compared to GPT-4o and GPT-4V respectively with complete input. \textbf{Second}, we found that the improved text-only-with-caption baseline merely improves accuracy by 2.7\% with GPT-4 (\autoref{main_result}). These results highlight the significance and indispensability of vision in EEE-Bench.


\section{Experiments}
In this section, we conduct a comprehensive evaluation of 17 existing LLMs and LMMs on EEE-Bench. \autoref{exp:set} introduces the experimental setup. Subsequently, \autoref{exp:results} presents the quantitative results. Finally, \autoref{exp: ana} provides a detailed, fine-grained error analysis.

\begin{table}[t]
    \centering
    \adjustbox{max width=\textwidth}{
    \begin{tabular}{lccccccccccc}
        \toprule
        Model & ALL & EMag & PEPS & SS & EMac & DLC & CorS & AC & CTNA & ComS & ED \\
        \midrule
        Random Chance &17.45&28.43&19.65&15.22&18.01&18.55&15.90&18.18&16.89&15.84&10.50\\
        \midrule
        \multicolumn{12}{c}{LLMs (text input)}\\
        \midrule
        GPT-4&25.52 & 29.41 & 25.33 & 25.36 & 26.71 & 28.23 & 27.20 & 26.59 & 22.47 & 28.71 & 19.18 \\

        GPT-4 w/ caption &28.22 & 28.43 & 28.82 & 26.81 & 31.06 & 31.25 & 25.10 & 25.21 & 27.20 & 34.65 & 30.59 \\
        \midrule
        \multicolumn{12}{c}{Open-source LMMs} \\
        \midrule
        Llava-1.6-7B & 21.36 & 32.35 & 27.51 & 26.09 & 26.09 & 24.60 & 20.50 & 19.55 & 19.09 & 13.86 & 11.42 \\
        Llava-1.6-13B & 20.73 & 26.47 & 20.96 & 17.39 & 21.74 & 26.01 & 21.34 & 18.70 & 21.11 & 13.86 & 14.16 \\
        SPHINX-v2-1k & 19.48 & 18.63 & 21.40 & 20.29 & 18.01 & 25.40 & 23.01 & 18.01 & 17.74 & 14.85 & 11.87 \\
        MiniCPM-V-2.6 & 25.21 & 26.47 & 24.45 & 18.84 & \cellcolor{blue!25}36.02 & \cellcolor{blue!25}34.07 & \cellcolor{blue!25}27.62 & 22.98 & 21.45 & 21.78 & 16.44 \\
        InternVL2-26B & \cellcolor{blue!25}26.89 & \cellcolor{blue!25}38.24 & \cellcolor{blue!25}29.26 & \cellcolor{blue!25}26.81 & 31.06 & 28.02 & 23.01 & \cellcolor{blue!25}26.93 & \cellcolor{blue!25}24.66 & 20.79 & \cellcolor{blue!25}26.48 \\
        DeepSeek-VL & 22.31 & 23.53 & 26.20 & 17.39 & 21.74 & 29.23 & 25.52 & 21.10 & 18.58 & 23.76 & 14.61 \\
        GLM-4V-9B & 24.06 & 26.47 & 23.14 & 22.46 & 27.95 & 26.01 & 30.13 & 23.50 & 19.76 & 22.77 & 24.66 \\
        InternLM-XC2.5 & 23.85 & 28.43 & 18.34 & 23.91 & 24.84 & 27.82 & 24.27 & 23.50 & 21.45 & \cellcolor{blue!25}28.71 & 22.37 \\
        Phi3V & 24.90 & 27.45 & 25.76 & 25.36 & 28.57 & 31.45 & 27.62 & 22.81 & 21.28 & 24.75 & 17.35 \\
        \midrule
        \multicolumn{12}{c}{Closed-source LMMs} \\
        \midrule
        GPT-4o & \cellcolor{red!25}46.78 & \cellcolor{red!25}50.98 & \cellcolor{red!25}47.60 & \cellcolor{red!25}46.38 & \cellcolor{red!25}49.69 & \cellcolor{red!25}47.78 & \cellcolor{red!25}46.03 & \cellcolor{red!25}44.60 & \cellcolor{red!25}40.88 & \cellcolor{red!25}60.40 & \cellcolor{red!25}56.16 \\
        GPT-4V & 40.21 & 44.12 & 38.86 & 44.93 & 32.30 & 42.94 & 41.00 & 37.74 & 36.32 & 48.51 & 48.86 \\
        GPT-4o-mini & 36.99 & 38.24 & 34.50 & 42.03 & 30.43 & 39.52 & 39.75 & 33.62 & 32.77 & 43.56 & 49.32 \\
        Gemini-Pro 1.5 & 41.99 & 50.00 & 46.29 & 42.75 & 36.65 & 38.71 & 46.03 & 41.85 & 38.34 & 54.46 & 44.75 \\
        Gemini-Flash 1.5 & 33.95 & 37.25 & 32.75 & 35.51 & 35.40 & 35.89 & 40.59 & 30.87 & 31.93 & 42.57 & 29.68 \\
        Qwen-VL-Max & 25.73 & 25.49 & 24.45 & 22.46 & 26.71 & 28.02 & 26.36 & 26.24 & 24.66 & 26.73 & 23.74 \\
        Qwen-VL-Plus & 21.33 & 26.47 & 22.71 & 13.77 & 23.60 & 25.40 & 19.67 & 21.10 & 20.78 & 16.83 & 17.35 \\
        \bottomrule
    \end{tabular}
    }
    \caption{ Comparison of model performances across various EEE subjects. Subjects: EMag: Electromagnetics, PEPS: Power Electronics and Power Systems, SS: Signals and Systems, EMac: Electrical Machines, DLC: Digital Logic Circuits and Microprocessor Design, CorS: Control Systems, AC: Analog Circuits, CTNA: Circuit Theory and Network Analysis, ComS: Communication Systems, ED: Electronic Devices. The highest accuracy
    for \colorbox{red!25}{\textbf{closed-source}} and \colorbox{blue!25}{\textbf{open-source}} LMMs is marked in \colorbox{red!25}{\textbf{red}} and \colorbox{blue!25}{\textbf{blue}} respectively.}
    \label{main_result}
    \vspace{-1em}
\end{table}
\subsection{Experiment Setup}
\label{exp:set}
\paragraph{Model selection.} We evaluate a total of 1 LLM and 16 LMMs against EEE-Bench. The abundant model selection captures the diversity of LMMs today and ensures a thorough and comparable assessment. In detail, the LMM set includes \textbf{(a) 9 open-source} LMMs: LLaVA1.6-7B, LLaVA-1.6-13B \citep{liu2023improvedllava, liu2024visual}, DeepSeek-VL-7B \citep{lu2024deepseek}, Phi3-Vision-4.2B \citep{abdin2024phi}, MiniCPM-Llama3-V 2.6 \citep{yao2024minicpm}, SPHINX-v2-1k \citep{liusphinx}, InternVL2-26B \citep{chen2024far,chen2024internvl}, GLM-4V-9B \citep{glm2024chatglm}, and InternLM-XC2.5 \citep{zhang2024internlm},  \textbf{(b) 7 closed-source} LMMs: GPT-4o \citep{openai2024gpt4}, GPT-4o-mini \citep{openai2024gpt4o}, GPT-4V \citep{gptvision2023}, Gemini 1.5 Pro, Gemini 1.5 Flash \citep{reid2024gemini, team2023gemini}, Qwen-VL-Max and Qwen-VL-Plus \citep{bai2023qwen}. We provide example responses from various models in \autoref{all_comparison_1} and \autoref{all_comparison_2}. To further ground the impact of vision in LMMs, we provide two additional input-only evaluation baselines using GPT-4 \citep{achiam2023gpt}: 1) a text-only baseline where the visual inputs are entirely hidden and 2) an improved baseline of text-only input but with image captions generated by GPT-4o. The prompt for caption generation is available in \autoref{app:prompt}. More details about all models can be found in \autoref{app:imp}. 

\paragraph{Evaluation.} We utilize an evaluation pipeline similar to MathVista \citep{lu2023mathvista}. Raw responses are first generated from LMMs and then processed by GPT-4o-mini to extract the concise answers given an exemplary system prompt. Finally, accuracy is obtained through calculation against the ground truth.



\paragraph{Implementation details.}  We evaluate all models in a zero-shot manner for better generalization examination. We run each evaluation setting twice for each LMM and record their average scores for the final report. All experiments for open-source models are conducted on NVIDIA A100 GPUs. We also provide all hyperparameters and model setups used for evaluations in \autoref{app:imp}. 

\subsection{Experiment Results}
\label{exp:results}

In this section, we compare the performance of 17 open-source and closed-source models. We report the overall accuracy and the accuracy among each subject. The results are shown in \autoref{main_result}. 

\begin{wrapfigure}{r}{0.4\textwidth}  
  \centering
  \includegraphics[width=0.42\textwidth]{   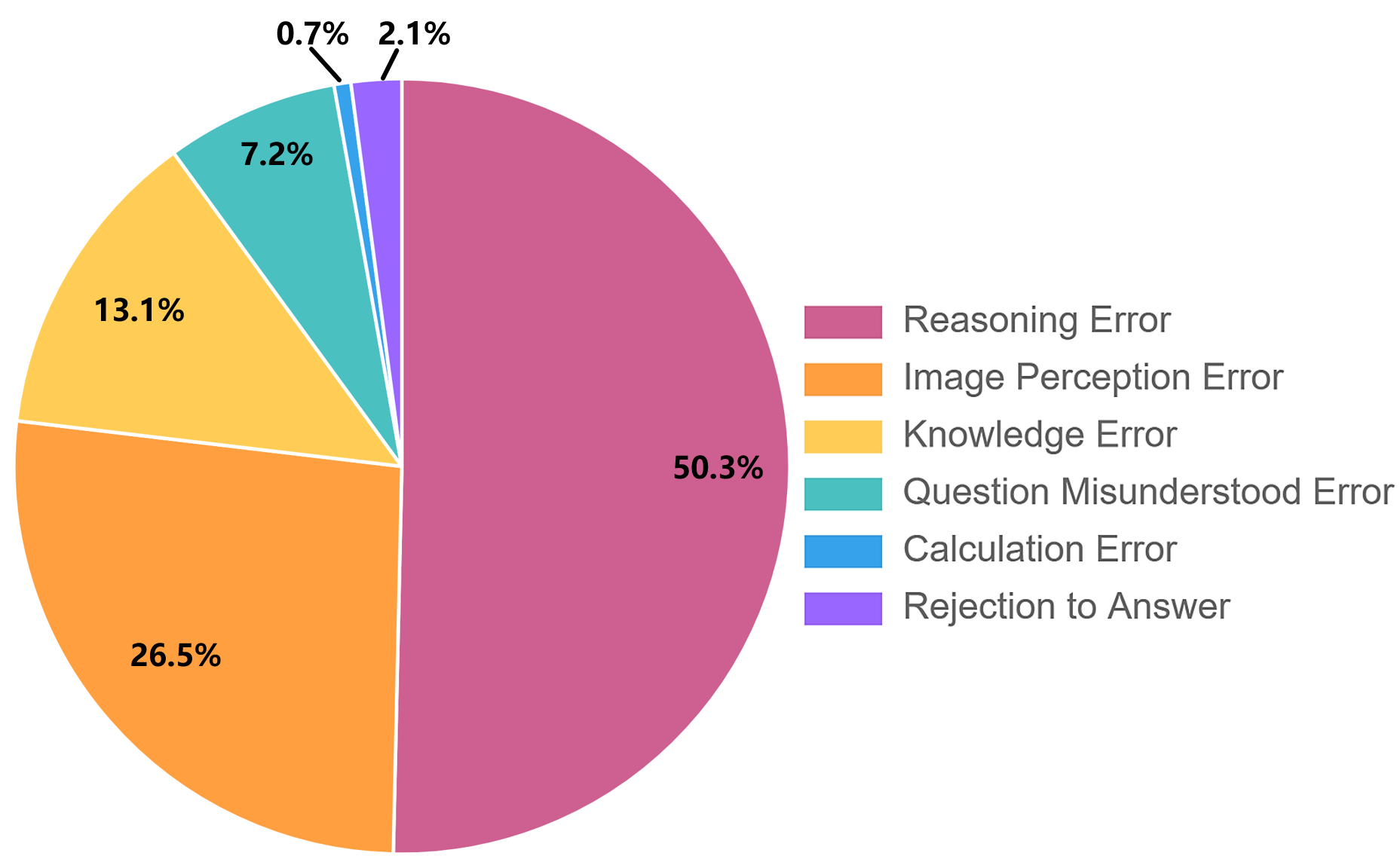}  
  \caption{ Error distribution of the wrong results from GPT-4o
on EEE-Bench.}
\vspace{-1em}
  \vskip -10px
  \label{err_fig}
\end{wrapfigure}

\paragraph{Finding 1: Existing LMMs struggle to address EEE problems.} The results presented in \autoref{main_result} highlight that existing LMMs can not deal well with EEE problems in visual contexts. A representative failure case is shown in \autoref{all_comparison_2}.  For example, GPT-4o achieves the highest average accuracy of just 46.78\% (less than half), and the average accuracies of proprietary and open-source models are only 35.28\% and 23.19\%, respectively.  These results indicate that current LMMs are inadequate in providing satisfactory assistance for real-world engineering tasks with their existing performance levels.

\paragraph{Finding 2: Open-source models see superior generalization compared to closed-source models. } 
Similar to prior works on benchmarking the performance of LLMs and LMMs in mathematics and sciences \citep{lu2023mathvista, yue2024mmmu}, we found it evident that a significant performance disparity exists between closed-source and open-source models, as shown in \autoref{main_result}.. The top two best-performing closed-source LMMs, GPT-4o and Gemini-Pro 1.5, significantly outperform their open-source counterparts. For instance, GPT-4o, the best-performing closed-source LMM, has an average accuracy of 46.78\%, while the highest-scoring open-source LMM, InternVL2-26B, achieves only 26.89\%—a gap of nearly 20 percentage points. Moreover, even the lightweight closed-source models, GPT-4o-mini and Gemini-Flash 1.5, surpass all open-source LMMs in performance. This discrepancy not only underscores the superior generalization capability of state-of-the-art (SOTA) closed-source models but also underscores the pressing need for the development of open-source LMMs that are proficient in EEE tasks.

\paragraph{Finding 3: Diverse subdomains present different challenges.} 
As discussed in \autoref{bench_ana}, problems across various subjects demand specific capabilities for effective resolution.  This section delves into the performance of existing LMMs across diverse subjects and discusses their strengths and weaknesses.

It is noteworthy that most closed-source LMMs excel in handling problems related to Communication Systems and Electronic Devices, yet they perform significantly poorer on problems from Analog Circuits and Circuit Theory and Network Analysis. For instance, the best-performing LMM GPT-4o achieves average accuracies of 60.40\% and 56.16\% in Communication Systems and Electronic Devices respectively, which are substantially higher than the overall average accuracy of 46.78\%. In contrast, its performance drops to 44.60\% and 40.88\% in Analog Circuits and Circuit Theory and Network Analysis, respectively, both of which fall below the overall average.

Problems within Communication Systems and Electronic Devices typically present straightforward visual information, yet solving them demands considerable computational effort. Conversely, problems from Analog Circuits and Circuit Theory and Network Analysis often feature complex visual contexts, such as intricate circuit diagrams, necessitating the use of advanced mathematical and computational tools. Thus, tackling issues in these domains requires not just strong numerical capabilities but also sophisticated vision recognition and reasoning skills. This performance disparity highlights that while many existing closed-source LMMs are adept at managing computationally intensive tasks due to their robust numerical calculation capabilities, they struggle with EEE problems that require more nuanced vision recognition and complex reasoning abilities. This suggests a crucial area for further development in enhancing the vision and reasoning capabilities of LMMs to better handle the full spectrum of EEE challenges. 

Conversely, open-source LMMs face challenges in handling tasks related to Communication Systems and Electronic Devices, despite the simplicity of the associated visual contexts. This performance discrepancy between open-source and closed-source LMMs in these domains underscores a significant shortfall in the computational capabilities of the open-source models. Urgent development is needed to enhance the computational efficiency and numerical problem-solving abilities of open-source LMMs. 

\subsection{Error Analysis} 
\label{error analysis}
\begin{figure*}[t]
    \centering
    \begin{minipage}{0.99\linewidth}
            \centering
            \vskip-5pt
            \includegraphics[width=\linewidth]{   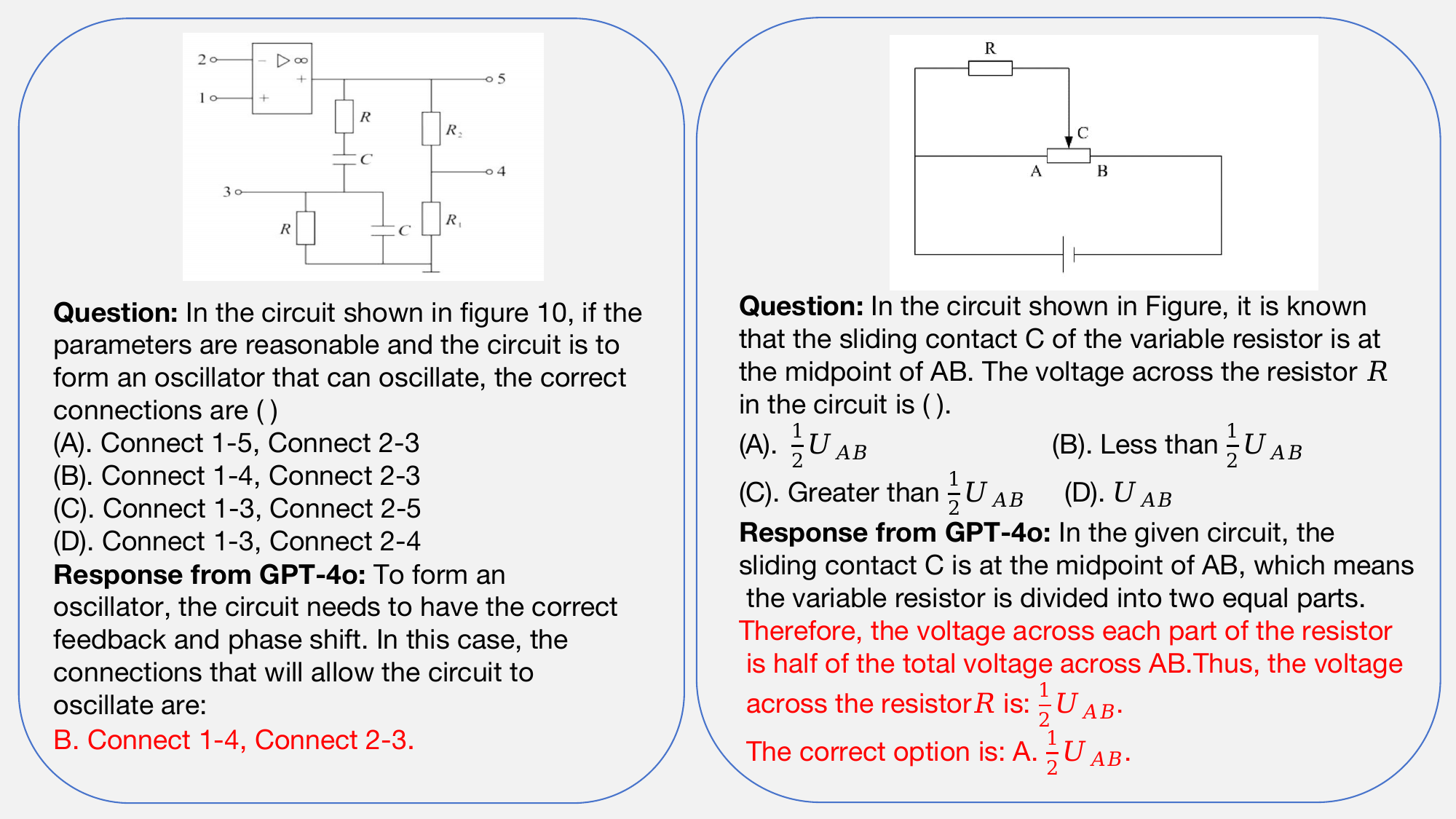}
            \vskip-5pt
    \end{minipage}
    \caption{ Error examples of response from GPT4o. Left: Image Perception Error. Right: Reasoning Error. Error part is marked with {\color{red}red} color.}
    \label{error-exp}
    \vskip -10px
\end{figure*}

In this section, we provide an in-depth analysis of the common errors observed in evaluations on EEE-Bench using GPT-4o as an example. \autoref{err_fig} shows the breakdown by error measured by 6 types: reasoning error, image perception error, question misunderstood error, knowledge error, calculation error, and rejection to answer. We focus our analysis on the two largest error groups, \textbf{reasoning error} and \textbf{perception error}, constituting \textbf{50.3\%} and \textbf{26.5\%} of total error respectively.  
\paragraph{Lack of reasoning.} These happen when LMMs can read the image but cannot reason correctly to arrive at the answer. For instance, on the right of \autoref{error-exp}, the LMM correctly reads the location of contact C but fails to consider the location of resistor $R$ in the overall circuit, thus leading to the wrong answer. 
\paragraph{Lack of visual understanding.} These happen when LMM fails to answer correctly due to misunderstandings of the image compositionally. Concretely, the circuit diagram on the left of \autoref{err_fig} shows a filter circuit using an operational amplifier with multiple feedback loops involving resistors and capacitors. In such cases even if the explanation is fine the verdict is wrong as the diagram is more complex than current LMMs can handle.
\label{exp: ana}

The examples above clearly underline the challenge of engineering problems to LMMs. We provide more of such instances in \autoref{appendix: more failure example}. Compared to other domains such as mathematics, these results demonstrate that engineering problems are intrinsically more \textbf{complex} and \textbf{dynamic} as components in domains such as electric engineering could be arranged in very different but meaningful ways or layered up to form more complicated diagrams whereas their counterparts in, for example, mathematics are mostly stagnated and fixed. On the other hand, it highlights the fact that LMMs struggle to understand electric engineering diagrams, and are especially prone to increased visual complexity and details such as resisters, signs, etc. Overall, it indicates that solving these electric engineering problems requires more delicate incorporations of both visual and textual inputs.

\section{Discussion}
\label{discussion}
\paragraph{The unique challenge of EEE problems.} Here we address the unique challenge of visual complexity in EEE-Bench. \autoref{exp:results} demonstrates that even SOTA LMMs struggle to understand and reason about EEE problems. To understand this phenomenon further, we design and conduct a simple experiment which is to provide additional information in the prompt for assistance. Concretely, we generate for each image a caption $X_c$ using GPT-4o to let it describe very detailed key visual information related to potential EEE problem-solving (the prompt for caption generation is provided in \autoref{app:prompt}) and prepend such caption to the original question prompt $X_q$ such that the new query is $(X_{ins} + X_c) + X_q$ where $X_{ins}$ is a system instruction prompt which reads ``\textit{Image caption context:}".
Surprisingly, as shown in \autoref{tab:model_comparison}, the improvements on the EEE-Bench with such caption are \textbf{negative} across 7 out of 10 subdomains, with an average performance change of -1.43\% using GPT-4o.
\autoref{error_caption_1} and \autoref{error_caption_2} show two of such examples. Technically speaking, captions generated by GPT-4o should include key and correct information from the images. However, in our case, it is apparent that such knowledge fails to be useful and even provides negative guidance in some cases. Importantly, such an effect evidently implies the intense difficulty of the images in EEE-Bench, which we believe is due to the high level of \textbf{abstraction} and \textbf{versatility} as well as the abundance of \textbf{domain-specific details} contained in these images, especially compared to other domains like mathematics. More intriguingly, it raises a more significant question about LMMs: \textbf{\textit{do LMMs rely more on vision or text?}} 


\paragraph{Your LMM might be lazy.} In this section we attempt to uncover the underlying cause of the phenomenon raised above. Previous works \citep{liu2024phd,qian2024easy,liu2023mitigating} have explored hallucination problems of recognition of nonexisting items in natural images. In contrast, our research delves into this issue with technical images, emphasizing the comprehension of image information and reasoning rather than object recognition. We design a straightforward experiment that prepends, instead of the accurate caption $X_c$, a spurious caption $X'_{c}$ which contains a spurious caption of the image augmented from the original caption by GPT-4 (the prompt for spurious caption generation is provided in \autoref{app:prompt}). Instead of focusing on objects, we intentionally make such argumentation small in change but major in the meaning of existing objects in images, for example, changing ``\textit{increasing}" into ``\textit{decreasing}" when describing a function like shown in \autoref{lazy_fig}. These small changes need not only careful recognition but also thorough understanding about the images. It is important to note that we ask specifically in our query for LMMs to answer based on the \textit{figure} or \textit{image} such that the captions merely serve as additional information. We display results on EEE-Bench in \autoref{tab:model_comparison} and on MathVista \citep{lu2023mathvista} in \autoref{spurious_mathvista}.
 Interestingly, across the board, the results show a significant negative impact, measuring a 7.79\% and 8.4\% drop on EEE-Bench and MathVista respectively using GPT-4o, where the same can be found with Gemini-Pro 1.5. For example, \autoref{lazy_fig} shows a case on EEE-Bench where LMMs fail to reason correctly due to spurious instruction which merely changes ``\textit{increasing}" into ``\textit{decreasing}", even if the function in the image is clearly increasing. More of such examples are displayed in \autoref{lazy_1} through \autoref{lazy_8} for EEE-Bench and in \autoref{lazy_9} through \autoref{lazy_10} for MathVista for reference. These results demonstrate that LMMs are severely distracted by the spurious prompt. On one hand, spurious captions may indeed hallucinate LMM's reasoning process. On the other hand, the instruction clearly states answering questions with the \textit{figure} or \textit{image}. This observation unveils a key shortcoming of LMMs that they may simply be lazy when both visual and textual inputs are given. We note that a recent work \citep{shi2023largelanguagemodelseasily} indeed finds that LLMs are easily disturbed by irrelevant content. We thus believe that the shortcoming revealed by our study is noteworthy for further detailed investigations as it is a major issue in many practical cases.
 


\begin{figure}[t]
\caption{Examples of LMM laziness in EEE-Bench. The spurious text in the caption and error in response is marked with {\color{red}red} color. The LMM used here is GPT-4o.}
\label{lazy_fig}
\hrule
\centering
\begin{minipage}{0.2\textwidth}
    \centering
    \includegraphics[width=\textwidth]{  } 
\end{minipage}
\hfill
\begin{minipage}{0.7\textwidth}
\small
    \textbf{Question: } Given that the waveform of \( \boldsymbol{f}(\boldsymbol{t}) \) is as shown in the figure, the expression for \( \boldsymbol{f}(\boldsymbol{t}) \) is ( ).

A. \( \boldsymbol{f}(\boldsymbol{t}) = \boldsymbol{t} \varepsilon(\boldsymbol{t}) - \boldsymbol{t} \varepsilon(\boldsymbol{t} - 1) \)  
B. \( \boldsymbol{f}(\boldsymbol{t}) = (1 - \boldsymbol{t}) \varepsilon(\boldsymbol{t}) - (1 - \boldsymbol{t}) \varepsilon(\boldsymbol{t} - 1) \)  \\
C. \( \boldsymbol{f}(\boldsymbol{t}) = \boldsymbol{t} \varepsilon(\boldsymbol{t}) - (\boldsymbol{t} - 1) \varepsilon(\boldsymbol{t} - 1) \)  
D. \( \boldsymbol{f}(\boldsymbol{t}) = (1 + \boldsymbol{t}) \varepsilon(\boldsymbol{t}) - (\boldsymbol{t} + 1) \varepsilon(\boldsymbol{t} + 1) \)

\end{minipage}
\hfill
\begin{minipage}[c]{0.98\textwidth}
\small
\textbf{Spurious context: }Image caption context:

The diagram is a plot of a function \( f(t) \) versus time \( t \). Here is a detailed description:

- Axes: 
  - The horizontal axis represents time \( t \).
  - The vertical axis represents the function \( f(t) \).

- Function Behavior:
  - The function starts at the origin, \( (0, 0) \).
  - From \( t = 0 \) to \( t = 1 \), the function {\color{red}decreases linearly from 1 to 0}. This indicates a slope of {\color{red}-1}.
  - At \( t = 1 \), the function reaches a value of {\color{red}0}.
  - For \( t > 1 \), the function remains constant at \( f(t) = {\color{red}0} \).

- Key Points:
  - \( f(0) = {\color{red}1} \)
  - \( f(1) = {\color{red}0} \)
  - The slope from \( t = 0 \) to \( t = 1 \) is {\color{red}-1}.
  - The function is constant for \( t > 1 \).

This plot represents a {\color{red}unit ramp function that transitions into a unit step function at \( t = 1 \)}.

\textbf{Correct Response w/o spurious caption: }
The waveform \( f(t) \) is a ramp function starting at \( t = 0 \) and ending at \( t = 1 \), where it becomes constant. This can be expressed using unit step functions \( \varepsilon(t) \).

The correct expression is:
\[ f(t) = t \varepsilon(t) - (t - 1) \varepsilon(t - 1) \]
This matches option C.
 
\textbf{Wrong Response w/ spurious caption: }
The function \( f(t) \) {\color{red}starts at 1 and decreases linearly to 0 from \( t = 0 \) to \( t = 1 \), then remains constant at 0 for \( t > 1 \)}. This behavior can be represented using the unit step function \( \varepsilon(t) \).

The correct expression for \( f(t) \) is:
\[ f(t) = (1 - t) \varepsilon(t) - (1 - t) \varepsilon(t - 1) \]

This matches the description of the function, where it decreases linearly from 1 to 0 and then stays at 0.
 
{\color{red}The correct option is:
**B**}
\vspace{0.4em}
\end{minipage}
\hrule
\hrule
\end{figure}




\section{Related Works}
\subsection{Large Multimodal Models}
Following the rapid development and success of Large Language Models (LLMs) \citep{brown2020language,jiang2024mixtral,touvron2023llama,wei2021finetuned}, Large Multimodal Models (LMMs) have been constructed by integrating LLMs with large vision encoders \citep{radford2021learning,caron2021emerging,oquab2023dinov2,chen2021empirical}. Notably, closed-source versatile LMMs such as GPT-4o \citep{openai2024gpt4} and Gemini-Pro 1.5 \citep{team2023gemini} have achieved exceptional visual reasoning capabilities across various domains, including Visual Question Answering \citep{li2024seed} and Mathematics \citep{lu2023mathvista}. In response to the need for publicly accessible versatile LMMs, several open-source models such as LLaVA \citep{liu2024visual} have been developed \citep{li2022blip,liu2024visual,liu2023improvedllava,gao2023llama,zhang2023llama,li2023blip,chen2023minigpt,ye2023mplug,tong2024cambrian1}. Moreover, finetuning \citep{zhou2022learning,li2024vision,hu2021lora,zhou2022conditional} aims to improve the performance of LMMs using downstream data. Specialized LMMs finetuned on specific subdomain data \citep{wang2023parameter,hu2021lora} have been introduced, demonstrating performance on par with closed-source models in specific subject areas like Mathematics \citep{shi2024math,zhang2024mavis,zhuang2024math} and chart \citep{meng2024chartassisstant,hu2024mplug}. Recent efforts have also focused on leveraging large models for engineering tasks, such as chip design \citep{bose2000energy,bose2020power,irwin2020basic,chang2024data} and optimizing operations in the energy sector \citep{majumder2024exploring}. However, the reasoning capabilities of existing LMMs in visual contexts have not yet been thoroughly examined for electrical and electronics engineering (EEE) problems. Therefore, this paper proposes EEE-Bench to comprehensively evaluate the EEE multimodal reasoning abilities of current LMMs.
\begin{table}[t!]
    \vskip -5px
    \centering 
     \adjustbox{max width=\textwidth}{
    \begin{tabular}{lccccccccccc}
        \toprule
        Model & ALL & EMag & PEPS & SS & EMac & DLC & CorS & AC & CTNA & ComS & ED \\
        \midrule
        \multicolumn{12}{c}{GPT-4o} \\
        \midrule
        GPT4o & 46.78 & 50.98 & 47.6 & 46.38 & 49.69 & 47.78 & 46.03 & 44.6 & 40.88 & 60.4 & 56.16 \\
        w/ caption & 45.35 & 48.04 & 41.48 & 44.93 & 44.72 & 43.75 & 44.35 & 46.31 & 42.23 & 48.51 & 57.99 \\
        $\Delta$& \textcolor{red}{-1.43} & \textcolor{red}{-2.94} & \textcolor{red}{-6.12} & \textcolor{red}{-1.45} & \textcolor{red}{-4.97} & \textcolor{red}{-4.03} & \textcolor{red}{-1.68} & \textcolor{blue}{+1.71} & \textcolor{blue}{+1.35} & \textcolor{red}{-11.89} & \textcolor{blue}{+1.83} \\
         w/ spurious caption & 38.99 & 50.00 & 38.43 & 44.93 & 36.02 & 40.12 & 38.91 & 38.42 & 33.11 & 45.54 & 44.75 \\
        $\Delta$ & \textcolor{red}{-7.79} & \textcolor{red}{-0.98} & \textcolor{red}{-9.17} & \textcolor{red}{-1.45} & \textcolor{red}{-13.67} & \textcolor{red}{-7.66} & \textcolor{red}{-7.12} & \textcolor{red}{-6.18} & \textcolor{red}{-7.77} & \textcolor{red}{-14.86} & \textcolor{red}{-11.41} \\
        \midrule
        \multicolumn{12}{c}{Gemini-Pro 1.5} \\
        \midrule
        Gemini-Pro 1.5 & 41.99 & 50 & 46.29 & 42.75 & 36.65 & 38.71 & 46.03 & 41.85 & 38.34 & 54.46 & 44.75 \\
         w/ spurious caption & 35.21 & 46.08 & 41.92 & 38.41 & 37.89 & 36.9 & 38.49 & 31.39 & 29.39 & 38.61 & 36.07 \\
        $\Delta$ & \textcolor{red}{-6.78} & \textcolor{red}{-3.92} & \textcolor{red}{-4.37} & \textcolor{red}{-4.34} & \textcolor{blue}{+1.24} & \textcolor{red}{-1.81} & \textcolor{red}{-7.54} & \textcolor{red}{-10.46} & \textcolor{red}{-8.95} & \textcolor{red}{-15.85} & \textcolor{red}{-8.68} \\
        \bottomrule
    \end{tabular}
    }
    \caption{Performance comparison of GPT-4o and Gemini-Pro 1.5 with and without spurious captions. Positive changes are in \textcolor{blue}{blue} and negative changes in \textcolor{red}{red}.}
    \vskip -10px
    \label{tab:model_comparison}
\end{table}

\subsection{Multimodal benchmarks}
As Large Language Models (LLMs) and Large Multimodal Models (LMMs) continue to evolve rapidly, the creation of comprehensive and challenging benchmarks to assess their capabilities has become increasingly vital \citep{chang2024survey,cui2024survey}. These benchmarks are indispensable for evaluating the strengths and limitations of these models, thereby facilitating their ongoing development and refinement. To measure the abilities of LLMs, various text-only benchmarks have been introduced \citep{hendrycks2020measuring,hendrycks2021measuring,zheng2023judging,zellers2019hellaswag,jimenez2023swe}. For vision and language benchmarks, early efforts primarily concentrated on general-purpose Visual Question Answering (VQA) tasks \citep{mathew2021docvqa,gurari2018vizwiz,antol2015vqa,goyal2017making,singh2019towards,hudson2019gqa,mathew2021docvqa,marino2019ok}, which may not sufficiently test the reasoning abilities of today's more advanced LMMs. To provide a multi-faceted evaluation, several benchmarks have been introduced, including specialized ones \citep{lu2023mathvista, zhang2024mathverse,qiao2024we,wang2024measuring,wang2024charxiv,li2024mmcode,lu2021inter} and versatile benchmarks \citep{li2024seed,liu2023mmbench,liang2024scemqa,saikh2022scienceqa,yue2024mmmu,wang2023scibench,chen2023theoremqa}. For instance, MathVista \citep{lu2023mathvista} assesses the mathematical reasoning capabilities of LMMs across various visual contexts. However, the visual context in mathematical problems is much less complex compared with real-world engineering problems that appeared in EEE-Bench. MMMU \citep{yue2024mmmu} tackles college-level questions that require intricate, domain-specific knowledge across multiple subjects. Although MMMU also includes some EEE problems, its scope in terms of problem variety and subject coverage is limited, which may not thoroughly evaluate the reasoning capabilities of LMMs on EEE tasks. \cite{doris2024designqamultimodalbenchmarkevaluating} benchmarks LMMs' understanding of engineering requirement documents, yet they are not directly related to solving engineering problems. To the best of our knowledge, we respectively believe EEE-Bench, which comprises 2860 meticulously curated problems from ten pivotal EEE subjects, is the first in the direction to offer a robust and comprehensive assessment of LMMs’ reasoning abilities in complex visual contexts within EEE. 
\section{Conclusion}
This study focuses on benchmarking modern LMMs' performance on EEE problems, an important representation of the border engineering domain and a high-impact subject in our daily lives (\autoref{intro}). We start by creating a pioneering electric engineering-based benchmark termed \textbf{EEE-Bench} consisting of \textbf{2860} hand-picked and carefully curated questions covering \textbf{10 subdomains} of electric engineering in \autoref{benchmark summary}. We perform evaluations and in-depth analysis of \textbf{17} open and closed-source LLMs (1) and LMMs (16) on EEE-Bench, such diverse coverage ensures a thorough and comparable assessment (\autoref{exp:results}). Importantly, we conclude that, in the field of EEE, as well as engineering in general, today's foundation models still struggle to understand and reason through intricate EEE questions in visual context (\autoref{exp: ana}), revealing the unmet promise of human-level LMMs in this domain. Our analysis shows that such deficiency lies in the unique challenge of engineering problems compared to other domains, namely the high level of \textbf{abstraction} and \textbf{versatility}, as well as \textbf{domain-specific information} (\autoref{discussion}). 
Furthermore, our study reveals an interesting ``laziness" phenomenon of LMMs, namely the tendency of \textbf{overly reliance} on and \textbf{easy distraction} by textual input thus overlooking visual context, suggesting a limitation noteworthy for future research (\autoref{discussion}). We hope EEE-Bench can advance the research for more capable LMMs for EEE and for engineering by providing a robust and comprehensive evaluation tool focusing on both understanding and reasoning while harnessing the bright potential of the usage of LMMs in more complex and real-world scenarios.
\newpage

\bibliography{iclr2025_conference}
\bibliographystyle{iclr2025_conference}

\newpage
\appendix

\section{Appendix Table of Content}
We provide all additional details for our paper in the following sections.
\begin{itemize}
    \item \autoref{impact} discusses the broader impact of EEE-Bench.
    \item \autoref{app:prompt} provides all prompts used in paper.
    \item \autoref{appendix: more experimental results} provides more experimental results. 
    \item \autoref{app:imp} provides the information of the models used in experiments and their hyper-parameters.
    \item \autoref{appendix: more about eee-bench} provides more example images about EEE-Bench.
    \item \autoref{appendix:detail_intro} provides detailed introduction to the 10 pivotal subjects in EEE-Bench.
    \item \autoref{appendix:lmm_abilities} discusses the specific required abilities for LMM to address the challenges in each subject, and shows examples of problem solutions for each subject. 
    \item \autoref{more_error} provides more qualitative examples of error analysis. 
    \item \autoref{app:more_lazy} provides more examples of the laziness phenomenon of existing LMMs.
\end{itemize} 

\section{Broader Impact} 
\label{impact}
\paragraph{Limitations and future work.} While EEE-Bench rigorously covers 10 critical subdomains of EEE (\autoref{appendix: more about eee-bench}), it does not extend to the broader spectrum of engineering (\emph{e.g.}, civil, mechanical, and chemical engineering). Although we believe our results can provide compelling insights that may generalize to other engineering domains, comprehensive and domain-specific studies are needed to thoroughly and accurately evaluate LMMs' capabilities across the full engineering landscape. We plan to carry out such studies and extend EEE-Bench to broader engineering domains in future works.  

\paragraph{Ethics statement.}
We ensure that EEE-Bench complies with legal and ethical guidelines throughout
its construction process, with no violations.  We provide fair compensation to all annotators and graduate students involved. We collect problems from free-access public open exams \footnote{https://gate2025.iitr.ac.in/}  and resources under the CC Licenses \footnote{https://www.allaboutcircuits.com/worksheets/}\footnote{https://ocw.mit.edu/}.  We guarantee that EEE-Bench is solely for academic research purposes, and we uphold the strict prohibition of any commercial use. The spurious captions
generated in \autoref{discussion}  are only related to problem-solving and do not pose any harm to humans. 

\paragraph{Reproducibility statement.} We are committed to efficient and reproducible research. Our benchmark along with complete source code will be released.
\section{Used prompts}
\label{app:prompt}
\subsection{Prompt for Caption Generation} 
The prompt for caption generation using GPT-4o is as follows: 

\textit{Here is a diagram related to an Electrical and Electronics Engineering problem. Please describe the diagram in complete detail, ensuring to include all relevant components, symbols, connections, values (such as resistances, voltages, or frequencies), and any other key parameters. Your description should be precise enough to fully replace the diagram and be usable for solving the EEE problem across topics like circuits, signal processing, electromagnetics, power systems, and control systems.}

\subsection{Spurious caption generation prompt}
The prompt for spurious caption generation using GPT-4 on EEE-Bench is as follows:

\textit{I will give a caption of an Electrical and Electronics Engineering (EEE) problem image and the associated question. Please help me rewrite the image caption so that it contains incorrect information, which could lead to an incorrect solution. Please only output the revised incorrect caption.
Please do not change name of variables, e.g., F, G. Please Do not change too much. Here are some examples of how you can introduce incorrect information into an Electrical and Electronics Engineering (EEE) image caption to potentially mislead a model into generating a wrong solution:\\
1. Incorrect component identification:
Original: "The circuit contains a 10$\Omega$ resistor and a 5V voltage source."
Misleading: "The circuit contains a 100$\Omega$ resistor and a 12V voltage source."\\
2. Wrong power supply voltage:
Original: "The circuit is powered by a 5V DC supply."
Misleading: "The circuit is powered by a 15V AC supply."\\
3. Inverted polarity for components:
Original: "The diode is forward biased in this circuit."
Misleading: "The diode is reverse biased in this circuit."\\
4. Incorrect current direction:
Original: "The current flows clockwise through the loop."
Misleading: "The current flows counterclockwise through the loop."\\
5. Swapping component values:
Original: "The capacitor has a capacitance of 100$\mu$F, and the inductor is 10mH."
Misleading: "The capacitor has a capacitance of 10$\mu$F, and the inductor is 100mH."\\
6. Incorrect assumptions about the circuit operation:
Original: "The transistor is in active mode."
Misleading: "The transistor is in saturation mode."\\
7. Wrong frequency or waveform type:
Original: "The AC signal is a 60Hz sine wave."
Misleading: "The AC signal is a 50Hz square wave."\\
8. Incorrect type of circuit:
Original: "This is a simple RC low-pass filter."\\
Misleading: "This is a high-pass filter."
9. Incorrect labeling of nodes or points:
Original: "Measure the voltage at node A, which is connected to the positive terminal."
Misleading: "Measure the voltage at node A, which is connected to the negative terminal."\\
10. Misleading behavior of the circuit:
Original: "The capacitor charges fully after 5 seconds."
Misleading: "The capacitor discharges completely after 5 seconds."\\
11. Incorrect power consumption:
Original: "The total power consumed by the circuit is 50W."
Misleading: "The total power consumed by the circuit is 100W."\\
12. Mislabeling passive vs. active components:
Original: "The operational amplifier is used as an active component in this circuit."
Misleading: "The operational amplifier is used as a passive component in this circuit."\\
13. Incorrect reading:
Original: "Only when A=1,B=1,C=1, F=1."
Misleading: "when A=1,B=1,C=1, F=0, when A=0,B=1,C=1, F=1.}

\begin{table}[t]
\centering
\caption{ The task instructions for different question and answer types in answer extraction. Here,
Float (1) refers to a floating-point number with one decimal place, Float (2) refers to a floating-point number with two decimal places, and Float (3) refers to a floating-point number with three decimal places.}
\label{instru_prompt}
\begin{tabular}{ccp{10cm}}
\hline
\vspace{-0.2em}
\textbf{Question type} & \textbf{Answer type} & \textbf{Task instruction} \\ \hline
\multirow{2}{*}{multiple-choice} & \multirow{2}{*}{Text} & Please answer the question and provide the correct option letter, e.g., A, B, C, D, at the end. \\ \hline
\multirow{2}{*}{Free-form} & \multirow{2}{*}{Integer} & Please answer the question requiring an integer answer and provide the final value, e.g., 1, 2, 3, at the end. \\ \hline
\multirow{2}{*}{Free-form} & \multirow{2}{*}{Float (1)} & Please answer the question requiring a floating-point number with one decimal place and provide the final value, e.g., 1.2, 1.3, 1.4, at the end. \\ \hline
\multirow{3}{*}{Free-form} & \multirow{3}{*}{Float (2)} & Please answer the question requiring a floating-point number with two decimal places and provide the final value, e.g., 1.23, 1.34, 1.45, at the end. \\ \hline
\multirow{3}{*}{Free-form} & \multirow{3}{*}{Float (3)} & Please answer the question requiring a floating-point number with three decimal places and provide the final value, e.g., 1.200, 1.300, 1.400, at the end.\\
\hline
\multirow{2}{*}{Free-form} & \multirow{2}{*}{List} & Please answer the question requiring a Python list as an answer and provide the final list, e.g., [1, 2, 3], [1.2, 1.3, 1.4], at the end. \\ \hline
\end{tabular}
\end{table}

The prompt for spurious caption generation using GPT-4 on Math-Vista is as follows:

\textit{I will provide a caption of a Math problem image along with the associated question. Please help me rewrite the image caption so that it contains incorrect information, potentially leading to an incorrect solution. Please only provide the revised incorrect caption. Do not change the names of variables, such as F or G, and avoid making extensive changes. \\
Please do not change the name of variables, e.g., F, G. Please Do not change too much. Please do not change the essential meaning of original text and only change subtle places like values.\\
Below are some examples of how you can introduce incorrect information into a Math problem caption to mislead a model:\\
Incorrect value of constants: Original: 'The equation has a constant value of 3.' Misleading: 'The equation has a constant value of 5.'\\
Wrong interpretation of angles: Original: 'Angle ABC is 90$^\circ$.' Misleading: 'Angle ABC is 60$^\circ$.'\\
Incorrect function transformation: Original: 'The function is f(x) = 2x + 3.' Misleading: 'The function is f(x) = 2x - 3.'\\
Wrong area formula: Original: 'The area of the circle is given by $\pi r^2$.' Misleading: 'The area of the circle is given by $2\pi r$.'\\
Incorrect limit definition: Original: 'The limit of the function as x approaches 2 is 4.' Misleading: 'The limit of the function as x approaches 2 is 5.'\\
Wrong assumption in calculus: Original: 'The derivative of $x^2$ is 2x.' Misleading: 'The derivative of $x^2$ is x.'\\
Incorrect triangle property: Original: 'In this right triangle, the hypotenuse is 10.' Misleading: 'In this right triangle, the hypotenuse is 8.'\\
Wrong direction in vector problems: Original: 'The vector points in the positive x-direction.' Misleading: 'The vector points in the negative x-direction.}
\subsection{Prompts for response generation}
We follow MathVista \citep{lu2023mathvista} to prepend an instructional prompt header into each question. The prompt can be found in \autoref{instru_prompt}.

\begin{table*}[t]
    \centering
    \caption{Generating parameters for Closed-Source LMMs.}
    \begin{tabular}{l | p{0.6\textwidth}}
    \toprule
    \textbf{Model} & \textbf{Generation Setup} \\
    \midrule
    
    GPT-4 & "model" : "gpt-4", "temperature" : 0, "max\_tokens" : 1024\\
    \midrule
    GPT-4o & "model" : "gpt-4o-2024-08-06", "temperature" : 0, "max\_tokens" : 1024 \\
    \midrule
    GPT-4V & "model" : "gpt-4-turbo", "temperature" : 0, "max\_tokens" : 1024 \\
    \midrule
    GPT-4o-mini & "model" : "gpt-4o-mini", "temperature" : 0, "max\_tokens" : 1024 \\
    \midrule
    Gemini 1.5 Pro & "model" : "gemini-1.5-pro", "temperature" : 0, "max\_tokens" : 1024 \\
    \midrule
    Gemini 1.5 Flash & "model" : "gemini-1.5-flash", "temperature" : 0, "max\_tokens" : 1024\\
    \midrule
    Qwen-VL-Max & "model" : "qwen-vl-max", "temperature" : 0, "max\_tokens" : 1024 \\
    \midrule
    Qwen-VL-Plus & "model" : "qwen-vl-plus", "temperature" : 0, "max\_tokens" : 1024\\
    \bottomrule
    \end{tabular}
    \label{api_setup}
\end{table*}

\begin{table*}[t]
    \centering
    \caption{Generating parameters for Open-Source LMMs.}
    \begin{tabular}{l | p{0.6\textwidth}}
    \toprule
    \textbf{Model} & \textbf{Generation Setup} \\
    \midrule
    Llava-1.6-7B & do\_sample = False, temperature = 0, max\_new\_tokens = 1024 \\
    \midrule
    Llava-1.6-13B & do\_sample = False, temperature = 0, max\_new\_tokens = 1024 \\
    \midrule
    SPHINX-v2-1k & do\_sample = False, temperature = 0, max\_new\_tokens = 1024 \\
    \midrule
    MiniCPM-V-2.6 & do\_sample = False, temperature = 0.1, max\_new\_tokens = 1024 \\
    \midrule
    DeepSeek-VL-7B & do\_sample = False, max\_new\_tokens = 1024 \\
    \midrule
    Phi3-Vision-4.2B & do\_sample = False, temperature = 0, max\_new\_tokens = 1024 \\
    \midrule
    InternVL2-26B & do\_sample = False, temperature = 0, max\_new\_tokens = 1024\\
    \midrule
    InternLM-XComposer2.5-7B& do\_sample = False \\
    \midrule
    GLM-4V-9B & do\_sample = False, max\_length = 1024,  temperature = 0 \\
    \bottomrule
    \end{tabular}
    \label{open_setup}
\end{table*}

\section{More Experimental Results}
\label{appendix: more experimental results}
In this section, we provide more experimental results.

\begin{figure}[t]
    \centering
    \begin{subfigure}{0.45\textwidth}
        \centering
        \includegraphics[width=\linewidth]{       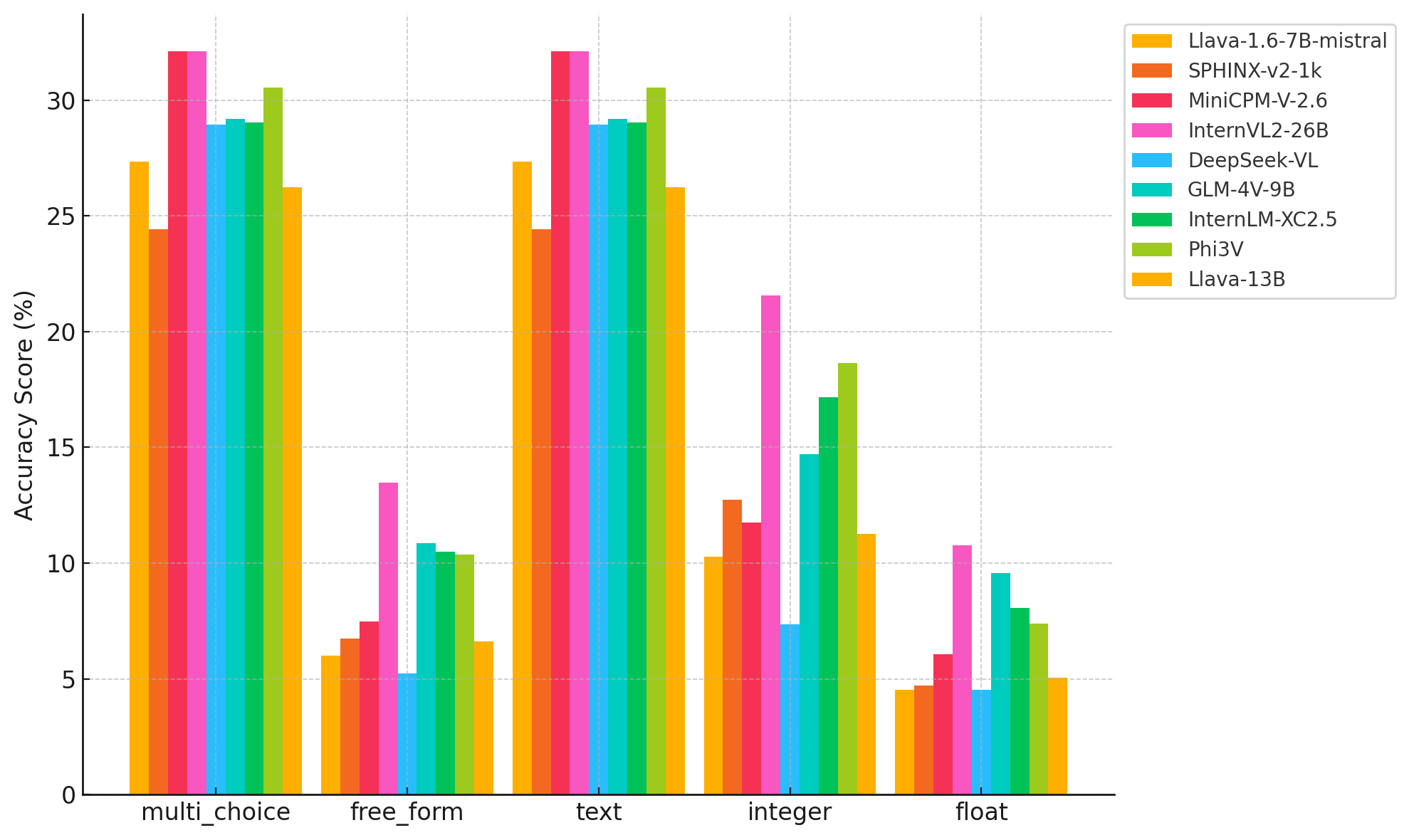}
        \caption{Open-source LMMs.}
    \end{subfigure}
    \hfill
    \begin{subfigure}{0.45\textwidth}
        \centering
        \includegraphics[width=\linewidth]{       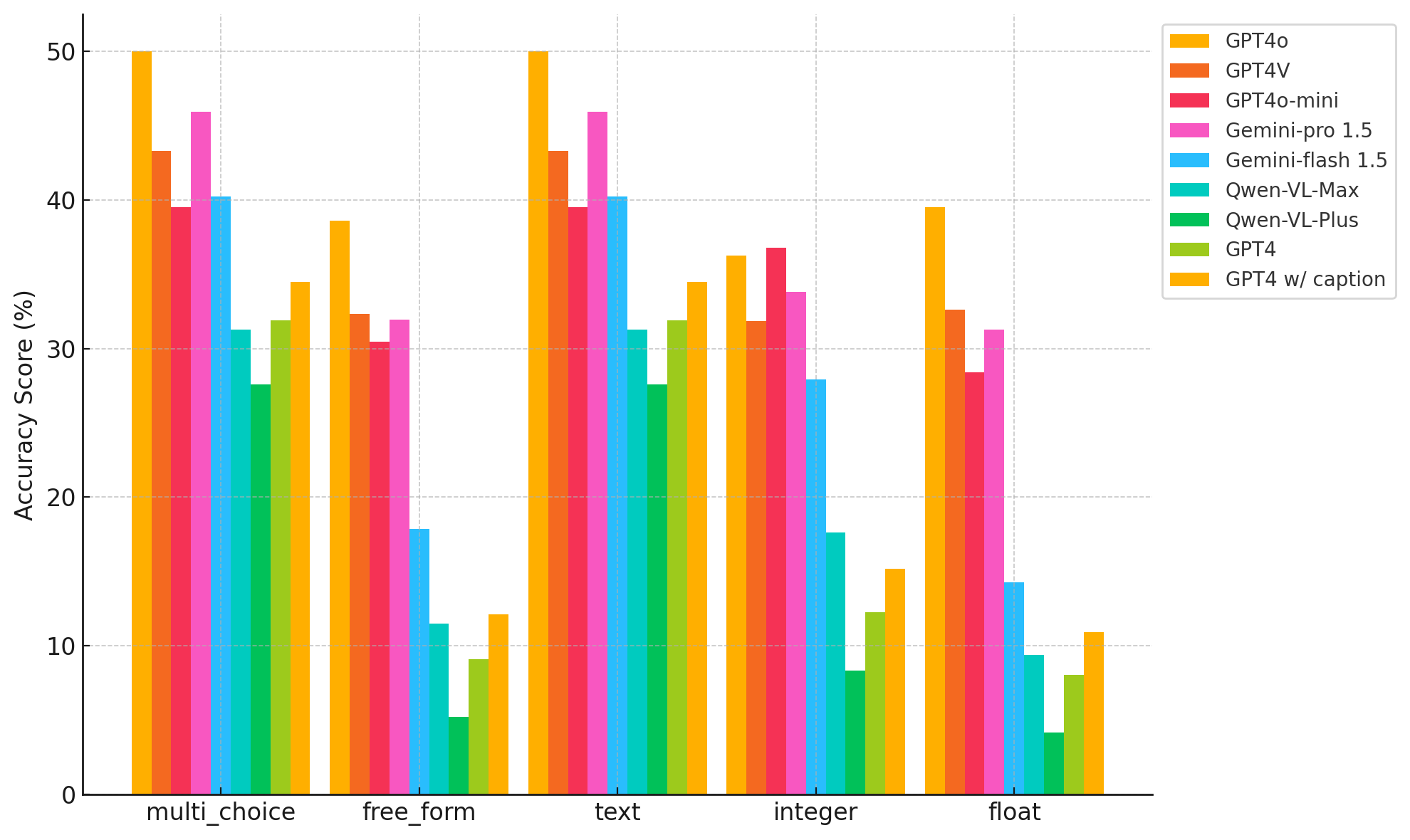}
        \caption{Cclosed-source LMMs.}
        
    \end{subfigure}
    
    \caption{Accuracy across different question types and answer types.}
    \label{app:type}
\end{figure}

\textbf{Results on different question types and answer types.} We first discuss the performance of LMMs on different question types: multi-choice and free-form, and answer types: text, integer, and float (since the number of list-type questions is too small, we ignore them here).  The results are shown in \autoref{app:type}. We can find that the open-source models perform very poorly on free-form problems. Since free-form problems need accurate answers, the computational ability of open-source models can not support them in obtaining accurate final answers. Compared with open-source models, closed-source models show a better balance of accuracy between multi-choice problems and free-form problems.

\textbf{Results on MathVista with spurious captions.} We provide the results on MathVista under the setting of spurious captions. The results are shown in \autoref{spurious_mathvista}. Similar to experiments done on EEE-Bench, adding spurious captions into text will significantly degrade LMMs' performance.

\begin{table}[t]
    \centering 
     \adjustbox{max width=\textwidth}{
    \begin{tabular}{lccccccccccccc}
        \toprule
        Model & ALL & GPS & TQA & FQA & MWP & VQA & SCI & STA & ALG & GEO & ARI & LOG & NUM \\
        \midrule
        \multicolumn{14}{c}{GPT-4o} \\
        \midrule
        $\Delta$ & \textcolor{red}{-8.40} & \textcolor{red}{-0.48} & \textcolor{red}{-12.03} & \textcolor{red}{-4.46} & \textcolor{red}{-21.51} & \textcolor{red}{-6.70} & \textcolor{blue}{+0.42} & \textcolor{red}{-3.20} & \textcolor{red}{-7.97} & \textcolor{blue}{+0.42} & \textcolor{red}{-12.18} & \textcolor{blue}{+2.70} & \textcolor{red}{-11.11} \\
        \midrule
        \multicolumn{14}{c}{Gemini-Pro 1.5} \\
        \midrule
        $\Delta$ & \textcolor{red}{-3.70} & \textcolor{red}{-6.20} & \textcolor{red}{-8.86} & \textcolor{blue}{+0.00} & \textcolor{red}{-5.38} & \textcolor{blue}{+0.00} & \textcolor{red}{-1.66} & \textcolor{red}{-1.66} & \textcolor{red}{-7.81} & \textcolor{red}{-6.27} & \textcolor{red}{-0.89} & \textcolor{red}{-2.02} & \textcolor{red}{-2.08} \\
        \bottomrule
    \end{tabular}
    }
    \caption{Performance Delta of using spurious caption as extra text on MathVista. ALL: overall accuracy. Task types: FQA:
figure question answering, GPS: geometry problem solving, MWP: math word problem, TQA: textbook question answering, VQA: visual question answering. Mathematical reasoning types: ALG:
algebraic reasoning, ARI: arithmetic reasoning, GEO: geometry reasoning, LOG: logical reasoning,
NUM: numeric commonsense, SCI: scientific reasoning, STA: statistical reasoning. Positive changes are in \textcolor{blue}{blue} and negative changes in \textcolor{red}{red}.}
\label{spurious_mathvista}
\end{table}

\textbf{Examples of response from LMMs. }
In \autoref{all_comparison_1} and \autoref{all_comparison_2}, we provide qualitative examples of responses for LMMs for reference. 

\textbf{Wrong responses from GPT-4o augmented with captions.} We discuss examples of wrong responses from GPT-4o with augmented captions. \autoref{error_caption_1} and \autoref{error_caption_2} show examples. We found that these errors stem from incorrectly generated captions, which aligns with the concept of LMM laziness.

\section{Model and Hyperparameter}
\label{app:imp}
The release time and source of open-source and closed-source models can be found in Table \ref{model_source}. For closed-sourced models, we access them with APIs and adopt the generation scheme shown in \autoref{api_setup}.
and then run the inference with CPUs, which typically completes within several hours. For all open-source
models, we utilize a cluster with 8 NVIDIA 40GB A100 GPUs to run the inference, and
we follow the hyper-parameter settings specified in \autoref{open_setup}.

\begin{table*}[th]
    \centering
    \caption{\textbf{The Release Time and Model Source of LMMs Used in EEE-Bench.}}
    \label{model_source}
    \resizebox{0.8\linewidth}{!}{ 
    \begin{tabular}{l@{\hspace{0.5cm}}c@{\hspace{1cm}}p{0.5\textwidth}}
    \toprule
    \textbf{Model} & \textbf{\makecell{Release\\ Time}} & \textbf{\makecell[c]{Source}} \\ \toprule
    GPT-4o    &   2024-05    & \url{https://gpt4o.ai/} \\
    \midrule
    \multirow{2}{*}{GPT-4V}    &   \multirow{2}{*}{2024-04}    & \url{https://openai.com/index/gpt-4v-system-card/} \\
    \midrule
    \multirow{2}{*}{GPT-4o-mini} & \multirow{2}{*}{2024-07} &{\href{https://openai.com/index/gpt-4o-mini-advancing-cost-efficient-intelligence}{\textmd{https://openai.com/index/gpt-4o-mini-advancing-cost-efficient-intelligence}}}\\
    \midrule
    \multirow{2}{*}{Gemini 1.5 Pro}  & \multirow{2}{*}{2024-05} & \url{https://deepmind.google/technologies/gemini/pro/}\\ 
    \midrule
    \multirow{2}{*}{Gemini 1.5 Flash}  & \multirow{2}{*}{2024-05}&\url{https://deepmind.google/technologies/gemini/flash/}
    \\
    \midrule
    \multirow{3}{*}{Qwen-VL-Plus}        & \multirow{3}{*}{2023-11}  & \url{https://help.aliyun.com/zh/dashscope/developer-reference/vl-plus-quick-start} \\
    \midrule
    \multirow{3}{*}{Qwen-VL-Max}            & \multirow{3}{*}{2024-01} & \url{https://help.aliyun.com/zh/dashscope/developer-reference/vl-plus-quick-start} \\
    \midrule
    \multirow{2}{*}{LLaVA-1.6-7B}         &    \multirow{2}{*}{2024-03}       & \url{https://huggingface.co/liuhaotian/llava-v1.6-mistral-7b} \\
    \midrule
    \multirow{2}{*}{LLaVA-1.6-13B}       &   \multirow{2}{*}{2024-03}    &        \url{https://huggingface.co/liuhaotian/llava-v1.6-vicuna-13b} \\
    \midrule
    \multirow{3}{*}{SPHINX-Plus }            & \multirow{3}{*}{2023-11} & \url{https://huggingface.co/Alpha-VLLM/LLaMA2-Accessory/tree/main/finetune/mm/SPHINX/SPHINX-v2-1k} \\
    \midrule
    {MiniCPM-V-2.6 }       &   {2024-08}    &        \url{https://github.com/OpenBMB/MiniCPM-V} \\
    \midrule
    \multirow{2}{*}{DeepSeek-VL-7B}       &   \multirow{2}{*}{2024-03}   &        \url{https://huggingface.co/deepseek-ai/deepseek-vl-7b-chat} \\
    \midrule
    \multirow{2}{*}{Phi3-Vision-4.2B}       &   \multirow{2}{*}{2024-05}    &        \url{https://huggingface.co/microsoft/Phi-3-vision-128k-instruct} \\
    \midrule
    \multirow{2}{*}{InternVL2-26B}        &   \multirow{2}{*}{2024-07}     & \url{https://huggingface.co/OpenGVLab/InternVL2-26B} \\
    \midrule
    \multirow{2}{*}{InternLM-XComposer2.5-7B}         &    \multirow{2}{*}{2024-07}       & \url{https://huggingface.co/internlm/internlm-xcomposer2d5-7b} \\
    \midrule
    \multirow{2}{*}{GLM-4V-9B} & \multirow{2}{*}{ 2024-06}    & \url{https://huggingface.co/THUDM/glm-4v-9b} \\
    \bottomrule
    \end{tabular}
    }
    \label{supp-t0.5}
\end{table*}

\section{More About EEE-Bench} 
\label{appendix: more about eee-bench}
\begin{figure*}[t]
  
    \centering
    \begin{minipage}{0.99\linewidth}
            \centering
            \vskip-5pt
            \includegraphics[width=\linewidth]{     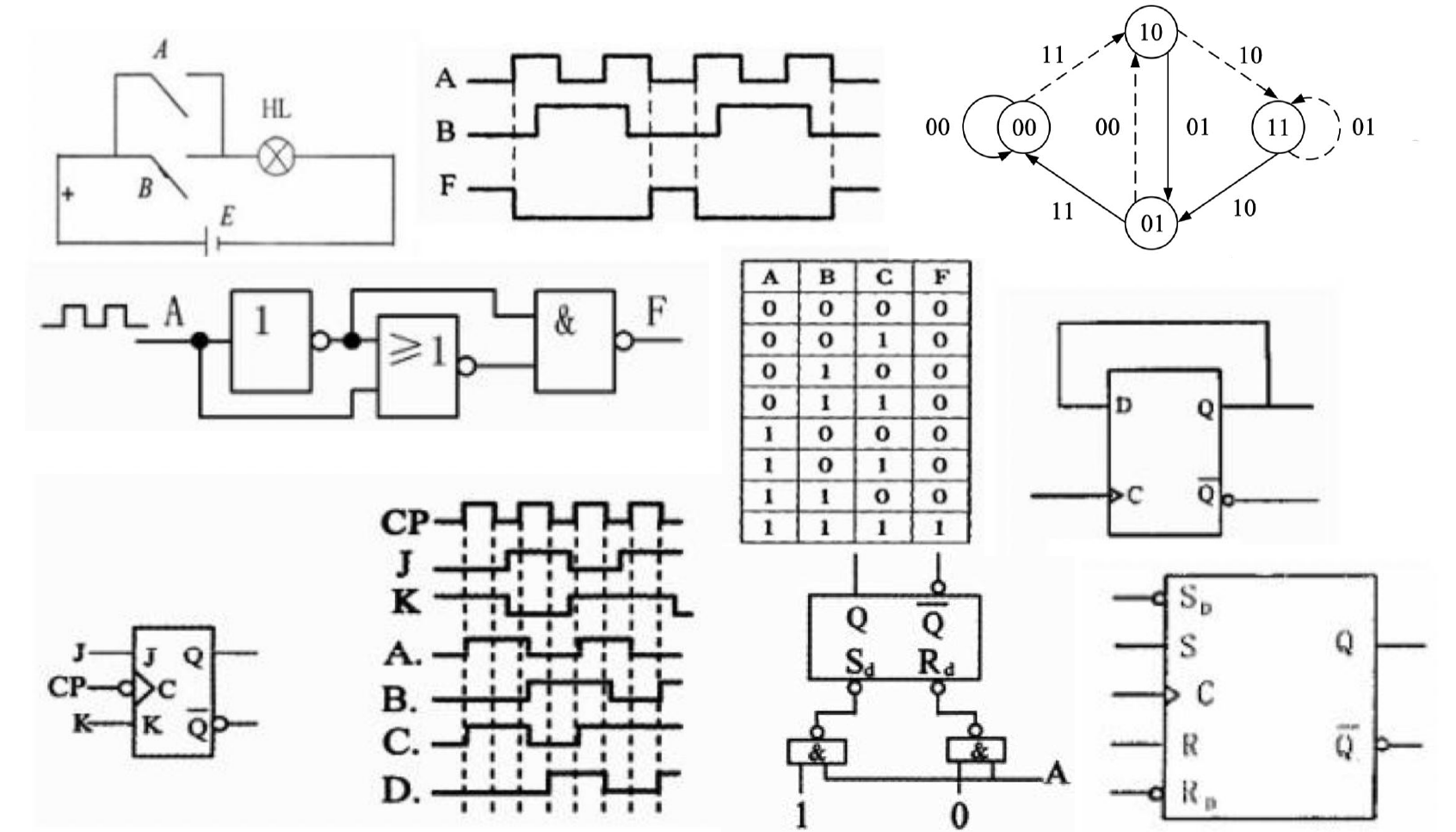}
            \vskip-5pt
    \end{minipage}
    \caption{ Examples of images from Digital Logic Circuits and
Microprocessor Design}
\label{fig:exam_1}
    \vskip -10px
\end{figure*}

\begin{figure*}[tp]
  
    \centering
    \begin{minipage}{0.99\linewidth}
            \centering
            \vskip-5pt
            \includegraphics[width=\linewidth]{     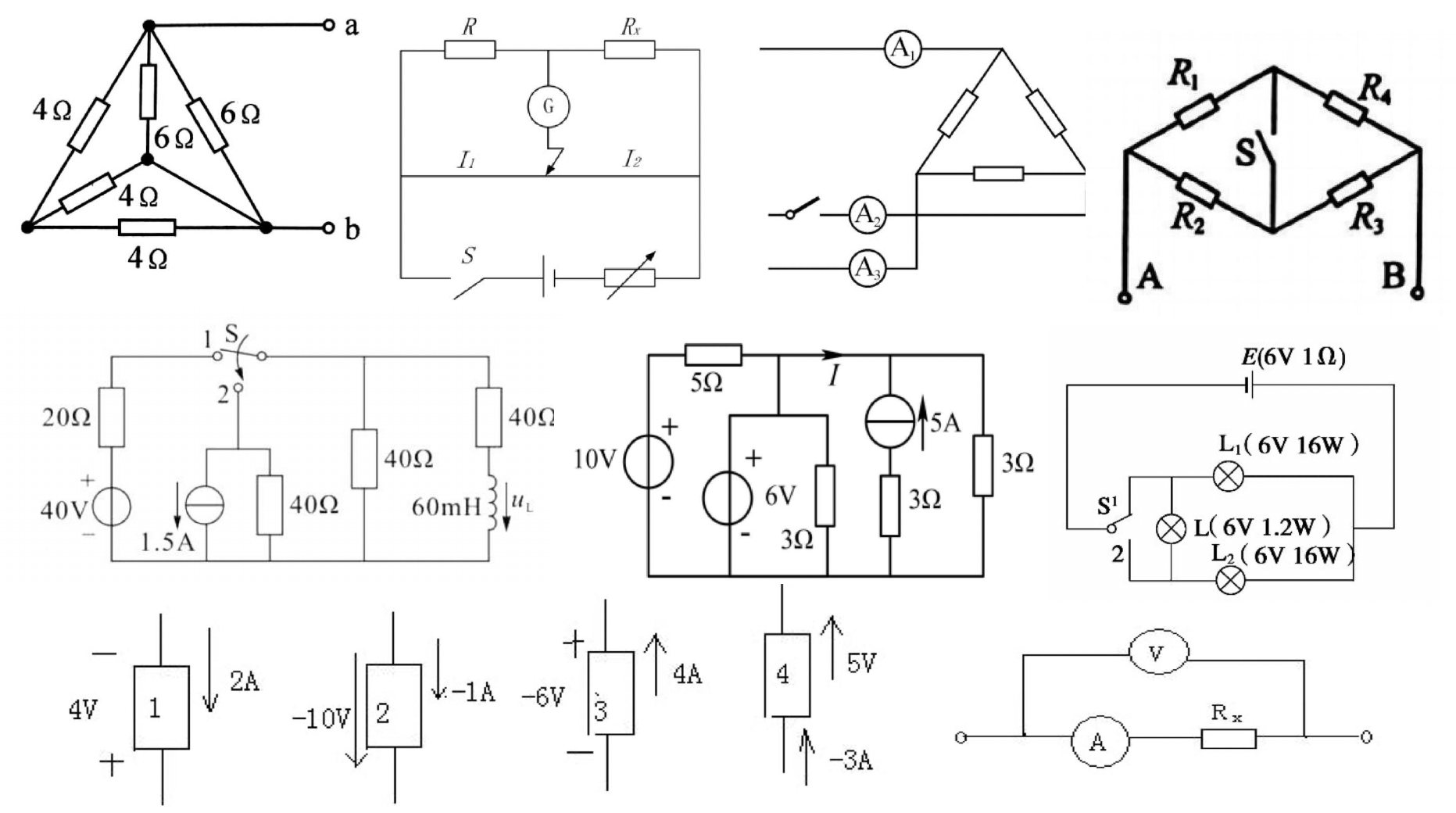}
            \vskip-5pt
    \end{minipage}
    \caption{ Examples of images from Circuit Theory and Network
Analysis}
\label{fig:exam_2}
    \vskip -10px
\end{figure*}

\begin{figure*}[t]
  
    \centering
    \begin{minipage}{0.99\linewidth}
            \centering
            \vskip-5pt
            \includegraphics[width=\linewidth]{     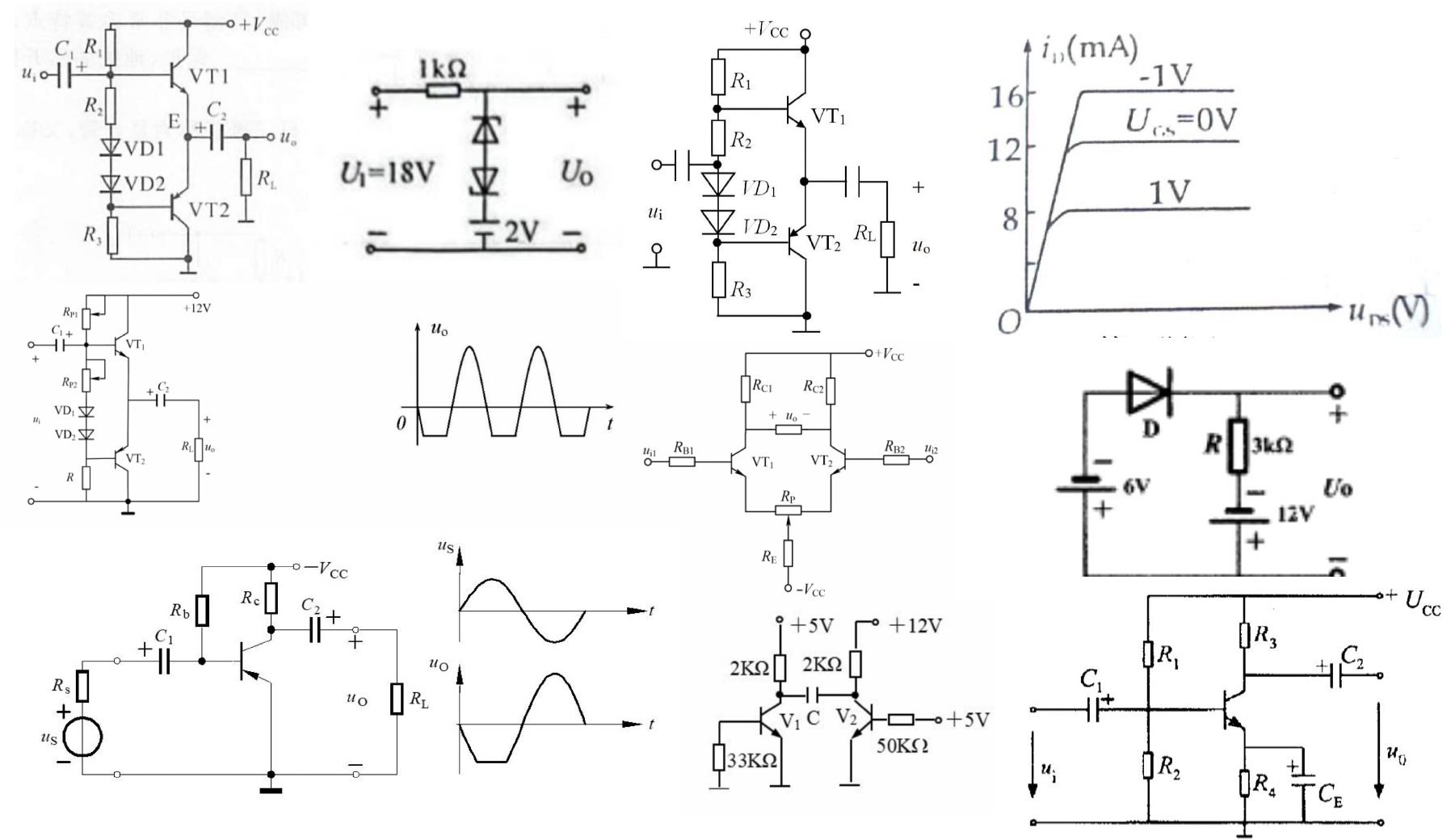}
            \vskip-5pt
    \end{minipage}
    \caption{ Examples of images from Analog Circuits}
\label{fig:exam_3}
    \vskip -10px
\end{figure*}

\begin{figure*}[t]
  
    \centering
    \begin{minipage}{0.99\linewidth}
            \centering
            \vskip-5pt
            \includegraphics[width=\linewidth]{     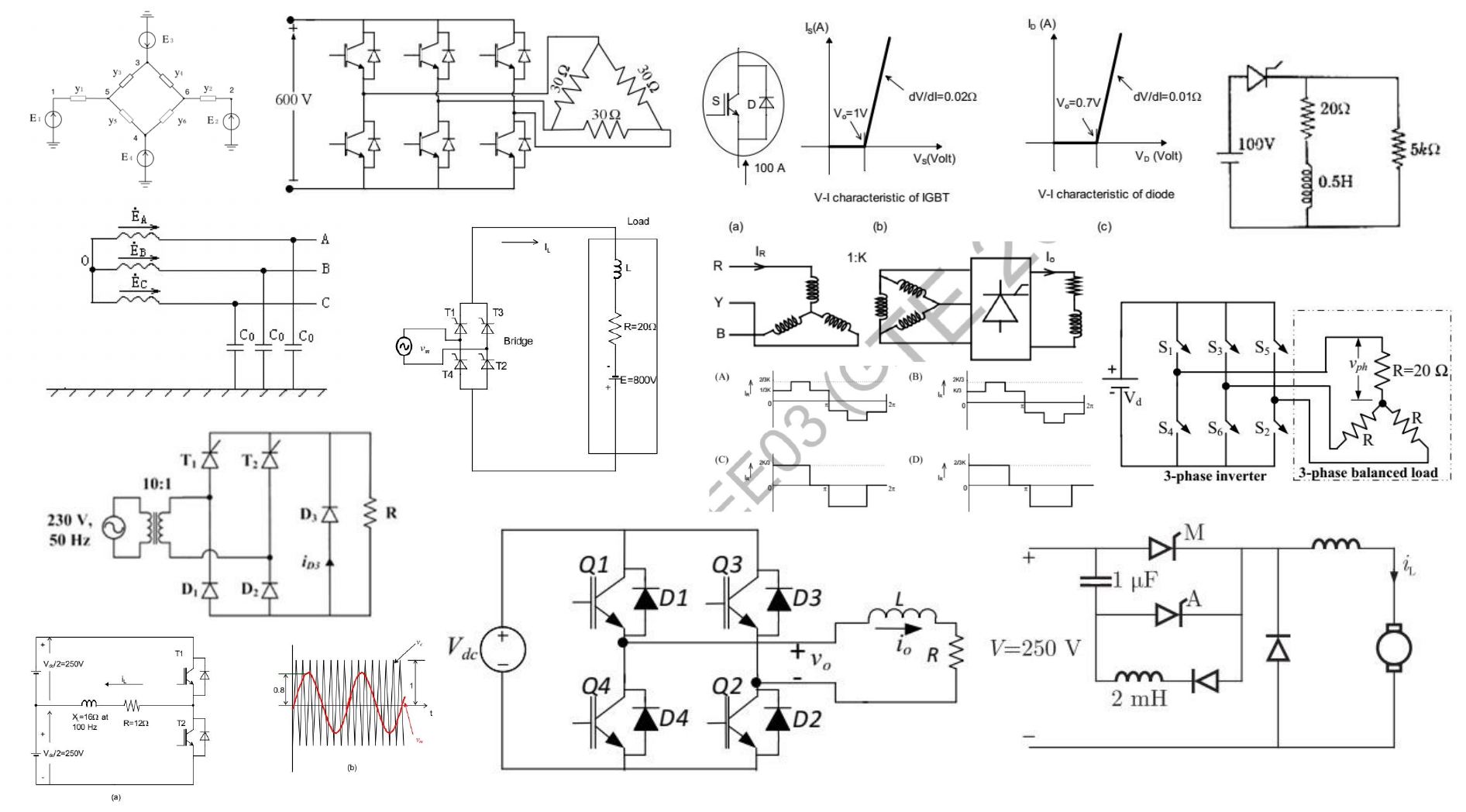}
            \vskip-5pt
    \end{minipage}
    \caption{ Examples of images from Power Electronics and Power
Systems}
\label{fig:exam_4}
    \vskip -10px
\end{figure*}

\begin{figure*}[t]
  
    \centering
    \begin{minipage}{0.99\linewidth}
            \centering
            \vskip-5pt
            \includegraphics[width=\linewidth]{     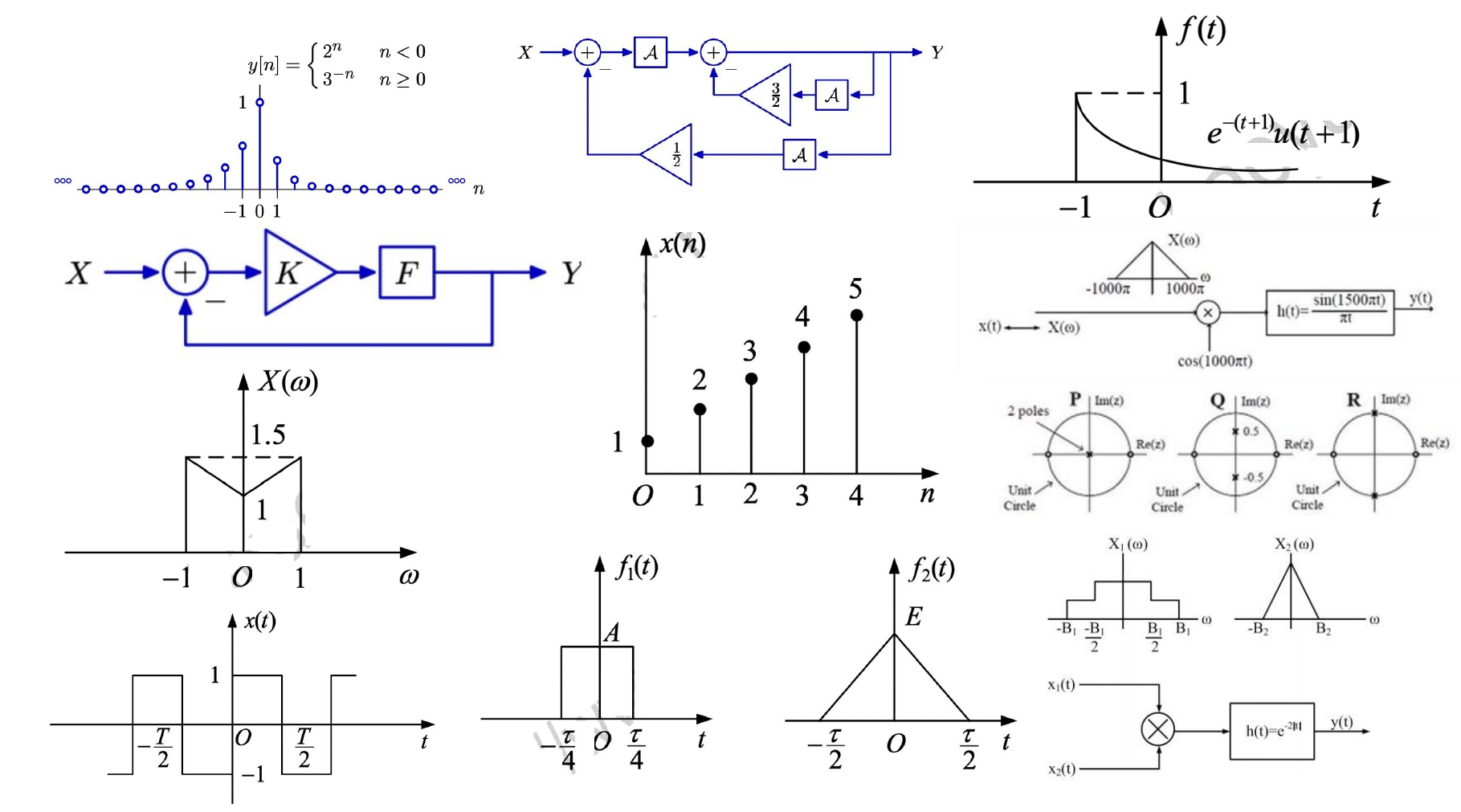}
            \vskip-5pt
    \end{minipage}
    \caption{ Examples of images from Signals and Systems}
\label{fig:exam_5}
    
\end{figure*}

\begin{figure*}[t]
  
    \centering
    \begin{minipage}{0.99\linewidth}
            \centering
            \vskip-5pt
            \includegraphics[width=\linewidth]{     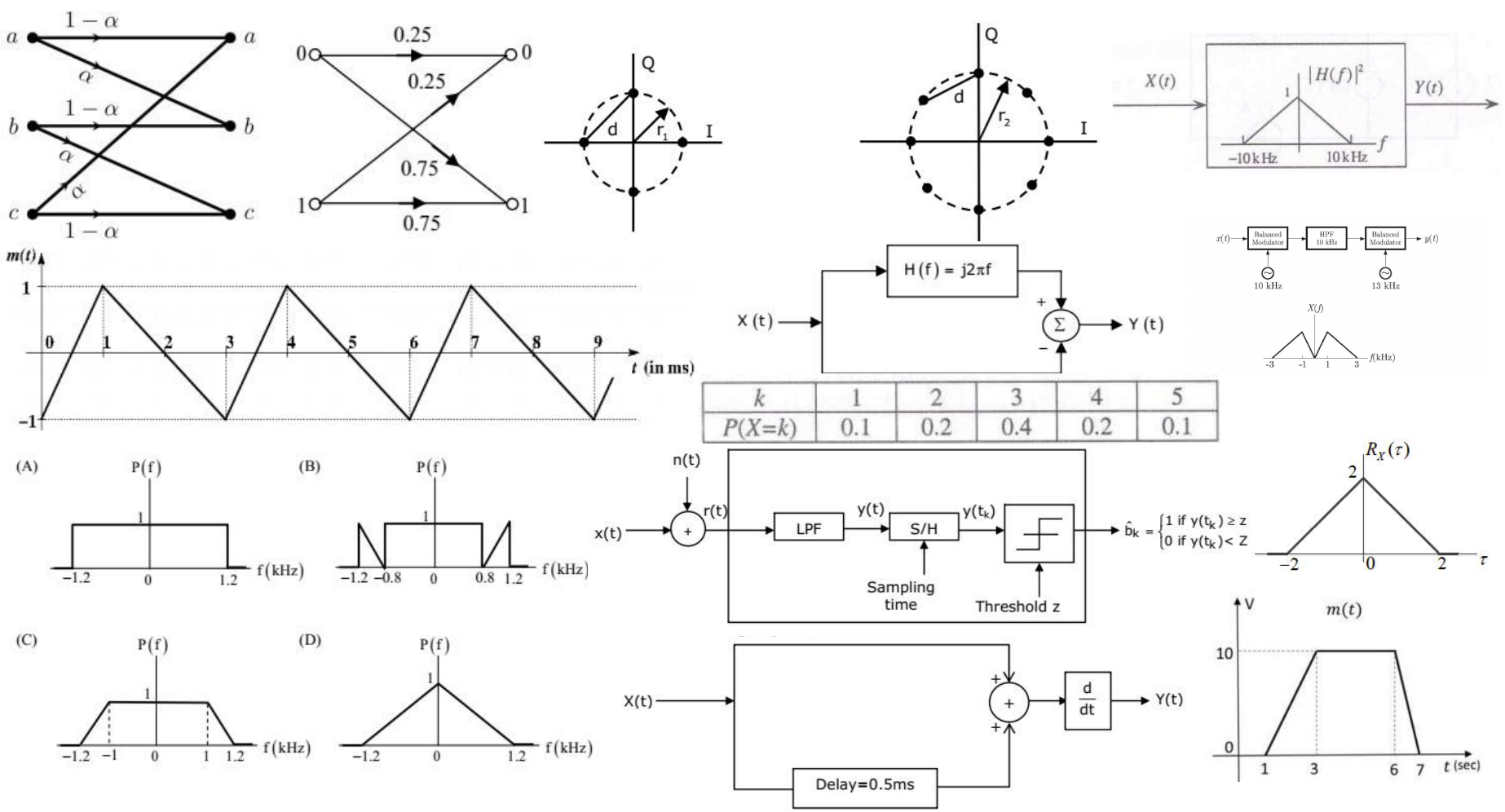}
            \vskip-5pt
    \end{minipage}
    \caption{ Examples of images from Communication Systems}
\label{fig:exam_6}
    \vskip -10px
\end{figure*}

\begin{figure*}[t]
  
    \centering
    \begin{minipage}{0.99\linewidth}
            \centering
            \vskip-5pt
            \includegraphics[width=\linewidth]{     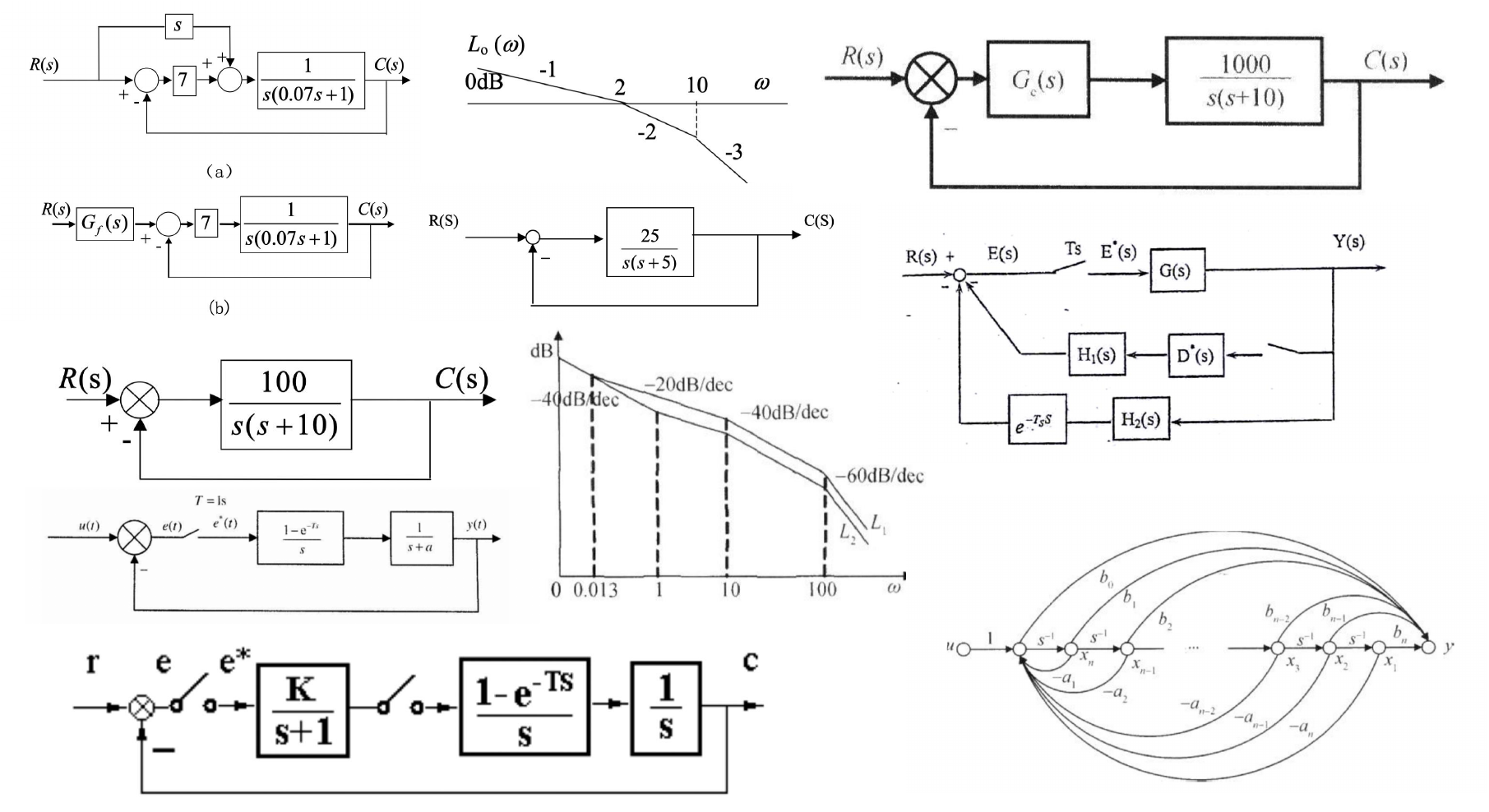}
            \vskip-5pt
    \end{minipage}
    \caption{ Examples of images from Control Systems }
\label{fig:exam_7}
    \vskip -10px
\end{figure*}

\begin{figure*}[t]
  
    \centering
    \begin{minipage}{0.99\linewidth}
            \centering
            \vskip-5pt
            \includegraphics[width=\linewidth]{     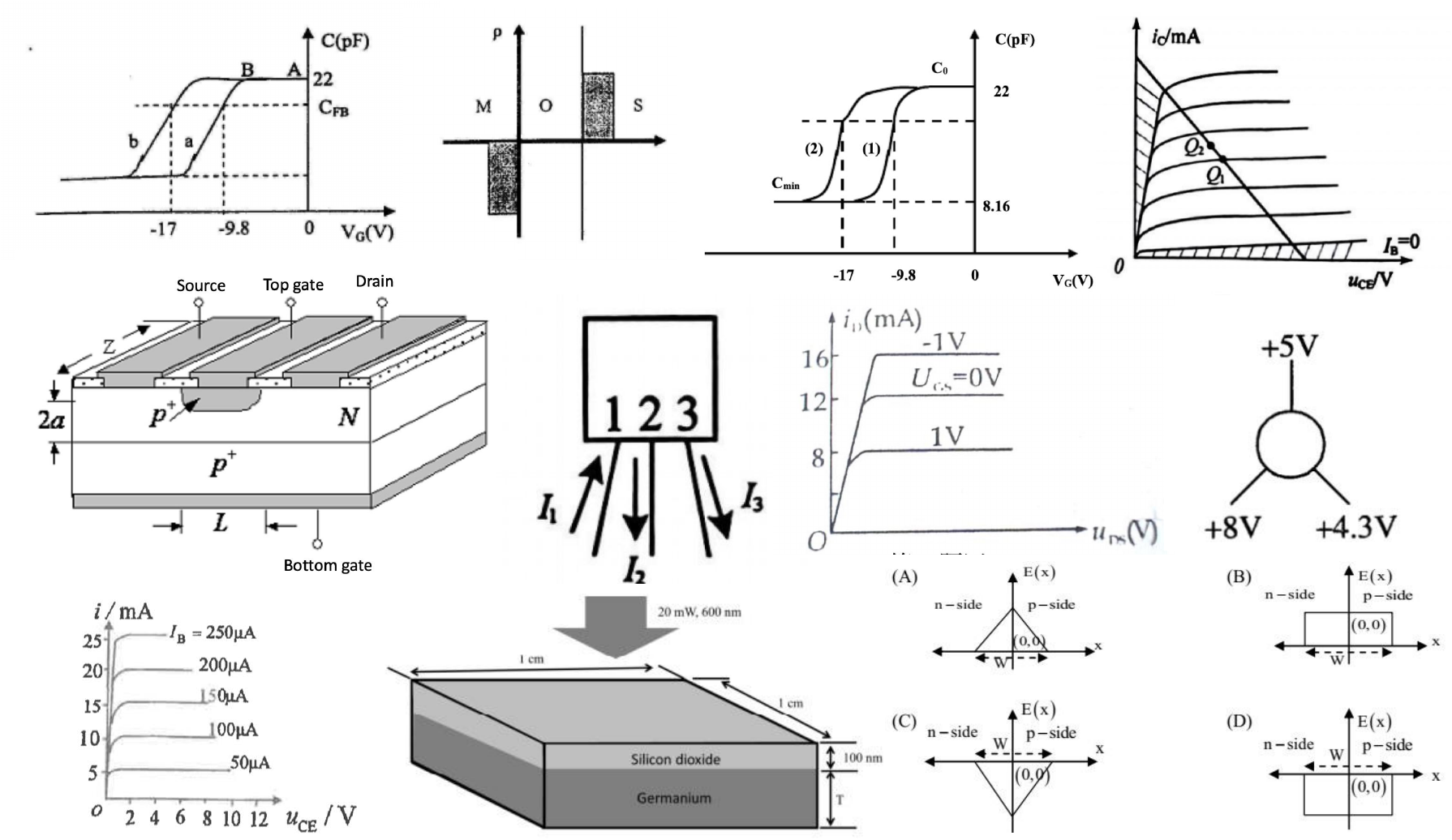}
            \vskip-5pt
    \end{minipage}
    \caption{ Examples of images from Electronic Devices }
\label{fig:exam_8}
    \vskip -10px
\end{figure*}

\begin{figure*}[t]
  
    \centering
    \begin{minipage}{0.99\linewidth}
            \centering
            \vskip-5pt
            \includegraphics[width=\linewidth]{     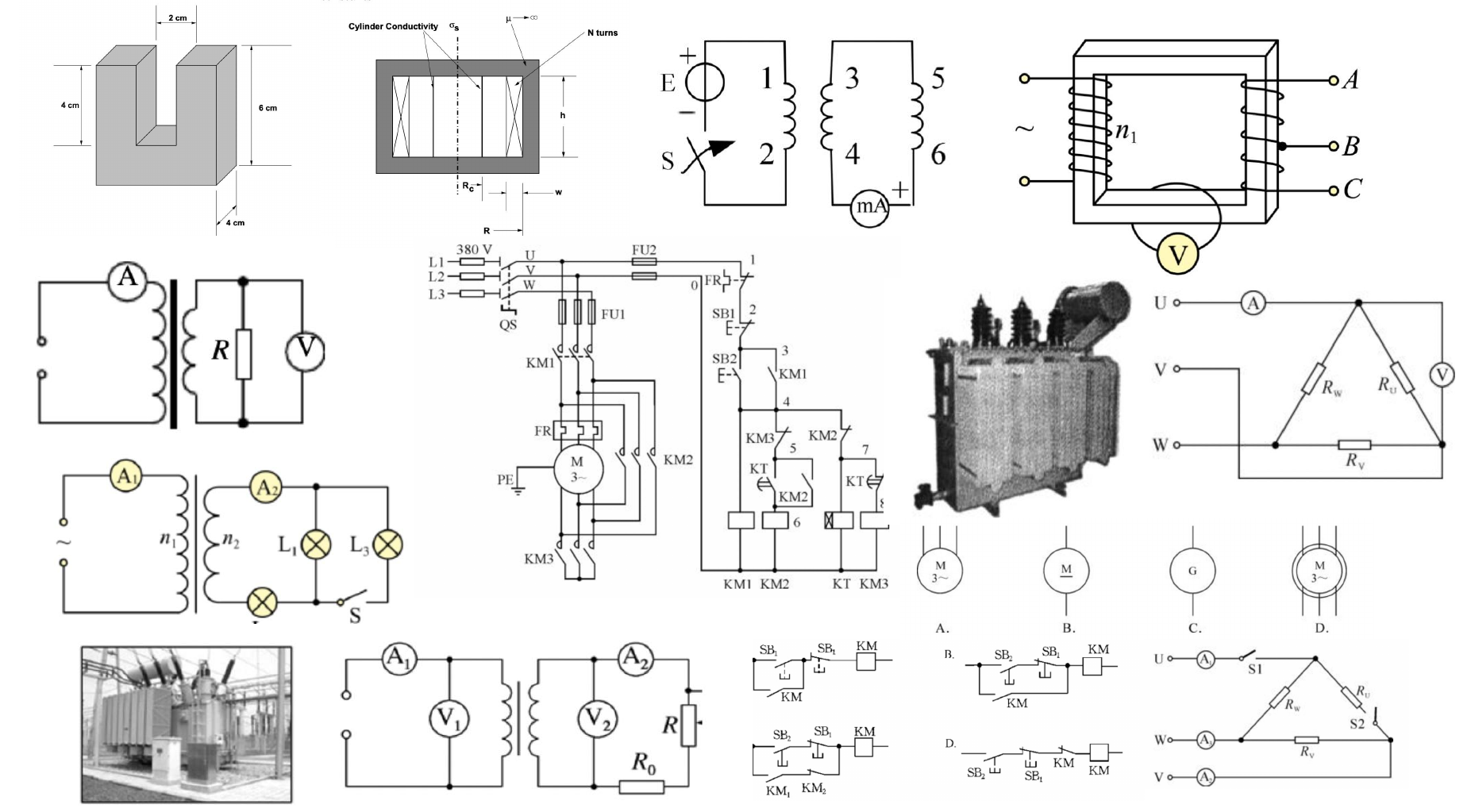}
            \vskip-5pt
    \end{minipage}
    \caption{ Examples of images from Electrical Machines }
\label{fig:exam_9}
    \vskip -10px
\end{figure*}

\begin{figure*}[t]
  
    \centering
    \begin{minipage}{0.99\linewidth}
            \centering
            \vskip-5pt
            \includegraphics[width=\linewidth]{     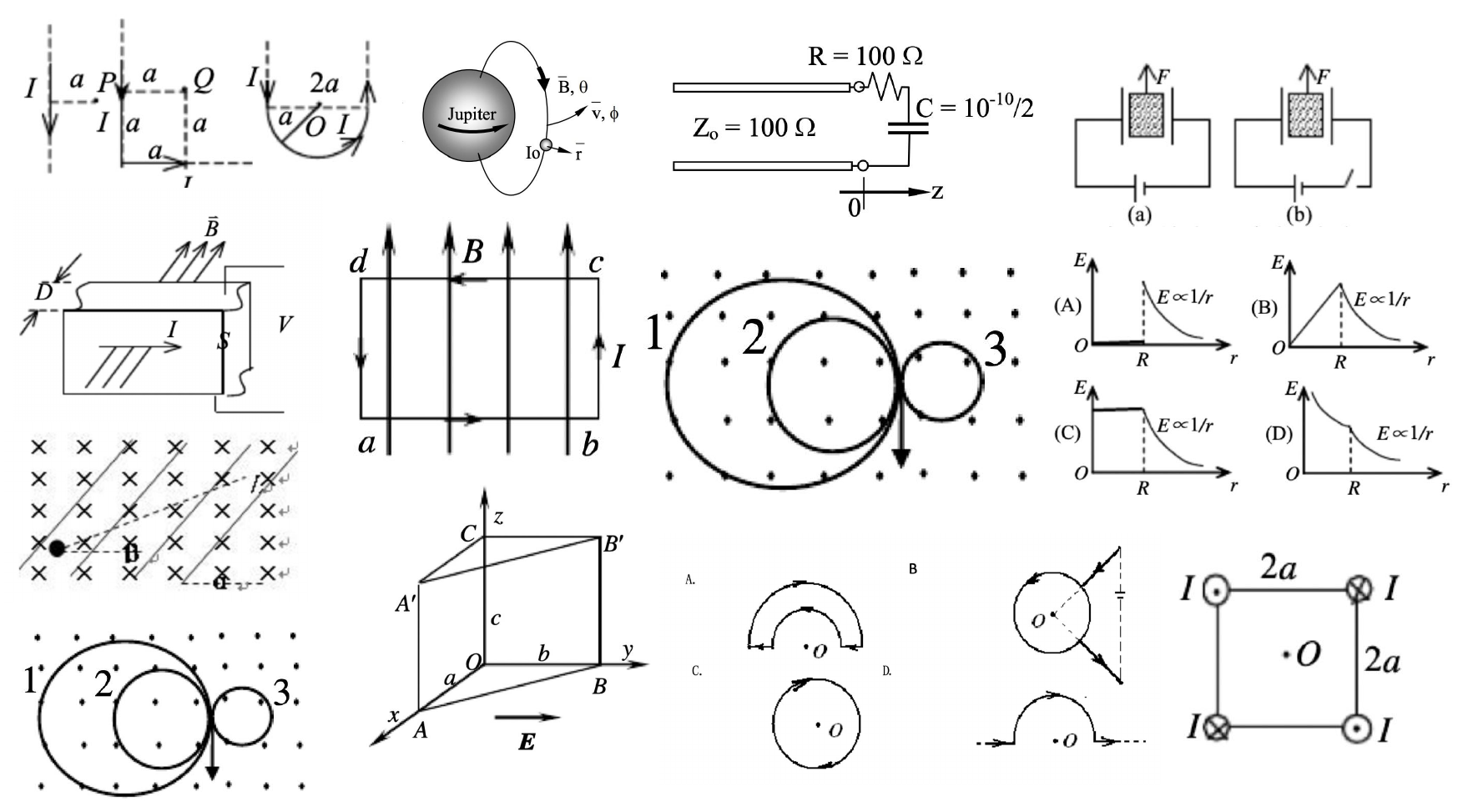}
            \vskip-5pt
    \end{minipage}
    \caption{ Examples of images from Electromagnetics }
\label{fig:exam_10}
    \vskip -10px
\end{figure*}

In this section, we provide more examples of images of EEE-Bench. \autoref{fig:exam_1} presents images from Digital Logic Circuits and
Microprocessor Design. 
\autoref{fig:exam_2} presents images from  Circuit Theory and Network
Analysis. \autoref{fig:exam_3} presents images from Analog Circuits.
\autoref{fig:exam_4} presents images from Power Electronics and Power
Systems. 
\autoref{fig:exam_5} presents images from Signals and Systems. 
\autoref{fig:exam_6} presents images from Communication Systems.
\autoref{fig:exam_7} presents images from Control Systems.
\autoref{fig:exam_8} presents images from Electronic Devices. 
\autoref{fig:exam_9} presents images from Electrical Machines.
\autoref{fig:exam_10} presents images from Electromagnetics.
EEE-Bench covers a wide range of images and thus can comprehensively evaluate the visual reasoning abilities of LMMs.

\begin{table}[th]
\centering
\caption{Description and Key Knowledge in EEE Subjects}
\label{tab:key_knowledge}
\begin{tabular}{>{\raggedright\arraybackslash}p{0.3\linewidth}p{0.65\linewidth}}
\toprule
\textbf{Subject} & \textbf{ Description and Key Knowledge} \\
\midrule
(1) Digital Logic Circuits and Microprocessor Design & 
Focuses on boolean logic, logic gates design, and the sequential circuits fundamental to digital systems. Key challenges include microprocessor architecture analysis, timing issues, and peripheral interfacing, requiring a deep understanding of logic operations and system integration. \\
\midrule
(2) Circuit Theory and Network Analysis & 
Involves detailed study of Ohm’s and Kirchhoff’s laws, network theorems, and electrical components. Emphasizes solving complex networks through techniques like mesh and nodal analysis, with problems often revolving around transient and steady-state behaviors in electrical circuits. \\
\midrule
(3) Analog Circuits & 
Covers operational amplifiers, filters, and oscillators, focusing on the design and analysis of circuits for continuous signal processing. Challenges include designing robust power amplifiers and ensuring accurate response analysis in varying operational conditions. \\
\midrule
(4) Power Electronics and Power Systems & 
Examines electrical power conversion and control using semiconductor devices. Key areas include designing efficient converters and inverters, and managing complex power systems with an emphasis on optimizing grid dynamics and energy distribution. \\
\midrule
(5) Signals and Systems & 
Deals with the representation and analysis of signals and the behavior of systems, employing tools like Fourier and Laplace transforms. Challenges include analyzing both time-domain and frequency-domain data to understand system properties and behaviors. \\
\midrule
(6) Communication Systems & 
Focuses on techniques for efficient signal transmission and reception. Involves studying modulation schemes, coding techniques, and noise analysis with problems often related to designing and optimizing communication links and networks for reliability and performance. \\
\midrule
(7) Control Systems & 
Addresses the design and analysis of dynamic systems using feedback. Involves stability analysis and controller design, with practical challenges in tuning controllers like PID for desired system responses and ensuring long-term stability and reliability. \\
\midrule
(8) Electronic Devices & 
Explores semiconductor physics and the application of devices like diodes and transistors. Problems focus on device operation under different conditions and the integration of special purpose devices such as LEDs and photodetectors in practical applications. \\
\midrule
(9) Electrical Machines & 
Studies the principles of operation and design of electric machines such as transformers and motors. Key challenges include analyzing performance, designing for efficiency, and understanding the electromechanical interactions in different operating conditions. \\
\midrule
(10) Electromagnetics & 
Explores Maxwell’s equations and their applications in electromagnetic field theory and wave propagation. Challenges include designing transmission lines, waveguides, and antennas, and ensuring electromagnetic compatibility in various applications. \\
\bottomrule
\end{tabular}
\label{subject_intro}
\end{table}

\begin{longtable}[t]{>{\raggedright\arraybackslash}p{0.3\linewidth}p{0.65\linewidth}}
\caption{Required Abilities for LMM among EEE Subjects} \label{tab:lmm_abilities} \\
\toprule
\textbf{Subject} & \textbf{Required LMM Abilities} \\
\midrule
\endfirsthead

\multicolumn{2}{c}%
{{\bfseries \tablename\ \thetable{} -- continued from previous page}} \\
\toprule
\textbf{Subject} & \textbf{Required LMM Abilities} \\
\midrule
\endhead

\midrule
\multicolumn{2}{r}{{Continued on next page}} \\
\bottomrule
\endfoot

\bottomrule
\endlastfoot

(1) Digital Logic Circuits and Microprocessor Design & Solving problems in this domain requires models to deeply understand intricate logic relationships in digital circuits and reason effectively. While these problems demand strong logical structuring and timing analysis capabilities, they do not heavily rely on computational power or advanced mathematical computations. The ability to recognize logic patterns and reason through circuit operations is essential. \\
\midrule

(2) Circuit Theory and Network Analysis & LMMs tackling problems in this subject require extensive computational processing abilities. The models must be proficient in solving complex equations and applying numerical methods to analyze and interpret circuit behaviors under varying conditions. Although less complex in terms of logic, these problems require strong numerical and computational skills to manage transient and steady-state responses. \\
\midrule

(3) Analog Circuits & For analog circuits, LMMs must handle complex visual contexts, including intricate circuit diagrams. These problems demand both sophisticated vision recognition abilities to interpret circuit structures and strong computational skills to predict continuous signal behavior. Advanced mathematical tools, such as Laplace transforms and frequency domain analysis, are required for solving circuit dynamics and performance analysis. \\
\midrule

(4) Power Electronics and Power Systems & LMMs solving power electronics and power systems problems must manage large-scale computational tasks related to energy conversion, grid dynamics, and efficiency optimization. Although the visual elements might not be as complex as other subjects, the computational effort needed for modeling and simulating power systems is significant. Strong numerical capabilities and reasoning about energy flow are crucial. \\
\midrule

(5) Signals and Systems & This domain requires LMMs to process and analyze signals in both time and frequency domains. Vision recognition skills are needed to understand the visual representations of signals, such as waveforms and spectrograms, while computational tools like Fourier and Laplace transforms must be applied to analyze system responses. The models need to handle both symbolic reasoning and numerical calculations. \\
\midrule

(6) Communication Systems & Although the visual information in communication systems tends to be straightforward, solving these problems demands considerable computational effort. LMMs must decode and simulate communication protocols, analyze noise, and optimize transmission systems. Numerical accuracy and simulation capabilities are key for optimizing communication channels and analyzing signal integrity. \\
\midrule

(7) Control Systems & LMMs addressing control system problems must excel in both reasoning and computation. Feedback systems require sophisticated reasoning to design and stabilize control loops, while mathematical modeling and simulation of dynamic systems call for strong computational abilities. Complex visual representations of system dynamics, such as block diagrams, require models to recognize system structures and respond accordingly. \\
\midrule

(8) Electronic Devices & In this subject, LMMs must understand the physical principles of semiconductor devices, but the problems primarily involve numerical computation to predict device behaviors. Visual contexts, such as device schematics, are generally simple, but the computational tasks can be intensive, requiring models to handle material properties and electrical characteristics with high precision. \\
\midrule

(9) Electrical Machines & Problems in this domain involve electromechanical energy conversion, and LMMs must reason about mechanical and electrical interactions. The visual complexity is moderate, but models must be proficient in numerical simulations to predict machine performance under different conditions. Computational tools are essential for solving mechanical equations and analyzing energy transfer processes. \\
\midrule

(10) Electromagnetics & LMMs need to possess strong reasoning skills to interpret complex visual contexts like field distributions and wave propagation diagrams. Additionally, these problems demand high computational power to solve Maxwell's equations and simulate electromagnetic fields. Numerical accuracy and proficiency in solving partial differential equations are key to successfully addressing challenges in this domain. \\
\end{longtable}
\section{Detailed Introduction to Ten Subjects}
\label{appendix:detail_intro}
To better help readers understand the key points of these ten subjects, we provide a detailed introduction for each in this section. Further information on each subject is available in \autoref{subject_intro}.

\section{Required Ability for LMMs to Solve EEE Problems}
\label{appendix:lmm_abilities}
\begin{figure}[h]
\caption{Examples of problems in EEE-Bench.}
\hrule
\centering
\begin{minipage}[c]{0.4\textwidth}
\includegraphics[width=\textwidth]{      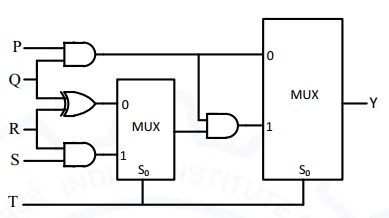}
\end{minipage}
\hfill
\begin{minipage}[c]{0.98\textwidth}
\textbf{Subject:} Digital Logic Circuits and Microprocessor Design\\
\textbf{Question:} The propagation delays of the XOR gate, AND gate and multiplexer (MUX) in the circuit shown in the figure are $4 \mathrm{~ns}, 2 \mathrm{~ns}$ and 1 ns , respectively.

If all the inputs P, Q, R, S and T are applied simultaneously and held constant, the maximum propagation delay of the circuit is ( )\\
(A) 3 ns
(B) 5 ns
(C) 6 ns
(D) 7 ns
\\ 
\textbf{Solution:} Case -1 : when T=0 \\
\text{Propagation delay} = $t_{AND1} + t_{MUX2} = 2 + 1 = 3ns$ \\
Case -1 : when T=1 \\
\text{Propagation delay} $= t_{AND2} + t_{MUX1} t_{AND3} + t_{MUX2} = 2 + 1 + 2 + 1 = 6ns.$
\\
\textbf{Answer:} C
\vspace{0.4em}
\end{minipage}
\hrule
\begin{minipage}[c]{0.4\textwidth}
\includegraphics[width=\textwidth]{      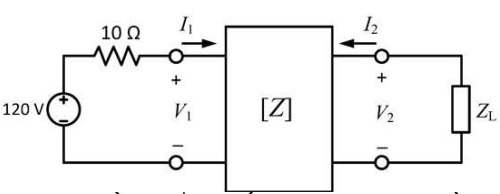}
\end{minipage}
\hfill
\begin{minipage}[c]{0.98\textwidth}
\textbf{Subject:} Circuit Theory and Network
Analysis\\
\textbf{Question:} In the given circuit shown in figure, the two-port network has the impedance matrix [Z] = 
$\begin{bmatrix}
40 & 60 \\
60 & 120
\end{bmatrix}$.
The value of $Z_L$ for which maximum power is transferred to the load is ( ) $\Omega$.\\
A. 48 B. 38 C. 40 D. 54
\\ 
\textbf{Solution:}From maximum power transfer theorem
$$Z_L = Z_{th}, \ Z_{th} = Z_{22} - \frac{Z_{12} \times Z_{21}}{R_s + Z_{11}}$$

For given data,
$$Z_{th} = 120 - \frac{60 \times 60}{10 + 40} = 48 \Omega, \ Z_L = 48 \Omega.$$
\textbf{Answer:} A
\vspace{0.4em}
\end{minipage}
\hrule
\begin{minipage}[c]{0.4\textwidth}
\includegraphics[width=\textwidth]{      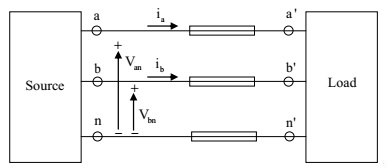}
\end{minipage}
\hfill
\begin{minipage}[c]{0.98\textwidth}
\textbf{Subject:} Power Electronics and Power
Systems\\
\textbf{Question:} A source is supplying a load through a 2-phase, 3-wire transmission system as shown in the figure. The instantaneous voltage and current in phase-a are 
$$V_{an} = 220 \sin(100\pi t) \, V \quad \text{and} \quad i_a = 10 \sin(100\pi t) \, A,$$ 
respectively. Similarly, for phase-b the instantaneous voltage and current are 
$$V_{bn} = 220 \cos(100\pi t) \, V \quad \text{and} \quad i_b = 10 \cos(100\pi t) \, A,$$ 
respectively.
The total instantaneous power flowing from the source to the load is ( ).\\
(A): 2200 W (B): $2200 \sin^2(100\pi t)$ W 
(C): 440W 
(D): $2200 \sin(100\pi t) \cos(100\pi t)  W$
\\ 
\textbf{Solution:} $V_{an} = 220 \sin(100\pi t)  V 
, \ i_a = 10 \sin(100\pi t) A $,

$V_{bn} = 220 \cos(100\pi t)  V , \ 
i_b = 10 \cos(100\pi t)  A $

$p = V_{an} i_a + V_{bn} i_b 
= 2200  W$
\\
\textbf{Answer:} A
\vspace{0.4em}
\end{minipage}
\hrule
\vspace{0.4em}
\caption{Examples of problems in EEE-Bench.}
\label{exam_fig5}

\end{figure}

\begin{figure}[h]
\hrule
\centering
\begin{minipage}[c]{0.4\textwidth}
\includegraphics[width=\textwidth]{      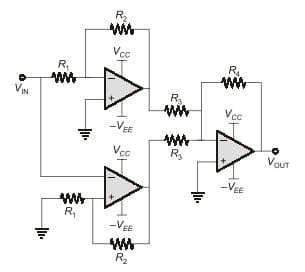}
\end{minipage}
\hfill
\begin{minipage}[c]{0.98\textwidth}
\textbf{Subject:} Analog Circuits \\
\textbf{Question:} The $\frac{V_{OUT}}{V_{IN}}$ of the circuit shown in figure is ( ).
$$
\text{A:} \  -\frac{R_4}{R_3}  \quad
\text{B:} \  \frac{R_4}{R_3}  \quad
\text{C:} \  1 + \frac{R_4}{R_3}  \quad
\text{D:} \ 1 - \frac{R_4}{R_3}
$$
\\ 
\textbf{Solution:} Here, $A_1$ is an inverting amplifier and $A_2$ is a non-inverting amplifier.

$$V_{01} = \frac{-R_2}{R_1} V_{in}, \ V_{02} = \left(1 + \frac{R_2}{R_1}\right) V_{in}$$

Also, $A_3$ is an inverting summing amplifier,

$$V_{out} = \frac{-R_4}{R_3} V_{01} + \frac{-R_4}{R_3} V_{02} = \frac{-R_4}{R_3} \left[\frac{R_2}{R_1} V_{in} + \left(1 + \frac{R_2}{R_1}\right) V_{in}\right]$$

$$V_{out} = \frac{-R_4}{R_3} V_{in}$$

Gain, 
$$\frac{V_{out}}{V_{in}} = \frac{-R_4}{R_3}$$
\\
\textbf{Answer:} A
\vspace{0.4em}
\end{minipage}
\hrule

\begin{minipage}[c]{0.4\textwidth}
\includegraphics[width=\textwidth]{      }
\end{minipage}
\hfill
\begin{minipage}[c]{0.98\textwidth}
\textbf{Subject:} Signals and Systems  \\
\textbf{Question:} Consider a six-point decimation-in-time Fast Fourier Transform (FFT) algorithm, for which the signal-flow graph corresponding to $X[1]$ is shown in the figure. Let $W_6 = \exp\left(-\frac{j 2 \pi}{6}\right)$. In the figure, what should be the values of the coefficients $a_1, a_2, a_3$ in terms of $W_6$ so that $X[1]$ is obtained correctly?

 $A: \ a_1 = -1, a_2 = W_6, a_3 = W_6^2 \quad B: \ a_1 = 1, a_2 = W_6^2, a_3 = W_6 $\\
$C: \ a_1 = 1, a_2 = W_6, a_3 = W_6^2 \quad D: \ a_1 = -1, a_2 = W_6^2, a_3 = W_6
$
\\ 
\textbf{Solution:} $X(k) = \sum_{n=0}^{N-1} x(n) e^{-j \frac{2 \pi}{N} kn}.$ \\
$X(1) = \sum_{n=0}^{5} x(n) W_6^n= x(0) + x(1) W_6 + x(2) W_6^2 + x(3) W_6^3 + x(4) W_6^4 + x(5) W_6^5 \dots (i)$

From the given flow graph, we have
$X(k) = [x(0) - x(3)] a_1 + [x(1) - x(4)] a_2 + [x(2) - x(5)] a_3 \dots (ii).$

By comparing equations (i) and (ii), we get 
$a_1 = 1, \quad a_2 = W_6, \quad a_3 = W_6^2.$
\\
\textbf{Answer:} C
\vspace{0.4em}
\end{minipage}
\hrule
\vspace{0.4em}
\caption{Examples of problems in EEE-Bench.}
\label{exam_fig4}
\end{figure}

\begin{figure}[h]
\hrule
\centering
\begin{minipage}[c]{0.4\textwidth}
\includegraphics[width=\textwidth]{      }
\end{minipage}
\hfill
\begin{minipage}[c]{0.98\textwidth}
\textbf{Subject:} Communication Systems  \\
\textbf{Question:} The frequency response $H(f)$ of a linear time-invariant system has magnitude as shown in the figure.

Statement I: The system is necessarily a pure delay system for inputs which are bandlimited to $-\alpha \leq f \leq \alpha$.

Statement II: For any wide-sense stationary input process with power spectral density $S_X(f)$, the output power spectral density $S_Y(f)$ obeys $S_Y(f) = S_X(f)$ for $-\alpha \leq f \leq \alpha$.
Which one of the following combinations is true?

(A): Statement I is correct, Statement II is correct (B): Statement I is correct, Statement II is incorrect \\
(C): Statement I is incorrect, Statement II is correct 
(D): Statement I is incorrect, Statement II is incorrect
\\ 
\textbf{Solution:} For the system to be a delay system
$$y(t) = x(t - t_d),\
y(F) = e^{-j\omega t_d} \times F$$
$$\Rightarrow H(F) = \frac{Y(F)}{X(F)} = e^{-j\omega t_d}$$
Here given system is constant, hence this is not a delay system, therefore statement I is incorrect.

$$S_y(f) = S_x(f) |H(f)|^2$$
and $|H(f)| = 1$(given).
Hence, 
$$S_y(f) = S_x(f) \, \text{for} \, -\alpha \leq f \leq \alpha$$
Statement II is correct.
\\
\textbf{Answer:} C
\vspace{0.4em}
\end{minipage}
\hrule

\begin{minipage}[c]{0.4\textwidth}
\includegraphics[width=\textwidth]{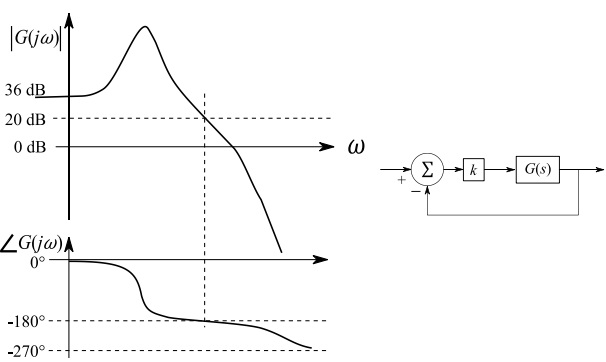}
\end{minipage}
\hfill
\begin{minipage}[c]{0.98\textwidth}
\textbf{Subject:} Control Systems  \\
\textbf{Question:} The figure below shows the Bode magnitude and phase plots of a stable transfer function $G(s) = \frac{n_0}{s^2 + d_2 s^2 + d_1 s + d_0}$.

Consider the negative unity feedback configuration with gain $k$ in the feedforward path. The closed loop is stable for $k < k_o$. The maximum value of $k_o$ is( ).\\
A. 0.1 B. 0.2 C. 0.3 D. 0.4
\\ 
\textbf{Solution:} For $G(s)$
$$M_{dB} (\omega_{pc}) = 20 dB$$

When cascaded with $k$,
$$GM_{dB} = -20 dB - 20 \log_{10}(k) > 0 dB$$

\[
20 + 20 \log_{10}(k) < 0
\]
\[
20 \log_{10}(k) < -20
\]
\[
k < 10^{-1} = 0.10
\]
So, $k_0 = 0.10$
\\
\textbf{Answer:} A
\vspace{0.4em}
\end{minipage}
\hrule
\vspace{0.4em}
\caption{Examples of problems in EEE-Bench.}
\label{exam_fig3}
\end{figure}

\begin{figure}[h]
\hrule
\centering
\begin{minipage}[c]{0.4\textwidth}
\includegraphics[width=\textwidth]{      }
\end{minipage}
\hfill
\begin{minipage}[c]{0.98\textwidth}
\textbf{Subject:} Electrical Machines  \\
\textbf{Question:} Windings 'A', 'B' and 'C' have 20 turns each and are wound on the same iron core as shown, along with winding 'X' which has 2 turns. The figure shows the sense (clockwise/anti-clockwise) of each of the windings only and does not reflect the exact number of turns. If windings 'A', 'B' and 'C' are supplied with balanced 3-phase voltages at 50 Hz and there is no core saturation, the no-load RMS voltage (in V, rounded off to 2 decimal places) across winding 'X' is ( ).\\
A. 36 B. 46 C. 12 D. 58
\\ 
\textbf{Solution:} $V_X = \frac{2}{20} \left(230 \angle 0^\circ - 230 \angle 120^\circ - 230 \angle -120^\circ \right) = 46 \angle 0^\circ \, \text{V}$
\\
\textbf{Answer:} B
\vspace{0.4em}
\end{minipage}
\hrule

\begin{minipage}[c]{0.4\textwidth}
\includegraphics[width=\textwidth]{      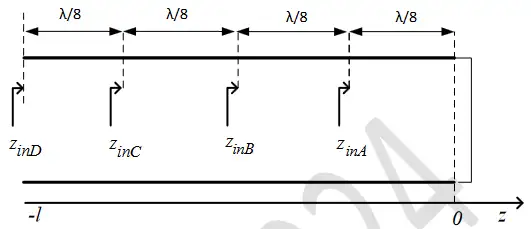}
\end{minipage}
\hfill
\begin{minipage}[c]{0.98\textwidth}
\textbf{Subject:} Electromagnetics  \\
\textbf{Question:} Consider a lossless transmission line terminated with a short circuit as shown in the figure. As one moves towards the generator from the load, the normalized impedances $z_{\text{in} \, A}, z_{\text{in} \, B}, z_{\text{in} \, C}, \text{ and } z_{\text{in} \, D} \text{ (indicated in the figure) are }$ ( ).
\begin{align*}
\text{A:} & \quad z_{\text{in} \, A} = +0.4j \, \Omega, z_{\text{in} \, B} = \infty, z_{\text{in} \, C} = 0, z_{\text{in} \, D} = -0.4j \, \Omega \\
\text{B:} & \quad z_{\text{in} \, A} = \infty, z_{\text{in} \, B} = +0.4j \, \Omega, z_{\text{in} \, C} = 0, z_{\text{in} \, D} = +0.4j \, \Omega \\
\text{C:} & \quad z_{\text{in} \, A} = -1j \, \Omega, z_{\text{in} \, B} = 0, z_{\text{in} \, C} = +1j \, \Omega, z_{\text{in} \, D} = \infty \\
\text{D:} & \quad z_{\text{in} \, A} = +1j \, \Omega, z_{\text{in} \, B} = \infty, z_{\text{in} \, C} = -1j \, \Omega, z_{\text{in} \, D} = 0 \quad 
\end{align*}
\\ 
\textbf{Solution:} $ Z_{s/c} = j Z_0 \tan \beta l$. \\
$\Rightarrow \text{Normalized impedance, } \bar{Z}_{S/C} = \frac{Z_{S/C}}{Z_0} = j \tan \beta l$ \\

$\bar{Z}_{in \, A} = j \tan \left( \frac{2\pi}{\lambda} \cdot \frac{\lambda}{8} \right) = j \tan\left( \frac{\pi}{4} \right) = j1 \, \Omega$ \\

$\bar{Z}_{in \, B} = j \tan \left( \frac{2\pi}{\lambda} \cdot \frac{\lambda}{4} \right) = \infty $\\

$\bar{Z}_{in \, C} = j \tan \left( \frac{2\pi}{\lambda} \cdot \frac{3\lambda}{8} \right) = -j1 \, \Omega$ \\

$\bar{Z}_{in \, D} = j \tan \left( \frac{2\pi}{\lambda} \cdot \frac{\lambda}{2} \right) = 0$
\\
\textbf{Answer:} D
\vspace{0.4em}
\end{minipage}
\hrule
\vspace{0.4em}
\caption{Examples of problems in EEE-Bench.}
\label{exam_fig2}
\end{figure}

\begin{figure}[th]
\hrule
\centering
\begin{minipage}[c]{0.4\textwidth}
\includegraphics[width=\textwidth]{      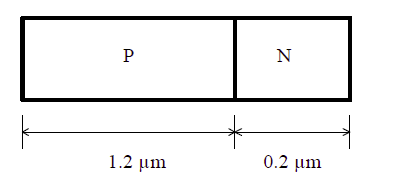}
\end{minipage}
\hfill
\begin{minipage}[c]{0.98\textwidth}
\textbf{Subject:} Electronic Devices  \\
\textbf{Question:} A silicon P-N junction is shown in the figure. The doping in the P region is $5 \times 10^{16} \, cm^{-3}$ and the doping in the N region is $10 \times 10^{16} \, cm^{-3}$. The parameters given are:

\begin{align*}
\text{Built-in voltage } (\Phi_{bi}) & = 0.8 \, V \\
\text{Electron charge } (q) & = 1.6 \times 10^{-19} \, C \\
\text{Vacuum permittivity } (\varepsilon_0) & = 8.85 \times 10^{-12} \, F/m
\end{align*}

Relative permittivity of silicon $(\varepsilon_{si}) = 12$

The magnitude of reverse bias voltage that would completely deplete one of the two regions (P or N) prior to the other (rounded off to one decimal place) is ( ) V (round to one decimal).
\\ 
\textbf{Solution:} Given that $N_A = 5 \times 10^{16} \, \text{cm}^{-3}; \quad N_D = 10 \times 10^{16} \, \text{cm}^{-3}$ \\
Built-in potential: $\Phi_{bi} = 0.8 \, \text{V}$ \\
Electron charge: $q = 1.6 \times 10^{-19} \, \text{C}$ \\
$\text{Vacuum permittivity, } \varepsilon_0 = 8.85 \times 10^{-12} \, \text{F/m} = 8.85 \times 10^{-14} \, \text{F/cm}$ \\
$\text{Relative permittivity of silicon, } \varepsilon_{si} = 12 $\\

$\Rightarrow \text{Doping on both sides is comparable, so the smaller region would deplete first.}$ \\
$\text{So, depletion region width on the N-side } = x_n = 0.2 \, \mu m$ \\
$\Rightarrow x_n = 0.2 \times 10^{-4} \, cm$ \\

$x_n = \sqrt{\frac{2 \varepsilon_{si}}{q} \left( \frac{N_A}{N_D} \right) \left( \frac{1}{N_A + N_D} \right) (\phi_{bj} + V_R)}$ \\

$\text{where, } V_R \rightarrow \text{Magnitude of reverse bias potential}$ \\
$\Rightarrow 0.2 \times 10^{-4} = \sqrt{\frac{2 \times 12 \times 8.85 \times 10^{-14}}{1.6 \times 10^{-19}} \cdot \frac{5 \times 10^{16}}{10 \times 10^{16}} \cdot \frac{1}{(15 \times 10^{16})} (\phi_{bj} + V_R)}$ \\

$\Rightarrow \phi_{bj} + V_R = 9.039$ \\
$\Rightarrow V_R = 9.039 - 0.8 $\\
$\Rightarrow V_R = 8.239 \, V \approx 8.2 \, V$
\\
\textbf{Answer:} 8.2
\vspace{0.4em}
\end{minipage}
\hrule
\vspace{0.4em}
\caption{Examples of problems in EEE-Bench.}
\label{exam_fig1}
\end{figure}

In this section, we discuss the characteristics of each subject’s problems and the required
ability for LMMs to solve them. The discussion can be found in \autoref{tab:lmm_abilities}. Some examples of problems and detailed solutions in EEE-Bench are shown in \autoref{exam_fig5},\autoref{exam_fig4},\autoref{exam_fig3},\autoref{exam_fig2},\autoref{exam_fig1}.

\section{More examples of error analysis}
\label{more_error}
\begin{figure}[th]
\hrule
\centering
\begin{minipage}[c]{0.2\textwidth}
\includegraphics[width=\textwidth]{    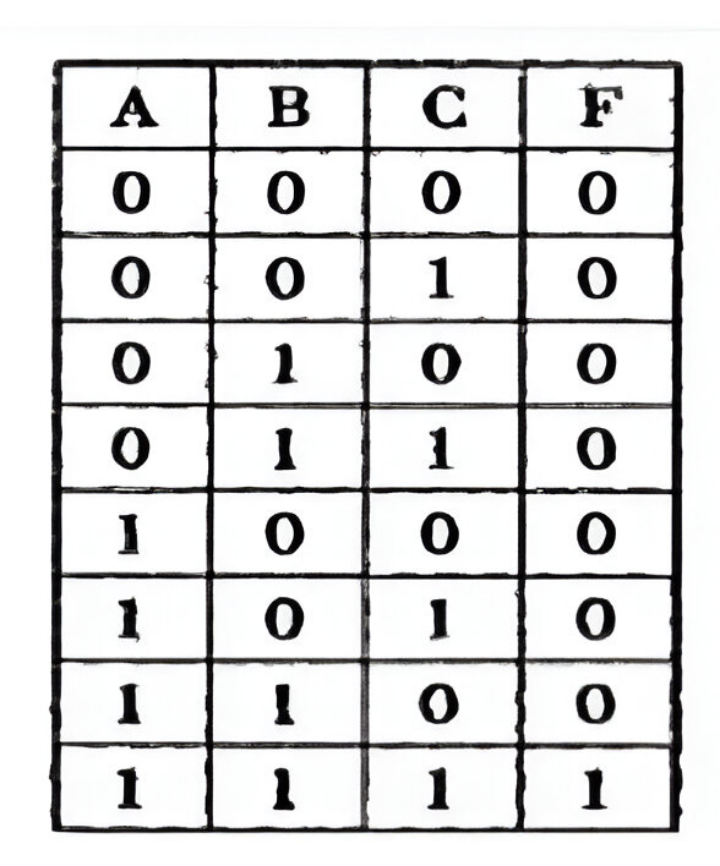}
\end{minipage}
\hfill
\begin{minipage}[c]{0.98\textwidth}
\small

\textbf{Question: }
The logical expression corresponding to the logic state table shown in figure is (  )\\
A. $F = A + B + C$ \\
B. $F = A\cdot B\cdot C$ \\
C. $F = A + B + \overline{C}$ \\
D. $F = A\cdot B\cdot \overline{C}$ \\

\textbf{Answer: }B. \\
\textbf{Response from GPT-4o: } To determine the logical expression, we need to analyze the truth table and identify when the output \( F \) is 1. {\color{red}From the table: F = 1 when  A=1,B = 1,C=0.} This corresponds to the logical expression$ F = A \cdot B \cdot \overline{C}$.{\color{red}Therefore, the correct option is: 
D. $F = A\cdot B\cdot \overline{C}$}

\vspace{0.4em}
\end{minipage}
\hrule

\begin{minipage}[c]{0.6\textwidth}
\includegraphics[width=\textwidth]{    }
\end{minipage}
\hfill
\begin{minipage}[c]{0.98\textwidth}
\small

\textbf{Question: }
Among the four circuits shown in the figure, the circuit that satisfies the equation \( U = -IR + E \) is ( ).\\
A. circuit A\\
B. circuit B\\
C. circuit C\\
D. circuit D\\

\textbf{Answer: }C.\\
\textbf{Response from GPT-4o: } The equation \( U = -IR + E \) suggests that the potential difference \( U \) is equal to the electromotive force \( E \) minus the voltage drop across the resistor \( IR \). This implies that the current \( I \) should flow from the positive terminal of the battery through the resistor.

{\color{red}In circuit B, the current \( I \) flows from the positive terminal of the battery through the resistor, which matches the equation \( U = -IR + E \).}

Therefore, the correct answer is:

{\color{red}B. circuit B}

\vspace{0.4em}
\end{minipage}
\hrule

\begin{minipage}[c]{0.5\textwidth}
\includegraphics[width=\textwidth]{    }
\end{minipage}
\hfill
\begin{minipage}[c]{0.98\textwidth}
\small

\textbf{Question: }
Given that the Fourier transform \( X(\omega) \) of the signal \( x(t) \) is as shown in the figure, determine the value of \( \int_{-\infty}^{+\infty} x(t) \, dt = \) ( ).

\textbf{Answer: }1.\\
\textbf{Response from GPT-4o: } To find the value of \( \int_{-\infty}^{+\infty} x(t) \, dt \), we need to evaluate \( X(0) \), the Fourier transform of \( x(t) \) at \( \omega = 0 \).

From the given figure,{\color{red} \( X(\omega) \) is a triangular function with a peak at \( \omega = 0 \) and a height of 1.5. Therefore, \( X(0) = 1.5 \).}

{\color{red}Thus, the value of \( \int_{-\infty}^{+\infty} x(t) \, dt \) is \( 1.5 \).

Final value: 1.5}

\vspace{0.4em}
\end{minipage}
\hrule
\vspace{0.4em}
\caption{ Image Perception Error examples of the response from GPT-4o. The error part in response is marked as {\color{red}red} color}.
\label{perception_error_1}
\end{figure}

\begin{figure}[th]
\hrule
\centering
\begin{minipage}[c]{0.5\textwidth}
\includegraphics[width=\textwidth]{    }
\end{minipage}
\hfill
\begin{minipage}[c]{0.98\textwidth}
\small

\textbf{Question: }
In the following circuits, which circuit (   ) exhibits the following phenomenon: After pressing the start button SB2, the contactor KM is engaged, but once the start button is released, the contactor KM immediately disengages?
  
A. graph A  
B. graph B  \\
C. graph C  
D. graph D  \\
\textbf{Answer: }C.\\
\textbf{Response from GPT-4o: }The circuit that exhibits the phenomenon where the contactor KM is engaged after pressing the start button SB2, but immediately disengages once the button is released, {\color{red}is circuit B. This is because circuit B lacks a holding contact to maintain the circuit after the button is released.}

{\color{red}Correct option: B. graph B}

\vspace{0.4em}
\end{minipage}
\hrule

\begin{minipage}[c]{0.2\textwidth}
\includegraphics[width=\textwidth]{    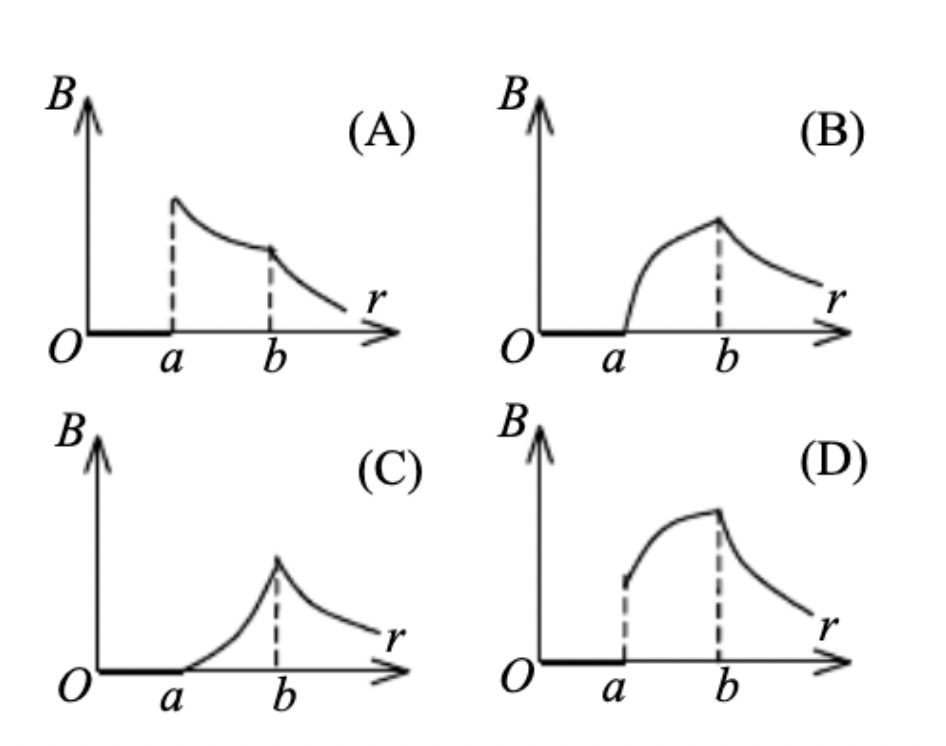}
\end{minipage}
\hfill
\begin{minipage}[c]{0.98\textwidth}
\small

\textbf{Question: }
The inner and outer radii of an infinitely long hollow cylindrical conductor carrying current are \( a \) and \( b \), respectively. The current is uniformly distributed over the cross-section of the conductor. The qualitative relationship between the magnitude of the magnetic field \( \vec{B} \) at various points in space and the distance \( r \) from the center axis of the cylinder is shown in the figure. The correct graph is:

A. graph A  
B. graph B  \\
C. graph C  
D. graph D \\
\textbf{Answer: }B.\\
\textbf{Response from GPT-4o: } To determine the correct graph, we need to consider the behavior of the magnetic field \( \vec{B} \) in different regions:

1. **Inside the hollow part (\( r < a \))**: The magnetic field is zero because there is no current enclosed.

2. **Within the conductor (\( a \leq r \leq b \))**: The magnetic field increases linearly with \( r \) because the current enclosed increases with \( r \).

3. **Outside the conductor (\( r > b \))**: The magnetic field decreases with \( r \) as the total current is enclosed and the field behaves like that of a long straight wire.

{\color{red}The correct graph that represents this behavior is graph D.

**Answer: D**}
\vspace{0.4em}
\end{minipage}
\hrule

\begin{minipage}[c]{0.2\textwidth}
\includegraphics[width=\textwidth]{    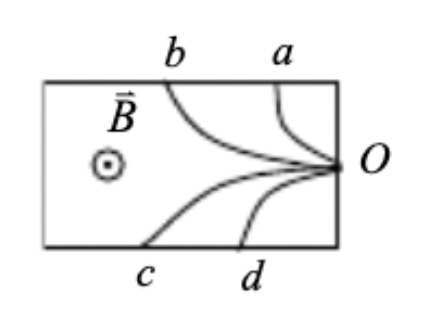}
\end{minipage}
\hfill
\begin{minipage}[c]{0.98\textwidth}
\small

\textbf{Question: }
The image shows the deflection trajectories of four charged particles after being injected into a uniform magnetic field at point \( O \) in the same direction, perpendicular to the magnetic field lines. The magnetic field direction is out of the plane of the paper. The masses of the four particles are equal, and their charges are also equal. The trajectory of the negatively charged particle with the greatest kinetic energy is:

(A) \( Oa \)  
(B) \( Ob \)  \\
(C) \( Oc \)  
(D) \( Od \)\\
\textbf{Answer: }C.\\
\textbf{Response from GPT-4o: } The trajectory of a charged particle in a magnetic field is determined by the Lorentz force, which causes the particle to move in a circular path. The radius of this path is given by the formula:

\[ r = \frac{mv}{|q|B} \]

where \( m \) is the mass, \( v \) is the velocity, \( q \) is the charge, and \( B \) is the magnetic field strength. Since the masses and charges are equal, the radius is directly proportional to the velocity, and hence the kinetic energy (since \( KE = \frac{1}{2}mv^2 \)).

The particle with the greatest kinetic energy will have the largest radius of curvature. In the image, the {\color{red}trajectory with the largest radius is \( Ob \).}

Therefore, the correct answer is:

{\color{red}(B) \( Ob \)}

\vspace{0.4em}
\end{minipage}
\hrule
\vspace{0.4em}
\caption{ Image Perception Error examples of the response from GPT-4o. The error part in response is marked as {\color{red}red} color}.
\label{perception_error_2}
\end{figure}

\begin{figure}[th]
\hrule
\centering
\begin{minipage}[c]{0.4\textwidth}
\includegraphics[width=\textwidth]{    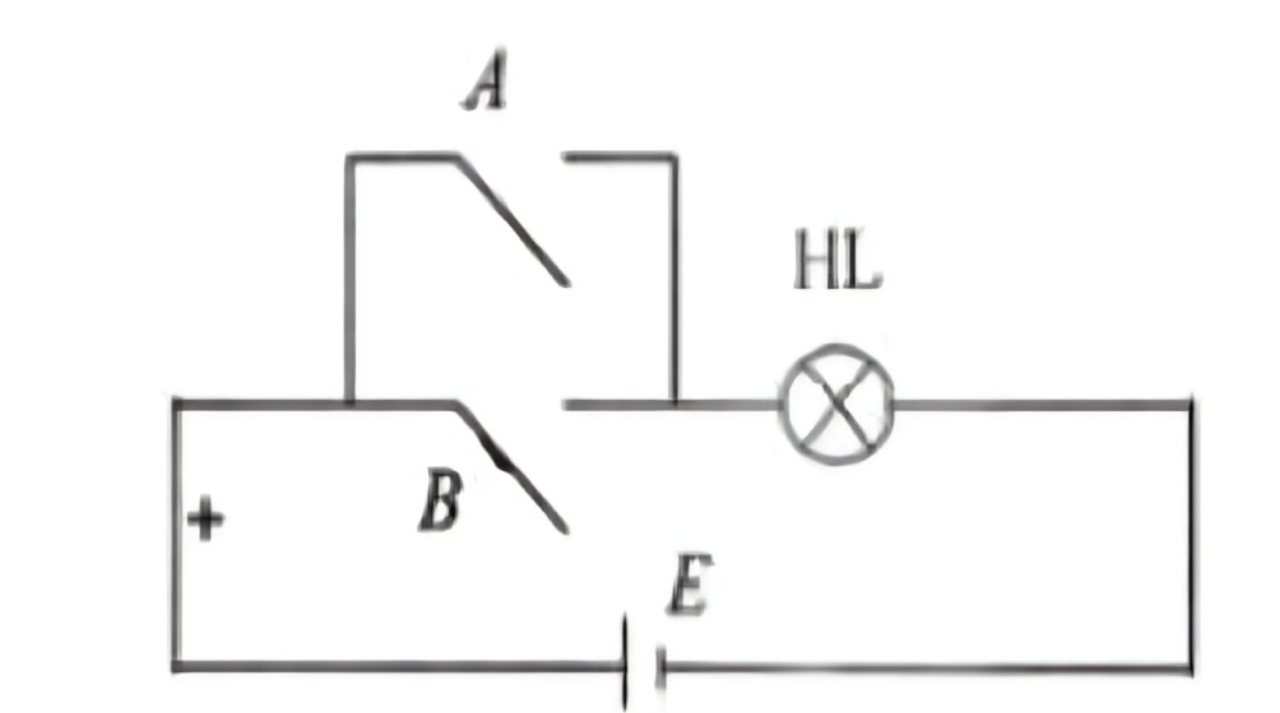}
\end{minipage}
\hfill
\begin{minipage}[c]{0.98\textwidth}
\small

\textbf{Question: }
In the logic circuit composed of switches shown in figure, set switches A and B to "1" (connected), and set to "0" (disconnected), if the light is on "1" and off "0", then this circuit is ( )\\
A. "AND" gate\\
B. "OR" gate\\
C. "NOT" gate\\
D. "NAND

\textbf{Answer: }B.\\
\textbf{Response from GPT-4o: } {\color{red}The circuit shown is an "AND" gate.} {\color{red}For the light to be on "1", both switches A and B must be closed (set to "1"). If either switch is open, the circuit is broken, and the light will be off ("0").Correct option: 
A. "AND" gate.}

\vspace{0.4em}
\end{minipage}
\hrule
\begin{minipage}[c]{0.3\textwidth}
\includegraphics[width=\textwidth]{    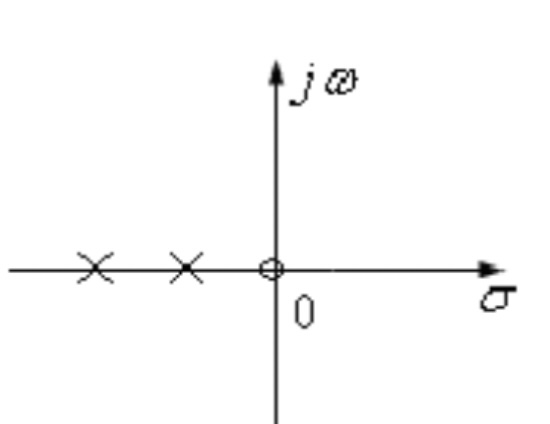}
\end{minipage}
\hfill
\begin{minipage}[c]{0.98\textwidth}
\small

\textbf{Question: }
If the pole-zero distribution of the system \( \mathrm{H}(\mathrm{s}) \) is as shown in Figure 1, determine which type of filter network it is:\\ 
A. Low-pass  \\
B. High-pass  \\
C. Band-pass  \\
D. Band-stop\\
\textbf{Answer: }C.\\
\textbf{Response from GPT-4o: } The pole-zero plot shows zeros on the left side of the imaginary axis and \color{red}a pole at the origin. {\color{red}This configuration typically represents a **high-pass filter**, as it attenuates low frequencies and allows high frequencies to pass.

Answer: B. High-pass}

\vspace{0.4em}
\end{minipage}
\hrule

\begin{minipage}[c]{0.4\textwidth}
\includegraphics[width=\textwidth]{    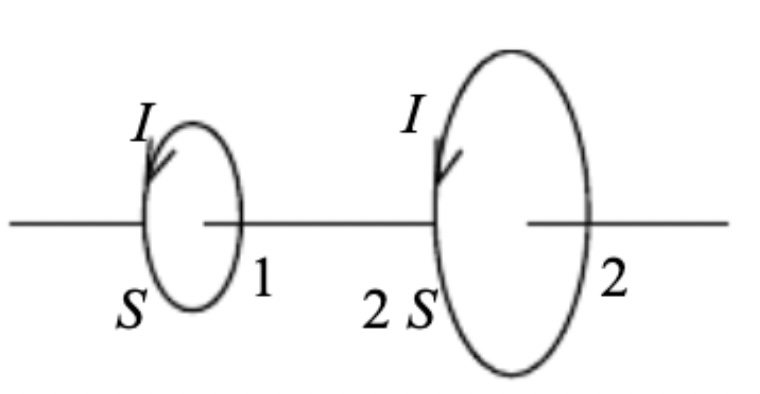}
\end{minipage}
\hfill
\begin{minipage}[c]{0.98\textwidth}
\small

\textbf{Question: }
Two circular coils, 1 and 2, with areas \( S \) and \( 2S \) respectively, are placed as shown in the image. Both coils carry the same current \( I \). The magnetic flux through coil 2 produced by the current in coil 1 is denoted as \( \Phi_{21} \), and the magnetic flux through coil 1 produced by the current in coil 2 is denoted as \( \Phi_{12} \). The relationship between \( \Phi_{21} \) and \( \Phi_{12} \) is:
 
(A) \( \Phi_{21} = 2 \Phi_{12} \)  \\
(B) \( \Phi_{21} > \Phi_{12} \)  \\
(C) \( \Phi_{21} = \Phi_{12} \)  \\
(D) \( \Phi_{21} = \frac{1}{2} \Phi_{12} \)\\

\textbf{Answer: }A.\\
\textbf{Response from GPT-4o: } The mutual inductance between two coils is the same regardless of which coil is the source of the magnetic field. Therefore, {\color{red}the magnetic flux through coil 2 due to coil 1 (\( \Phi_{21} \)) is equal to the magnetic flux through coil 1 due to coil 2 (\( \Phi_{12} \)).}

Thus, the relationship is:

{\color{red}(C) \( \Phi_{21} = \Phi_{12} \)}
\vspace{0.4em}
\end{minipage}
\hrule
\vspace{0.4em}
\caption{ Reasoning Error examples of the response from GPT-4o. The error part in response is marked as {\color{red}red} color}.
\label{reason_erro_1}
\end{figure}

\begin{figure}[th]
\hrule
\centering
\begin{minipage}[c]{0.4\textwidth}
\includegraphics[width=\textwidth]{    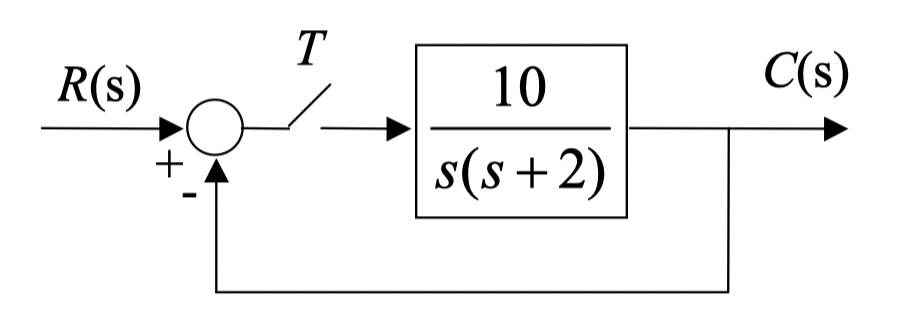}
\end{minipage}
\hfill
\begin{minipage}[c]{0.98\textwidth}
\small

\textbf{Question: }
Given the sampled control system as shown in figure, the open-loop pulse transfer function is \( G_{\mathrm{o}}(z) = \frac{5\left(1 - \mathrm{e}^{-2 T}\right) z}{(z - 1)\left(z - \mathrm{e}^{-2 T}\right)} \). When the sampling interval \( T = 0.1 \) seconds, is the system stable?
  
A. Yes  \\
B. No

\textbf{Answer: }A.\\
\textbf{Response from GPT-4o: } To determine the stability of the system, we need to examine the poles of the open-loop pulse transfer function \( G_{\mathrm{o}}(z) \).

The given transfer function is:

\[
G_{\mathrm{o}}(z) = \frac{5\left(1 - \mathrm{e}^{-2 T}\right) z}{(z - 1)\left(z - \mathrm{e}^{-2 T}\right)}
\]

For stability, all poles of the transfer function must lie inside the unit circle in the z-plane.

The poles are at \( z = 1 \) and \( z = \mathrm{e}^{-2T} \).

Given \( T = 0.1 \), we calculate:

\[
\mathrm{e}^{-2T} = \mathrm{e}^{-0.2} \approx 0.8187
\]

The poles are \( z = 1 \) and \( z \approx 0.8187 \).

{\color{red}Since the pole at \( z = 1 \) is on the unit circle, the system is marginally stable, not asymptotically stable.

Therefore, the system is not stable in the sense of all poles being strictly inside the unit circle.

**Answer: B. No**}

\vspace{0.4em}
\end{minipage}
\hrule
\begin{minipage}[c]{0.5\textwidth}
\includegraphics[width=\textwidth]{    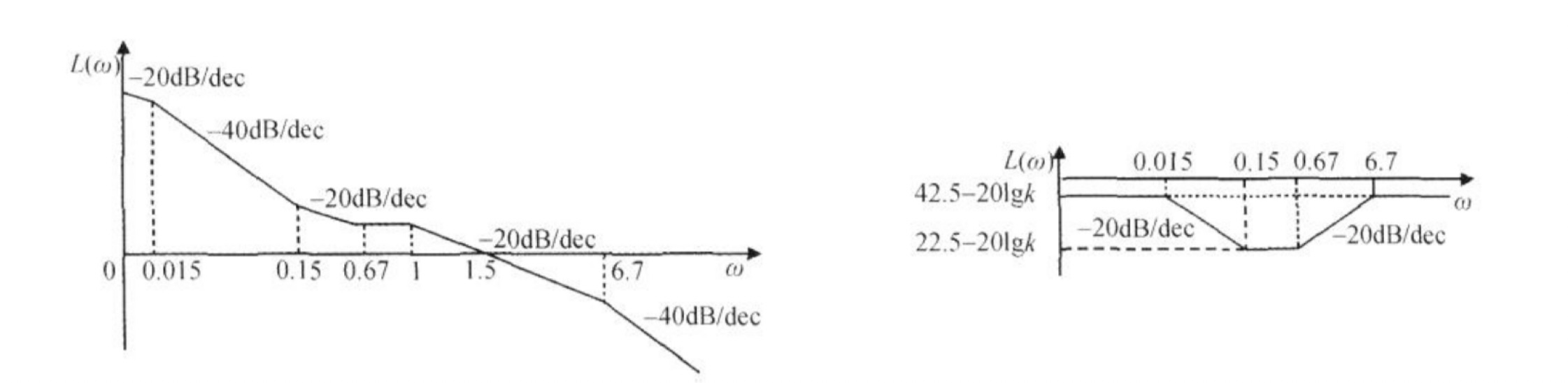}
\end{minipage}
\hfill
\begin{minipage}[c]{0.98\textwidth}
\small

\textbf{Question: }
The block diagram of a discrete system is shown in figure, where \( T = 0.25 \) s and the system's reference input signal is \( r(t) = 8 + 0.5t \). Determine the range of values for \( k \) that will ensure the steady-state error \( e_{ss} < 0.5 \).

A. \( 1 < k < 2.67 \)  \\
B. \( 0.5 < k < 2.5 \)  \\
C. \( 1 < k < 2.67 \)  \\
D. \( 0.8 < k < 2.2 \)\\
\textbf{Answer: }C.\\
\textbf{Response from GPT-4o: } To determine the range of values for \( k \) that will ensure the steady-state error \( e_{ss} < 0.5 \), we need to analyze the system's transfer function and apply the final value theorem.

The system is a discrete-time system with a transfer function given by:

\[ G(s) = \frac{k e^{-0.5s}}{s} \]

{\color{red}The reference input signal is \( r(t) = 8 + 0.5t \), which consists of a step input and a ramp input. The steady-state error for a ramp input in a type 1 system is given by:

\[ e_{ss} = \frac{1}{k_v} \]

where \( k_v \) is the velocity error constant. For a type 1 system, \( k_v \) is determined by:

\[ k_v = \lim_{s \to 0} sG(s) = \lim_{s \to 0} \frac{k e^{-0.5s}}{1} = k \]

Thus, the steady-state error for the ramp input is:
\[ e_{ss} = \frac{1}{k} \]
We want \( e_{ss} < 0.5 \), so:
\[ \frac{1}{k} < 0.5 \]
Solving for \( k \), we get:
\( k > 2 \)
Therefore, the range of values for \( k \) that ensures \( e_{ss} < 0.5 \) is:
\( k > 2 \)
From the given choices, the correct option is:

A. \( 1 < k < 2.67 \)

So, the answer is A.}

\vspace{0.4em}
\end{minipage}
\hrule
\vspace{0.4em}
\caption{ Reasoning Error examples of the response from GPT-4o. The error part in response is marked as {\color{red}red} color}.
\label{reason_erro_2}
\end{figure}

\begin{figure}[th]
\hrule
\centering
\begin{minipage}[c]{0.4\textwidth}
\includegraphics[width=\textwidth]{       }
\end{minipage}
\hfill
\begin{minipage}[c]{0.98\textwidth}
\small
\textbf{Caption context:} Image caption context:

The diagram consists of four subplots labeled A, B, C, and D. Each subplot represents a graph of current versus voltage, specifically for a transistor, likely a MOSFET or JFET, based on the axes labels. Here is a detailed description of each subplot:

 Subplot A:
- **Axes**: 
  - Horizontal axis: \( V_{GS} \) (Gate-Source Voltage)
  - Vertical axis: \( I_D \) (Drain Current)
- **Graph**: 
  - The curve starts in the third quadrant and moves into the first quadrant, indicating a negative \( V_{GS} \) initially, with \( I_D \) increasing as \( V_{GS} \) becomes positive.
  - The curve is nonlinear, showing a gradual increase in \( I_D \) with increasing \( V_{GS} \).

 Subplot B:
- **Axes**: 
  - Horizontal axis: \( V_{GS} \) (Gate-Source Voltage)
  - Vertical axis: \( I_D \) (Drain Current)
- **Graph**: 
  - The curve starts near the origin and sharply rises in the first quadrant, indicating a rapid increase in \( I_D \) with a small increase in \( V_{GS} \).
  - The curve is steep, suggesting a strong dependency of \( I_D \) on \( V_{GS} \).

Subplot C:
- **Axes**: 
  - Horizontal axis: \( V_{GS} \) (Gate-Source Voltage)
  - Vertical axis: \( I_D \) (Drain Current)
- **Graph**: 
  - The curve starts at the origin and rises steeply in the first quadrant, similar to subplot B, but with a more pronounced initial slope.
  - This indicates a very sensitive response of \( I_D \) to changes in \( V_{GS} \).

Subplot D:
- **Axes**: 
  - Horizontal axis: \( V_{GS} \) (Gate-Source Voltage)
  - Vertical axis: \( I_D \) (Drain Current)
- **Graph**: 
  - The curve starts in the third quadrant and moves into the first quadrant, similar to subplot A, but with a more pronounced initial curvature.
  - The curve shows a nonlinear increase in \( I_D \) as \( V_{GS} \) becomes positive, indicating a threshold behavior.

General Observations:
- Each subplot represents a different characteristic curve of a transistor, likely showing different modes of operation or types of transistors.
- The \( V_{GS} \) and \( I_D \) relationship is crucial for understanding the behavior of the transistor in different configurations or biasing conditions.
- The curves suggest different levels of sensitivity and threshold behavior, which are important for designing and analyzing circuits involving these components.
\\
\textbf{Question:}The graph that correctly describes the transfer characteristics of an enhancement-mode NMOSFET (assuming current always flows into the drain as the positive direction) is ().

A. graph A  
B. graph B  
C. graph C  
D. graph D
\\
\textbf{Response w/o caption:} The correct graph that describes the transfer characteristics of an enhancement-mode NMOSFET is graph C. 

C
\textbf{Response w/ caption: } {\color{red}The graph that correctly describes the transfer characteristics of an enhancement-mode NMOSFET is **graph B**. }

In an enhancement-mode NMOSFET, the drain current \( I_D \) starts to increase significantly only after the gate-source voltage \( V_{GS} \) exceeds a certain threshold voltage. {\color{red}This behavior is depicted in graph B, where \( I_D \) sharply rises in the first quadrant as \( V_{GS} \) increases.}
\vspace{0.2em}
\end{minipage}
\hrule

\begin{minipage}[c]{0.25\textwidth}
\includegraphics[width=\textwidth]{       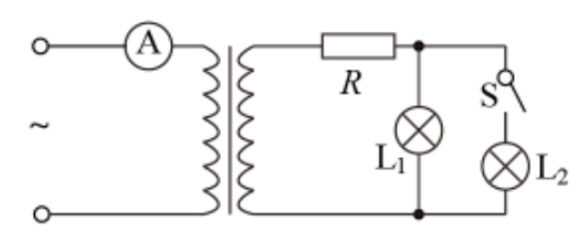}
\end{minipage}
\hfill
\begin{minipage}[c]{0.98\textwidth}
\small
\textbf{Caption context:} Image caption context:

The diagram depicts an electrical circuit with the following components and connections:

1. **AC Source**: 
   - Represented by the symbol for alternating current (AC) on the left side of the diagram.
   - It provides the input power to the circuit.

2. **Ammeter (A)**:
   - Connected in series with the primary coil of a transformer.
   - Measures the current flowing through the primary side.

3. **Transformer**:
   - Consists of two coils: the primary coil (connected to the AC source and ammeter) and the secondary coil.
   - The primary coil is on the left, and the secondary coil is on the right.

4. **Resistor (R)**:
   - Connected in series with the secondary coil of the transformer.
   - Represents a load or impedance in the circuit.

5. **Lamp L1**:
   - Connected in parallel with the resistor (R).
   - Indicates the presence of current through the secondary circuit.

6. **Switch (S)**:
   - Connected in series with another lamp (L2).
   - Controls the current flow to lamp L2.

7. **Lamp L2**:
   - Connected in series with the switch (S).
   - Illuminates when the switch is closed, indicating current flow through this path.

8. **Connections**:
   - The secondary coil of the transformer is connected to both the resistor (R) and lamp L1 in parallel.
   - The switch (S) and lamp L2 are connected in series, forming a parallel branch with the resistor and lamp L1.

This setup can be used to study the behavior of AC circuits, the effect of transformers, and the operation of parallel and series components in electrical systems.
\\
\textbf{Question:} As shown in the figure, the primary coil is input with a stable AC voltage, and switch \( S \) is in the open state. At a certain moment, bulb \( \mathbf{L}_1 \) becomes slightly dimmer. The possible reason could be:
 
A. The input voltage increased  
B. Switch \( \mathbf{S} \) was closed \\ 
C. The resistor \( \boldsymbol{R} \) short-circuited  
D. The ammeter experienced an open circuit
\\
\textbf{Response w/o caption:} The possible reason bulb \( \mathbf{L}_1 \) becomes slightly dimmer is:
B. Switch \( \mathbf{S} \) was closed

When switch \( S \) is closed, current can flow through bulb \( \mathbf{L}_2 \), reducing the current through \( \mathbf{L}_1 \) and making it dimmer.
\textbf{Response w/ caption: } The possible reason for bulb \( L_1 \) becoming slightly dimmer is:
{\color{red}C. The resistor \( \boldsymbol{R} \) short-circuited}

{\color{red}If the resistor \( R \) short-circuits, it would reduce the total resistance in the parallel branch, causing more current to flow through the shorted path and less through \( L_1 \), making it dimmer.}
\vspace{0.2em}
\end{minipage}
\hrule
\caption{Examples of LMM errors with \textit{captions} in EEE-Bench o fGPT-4o. We mark the errors in response with {\color{red}red} color}
\label{error_caption_1}
\end{figure}

In this section, we give more examples of error analysis. \autoref{perception_error_1} and \autoref{perception_error_2} are examples of image perception errors. \autoref{reason_erro_1} and \autoref{reason_erro_2} are examples of reasoning errors.
\label{appendix: more failure example}

\section{More examples of LMM laziness}
\label{app:more_lazy}
In this section, we provide more examples of the laziness phenomenon of existing LMMs. \autoref{lazy_1}, \autoref{lazy_2}, \autoref{lazy_3}, \autoref{lazy_4}, \autoref{lazy_5}, \autoref{lazy_6}, \autoref{lazy_7} and  \autoref{lazy_8} show examples of LMM laziness phenomenon on EEE-Bench. \autoref{lazy_9} and \autoref{lazy_10} show examples of LMM laziness phenomenon on MathVista. We can find that when there is (misleading) information for problem-solving in text, the LMMs will not try to read the relevant information in the images, even if this information is spurious. Consequently, LMMs generate wrong answers with this spurious information.

\begin{figure}
\hrule
    \centering
    \begin{minipage}[c]{0.35\textwidth}
\includegraphics[width=\textwidth]{       }
\end{minipage}
\hfill
\begin{minipage}[c]
{0.98\textwidth}
\small
\textbf{caption context:} Image caption context:

The diagram represents a periodic signal \( x_3(t) \) with a period of 10 units. Here is a detailed description of the waveform:

1. **Signal Type**: The signal is a square wave.

2. **Periodicity**: The function is periodic with a period \( T = 10 \). This is indicated by the equation \( x_3(t) = x_3(t + 10) \).

3. **Amplitude Levels**: The waveform alternates between two amplitude levels:
   - \( +1 \) 
   - \( -1 \)

4. **Time Intervals**:
   - The signal is at \( +1 \) from \( t = 0 \) to \( t = 1 \).
   - The signal drops to \( -1 \) from \( t = 1 \) to \( t = 3 \).
   - The signal returns to 0 from \( t = 3 \) to \( t = 10 \).

5. **Repetition**: This pattern repeats every 10 units along the time axis, both to the left and right, as indicated by the ellipses (\(\ldots\)).

6. **Key Points**:
   - At \( t = 0 \), the signal jumps to \( +1 \).
   - At \( t = 1 \), the signal drops to \( -1 \).
   - At \( t = 3 \), the signal returns to 0.
   - At \( t = 10 \), the pattern repeats.

This description captures all the essential features of the waveform, allowing for analysis or reconstruction without the visual diagram.
\\
\textbf{Question:} Determine the Fourier series coefficient $c_0$ for x3(t) shown in figure. $c_0$ = ()
\\
\textbf{Response w/o caption:} To find the Fourier series coefficient \( c_0 \), which represents the average value of the function over one period, we need to integrate the function over one period and divide by the period length.

Given the periodic function \( x_3(t) \) with period \( T = 10 \), we can calculate \( c_0 \) as follows:
\[ c_0 = \frac{1}{T} \int_0^T x_3(t) \, dt \]
From the graph, the function \( x_3(t) \) is:

- \( 1 \) from \( t = 0 \) to \( t = 1 \)
- \( -1 \) from \( t = 2 \) to \( t = 3 \)
- \( 0 \) elsewhere in the period

Now, calculate the integral:
\[ c_0 = \frac{1}{10} \left( \int_0^1 1 \, dt + \int_2^3 (-1) \, dt \right) \]
\[ c_0 = \frac{1}{10} \left( [t]_0^1 + [-t]_2^3 \right) \]
\[ c_0 = \frac{1}{10} \left( (1 - 0) + (-(3) - (-(2))) \right) \]
\[ c_0 = \frac{1}{10} \left( 1 - 1 \right) \]
\[ c_0 = \frac{1}{10} \times 0 \]
\[ c_0 = 0.0 \]
Thus, the Fourier series coefficient \( c_0 \) is \( 0.0 \).
\\
\textbf{Response w/ spurious caption:} 
To find the Fourier series coefficient \( c_0 \) for the signal \( x_3(t) \), we need to calculate the average value of the signal over one period. The coefficient \( c_0 \) is given by:
\[
c_0 = \frac{1}{T} \int_0^T x_3(t) \, dt
\]
Given the period \( T = 10 \), we can break the integral into segments based on the time intervals provided:

1. From \( t = 0 \) to \( t = 1 \), \( x_3(t) = 1 \).
{\color{red}2. From \( t = 1 \) to \( t = 3 \), \( x_3(t) = -1 \).}
3. From \( t = 3 \) to \( t = 10 \), \( x_3(t) = 0 \).

Now, calculate the integral:
\[
c_0 = \frac{1}{10} \left( \int_0^1 1 \, dt + {\color{red}\int_1^3 (-1) \, dt} + \int_3^{10} 0 \, dt \right)
\]
\[
= \frac{1}{10} \left( [t]_0^1 + [-t]_1^3 + [0]_3^{10} \right)
\]
\[
= \frac{1}{10} \left( (1 - 0) + (-(3 - 1)) + 0 \right)
\]
\[
= \frac{1}{10} \left( 1 - 2 \right)
\]
\[
= \frac{1}{10} \times (-1)
\]
\[
= -0.1
\]

{\color{red}Thus, the Fourier series coefficient \( c_0 \) is \(-0.1\).}
\vspace{0.4em}
\end{minipage}
\hrule
\vspace{0.4em}
\caption{Examples of LMM errors with \textit{captions} in EEE-Bench o f GPT-4o. We mark the errors in response with {\color{red}red} color}
\label{error_caption_2}
\end{figure}

\begin{figure}[th]
\hrule
\centering
\begin{minipage}[c]{0.3\textwidth}
\includegraphics[width=\textwidth]{       }
\end{minipage}
\hfill
\begin{minipage}[c]{0.98\textwidth}
\small
\textbf{Spurious context:} Image caption context:

The diagram depicts a diode {\color{red}{AND}} gate circuit with a {\color{red}{positive}} voltage reference. Here is a detailed description:

1. Components:
   - Two Diodes: The {\color{red}{cathodes}} of both diodes are connected to separate input voltages.
   - Resistor (R): Connected {\color{red}{in parallel}} with the {\color{red}{anodes}} of the diodes.

2. Connections:
   - Input Voltage \( V_A \): Connected to the {\color{red}{cathode}} of the first diode.
   - Input Voltage \( V_B \): Connected to the {\color{red}{cathode}} of the second diode.
   - The {\color{red}{anodes}} of both diodes are connected together at a junction point.
   - Output Voltage \( V_F \): Taken from the junction point where the {\color{red}{anodes}} of the diodes meet.
   - The other end of the resistor is connected to a {\color{red}{+12V}} supply.

3. Voltage Reference:
   - The circuit is referenced to a {\color{red}{+12V}} supply, which is connected to the other end of the resistor.

4. Functionality:
   - This configuration allows the output voltage \( V_F \) to follow the {\color{red}{lower}} of the two input voltages \( V_A \) or \( V_B \), {\color{red}{plus}} the diode forward voltage drop, as long as it is below {\color{red}{+12V}}.

This setup is typically used in applications where the {\color{red}{lowest}} of two input voltages needs to be selected, with a {\color{red}{positive}} voltage reference for biasing.
\\
\textbf{Question:}In the circuit shown in figure, $V_A = 3V,  V_B = 0V$, if the forward voltage drop of the diode is ignored, then $V_F$ is ( )
A. -12 V B. 9 V C. 0 V D. 3V
\\
\textbf{Response w/o spurious caption:} To find \( V_F \), we need to consider the behavior of the diodes. Since the forward voltage drop is ignored, the diodes will conduct if the voltage at their anode is higher than at their cathode.

1. **Diode connected to \( V_A \):** 
   - \( V_A = 3V \)
   - The diode will conduct because \( 3V > V_F \).

2. **Diode connected to \( V_B \):**
   - \( V_B = 0V \)
   - The diode will not conduct because \( 0V < V_F \).

Since the diode connected to \( V_A \) conducts, \( V_F \) will be equal to \( V_A \), which is \( 3V \).

Therefore, \( V_F = 3V \).

The correct option is D. 
\\
\textbf{Response w/ spurious caption:} 
In the given circuit, {\color{red}the output voltage \( V_F \) will follow the lower of the two input voltages \( V_A \) or \( V_B \) since the diodes will conduct for the lower voltage.} Here, \( V_A = 3V \) and \( V_B = 0V \).

Since \( V_B \) is lower, {\color{red}\( V_F \) will be \( 0V \)}.

{\color{red}The correct answer is: C. 0V}
\vspace{0.4em}
\end{minipage}
\hrule
\begin{minipage}[c]{0.35\textwidth}
\includegraphics[width=\textwidth]{       }
\end{minipage}
\hfill
\begin{minipage}[c]
{0.98\textwidth}
\small
\textbf{Spurious context:} Image caption context:

The diagram is a timing diagram showing three digital waveforms labeled A, B, and F. Here is a detailed description:

1. Waveform A:
   - The waveform is a {\color{red}{sine wave}} with a {\color{red}{75\% duty cycle}}.
   - It alternates between a high state and a low state.
   - The high state duration is {\color{red}{longer than the low state duration}}.
   - The waveform starts with a {\color{red}{low state}}.

2. Waveform B:
   - This waveform is also a {\color{red}{sine wave}} but with a {\color{red}{shorter high state}} compared to waveform A.
   - It starts with a {\color{red}{high state}}, then transitions to a {\color{red}{low state}}.
   - The high state duration is {\color{red}{shorter than the low state duration}}.

3. Waveform F:
   - This waveform is a combination of waveforms A and B.
   - It starts with a {\color{red}{high state}}.
   - The waveform transitions to a {\color{red}{low state}} when {\color{red}{either A or B is high}}.
   - It returns to a {\color{red}{high state}} when {\color{red}{both A and B are low}}.

4. Timing Relationships:
   - The transitions of waveforms A and B are {\color{red}{misaligned}} such that waveform F is {\color{red}{low}} only when both A and B are simultaneously {\color{red}{high}}.
   - The dashed vertical lines indicate the points of transition for synchronization.

This timing diagram is typically used in digital electronics to illustrate the behavior of logic circuits.
\\
\textbf{Question:}In the circuit shown in figure, if the waveform of the input terminal A and output terminal F of the NAND gate is as shown in figure, then this circuit is ( )\\
A. AND gate
B. OR gate
C. NOT gate
D. NAND gate
\\
\textbf{Response w/o spurious caption:} The waveform at terminal F is the inverse of the waveform at terminal A. This indicates that the circuit is functioning as a NOT gate, which inverts the input signal.

Correct option: C. NOT gate
\\
\textbf{Response w/ spurious caption:} 
{\color{red}The waveform F is low only when both A and B are high, which is characteristic of a NAND gate}. A NAND gate outputs low only when all its inputs are high.

Therefore, the circuit is a:

{\color{red}D. NAND gate}
\vspace{0.4em}
\end{minipage}
\hrule
\vspace{0.4em}
\caption{Examples of LMM laziness in EEE-Bench. We mark the \textit{spurious text} in the caption and error in response with {\color{red}red} color}
\label{lazy_1}
\end{figure}

\begin{figure}[th]
\hrule
\centering
\begin{minipage}[c]{0.45\textwidth}
\includegraphics[width=\textwidth]{       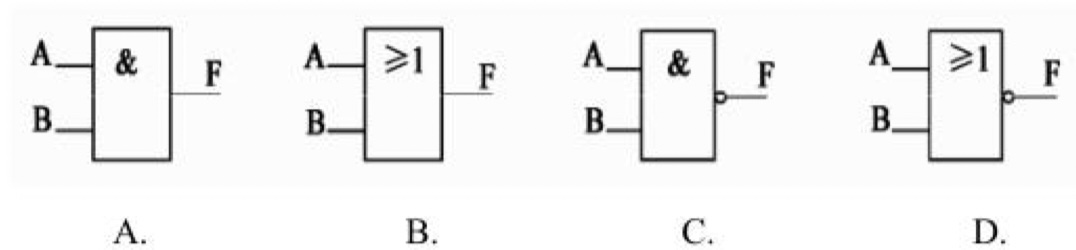}
\end{minipage}
\hfill
\begin{minipage}[c]{0.98\textwidth}
\small
\textbf{Spurious context:} Image caption context:

The diagram consists of four logic gate configurations labeled A, B, C, and D. Each configuration has two inputs labeled A and B, and one output labeled F. Here is a detailed description of each configuration:

A. {\color{red}OR Gate}
- **Symbol**: A rectangle with a $\ge $'1' inside.
- **Inputs**: Two inputs labeled A and B on the left side.
- **Output**: One output labeled F on the right side.
- **Function**: The output F is true (1) if at least one of the inputs A or B is true (1).

B. {\color{red}AND Gate}
- **Symbol**: A rectangle with an ampersand (\&) inside.
- **Inputs**: Two inputs labeled A and B on the left side.
- **Output**: One output labeled F on the right side.
- **Function**: The output F is true (1) only if both inputs A and B are true (1).

C. {\color{red}NOR Gate}
Symbol: A rectangle with a $"\ge 1"$ inside, followed by a small circle (representing negation) at the output.
- **Inputs**: Two inputs labeled A and B on the left side.
- **Output**: One output labeled F on the right side.
- **Function**: The output F is true (1) only if both inputs A and B are false (0).

D. {\color{red}NAND Gate}
- **Symbol**: A rectangle with an ampersand (\&) inside, followed by a small circle (representing negation) at the output.
- **Inputs**: Two inputs labeled A and B on the left side.
- **Output**: One output labeled F on the right side.
- **Function**: The output F is false (0) only if both inputs A and B are true (1); otherwise, it is true (1).

These configurations represent basic logic gates used in digital circuits, each performing a specific logical operation based on the inputs provided.
\\
\textbf{Question:}Among the logic symbols shown in figure, the one that can realize the logic function \( F = \overline{AB} \) is ( )

A. circuit A
B. circuit B
C. circuit C
D. circuit D
\\
\textbf{Response w/o spurious caption:} The logic function \( F = \overline{AB} \) is a NAND operation. The circuit that represents a NAND gate is circuit C, which shows an AND gate with a NOT symbol (a small circle) at the output.

Correct option: C
\\
\textbf{Response w/ spurious caption:} 
The logic function \( F = \overline{AB} \) is realized by a NAND gate. {\color{red} In the diagram, the NAND gate is represented by configuration D}.

{\color{red} So, the correct option is D.}
\vspace{0.4em}
\end{minipage}
\hrule
\begin{minipage}[c]{0.25\textwidth}
\includegraphics[width=\textwidth]{       }
\end{minipage}
\hfill
\begin{minipage}[c]
{0.98\textwidth}
\small
\textbf{Spurious context:} Image caption context:

The diagram appears to be a {\color{red}Bode plot}, which is commonly used in control systems to analyze the {\color{red}frequency response} of a system. Here is a detailed description:

1. Axes:
   - The horizontal axis is labeled as the {\color{red}frequency axis}.
   - The vertical axis is labeled as the {\color{red}magnitude in dB}.

2. Curves:
   - There is a single curve that shows the gain as a function of {\color{red}frequency}.

3. Points and Regions:
   - The origin is marked as \( O \).
   - There are two highlighted points:
     - One at a {\color{red}low frequency}, labeled \( M \).
     - One at a {\color{red}high frequency}, labeled \( S \).

4. Interpretation:
   - The plot is used to determine the {\color{red}gain and phase margins} of a system.
   - It helps in understanding how the system will respond to {\color{red}different frequencies}, particularly in terms of amplification and attenuation.

This description should help in analyzing the {\color{red}frequency response} and stability of a control system using the {\color{red}Bode plot} method.
\\
\textbf{Question:}For the ideal MOS structure shown in the figure, is the semiconductor in the question N-type or P-type?\\
A. N-type
B. P-type
C. can not be determined
\\
\textbf{Response w/o spurious caption:}To determine the type of semiconductor in the MOS structure, we need to look at the charge distribution. In the figure, the charge density ($\rho$) is positive in the semiconductor (S) region. This indicates that the semiconductor is P-type, as positive charge density in the semiconductor region is due to the presence of holes, which are the majority carriers in P-type semiconductors.

Answer: B. P-type
\\
\textbf{Response w/ spurious caption:} 
The image provided is not a Bode plot but rather a diagram related to a MOS structure. The vertical axis is labeled \( \rho \), which typically represents charge density. 

In the context of MOS structures:
- If the charge density is negative (below the horizontal axis) at point \( M \), it indicates the presence of electrons, suggesting an N-type semiconductor.
- If the charge density is positive (above the horizontal axis) at point \( S \), it indicates the presence of holes, suggesting a P-type semiconductor.

{\color{red}Given the diagram, the charge density at \( M \) is negative, indicating an N-type semiconductor.

Therefore, the correct answer is:

A. N-type}
\vspace{0.4em}
\end{minipage}
\hrule
\vspace{0.4em}
\caption{Examples of LMM laziness in EEE-Bench. We mark the \textit{spurious text} in the caption and error in response with {\color{red}red} color}
\label{lazy_2}
\end{figure}

\begin{figure}[th]
\hrule
\centering
\begin{minipage}[c]{0.45\textwidth}
\includegraphics[width=\textwidth]{       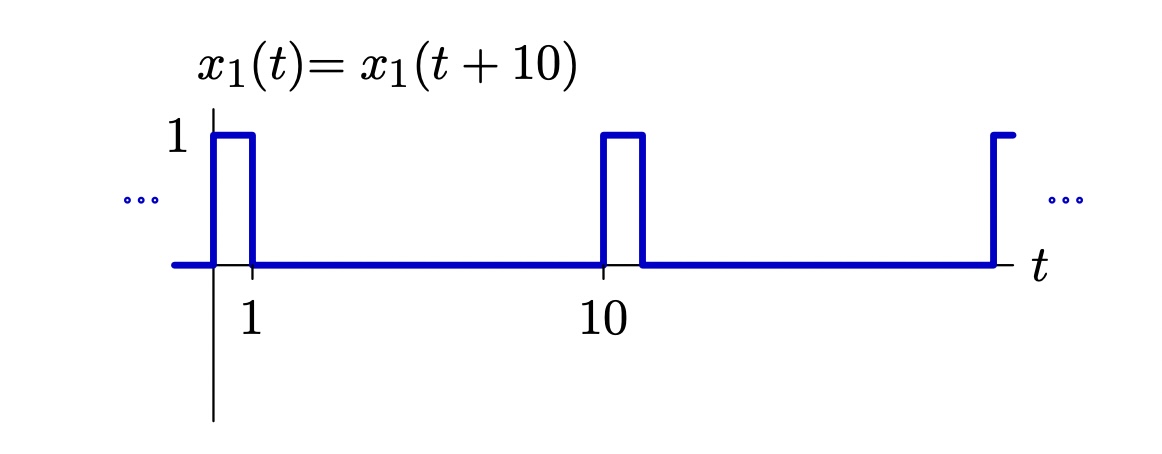}
\end{minipage}
\hfill
\begin{minipage}[c]{0.98\textwidth}
\small
\textbf{Spurious context:} Image caption context:

The diagram represents a periodic rectangular pulse signal \( x_1(t) \) with the following characteristics:

1. Signal Description:
   - The signal is periodic with a period of {\color{red}20} units of time.
   - The mathematical expression for the signal is \( x_1(t) = x_1(t + {\color{red}20}) \), indicating its periodic nature.

2. Pulse Characteristics:
   - Each pulse has an amplitude of {\color{red}2}.
   - The pulse width is {\color{red}2} units of time, starting at integer multiples of {\color{red}5} (e.g., 0, 5, 10, etc.).

3. Time Axis:
   - The horizontal axis represents time \( t \).
   - The vertical axis represents the amplitude of the signal.

4. Pulse Position:
   - The first pulse starts at \( t = 0 \) and ends at {\color{red}t = 2}.
   - The second pulse starts at {\color{red}t = 5} and ends at \( t = {\color{red}7} \).
   - This pattern continues indefinitely in both directions, as indicated by the ellipses (…).

5. Signal Behavior:
   - The signal is zero for all time intervals except during the pulse width of {\color{red}2} units at each period.

This description provides a complete understanding of the periodic rectangular pulse signal depicted in the diagram, suitable for analysis in various electrical and electronics engineering contexts.
\\
\textbf{Question:}Determine the Fourier series coefficients $a_0$ for $x_1(t)$ shown in figure. $a_0$=().
\\
\textbf{Response w/o spurious caption:} To find the Fourier series coefficient \( a_0 \) for the periodic function \( x_1(t) \), we need to calculate the average value of the function over one period.

The function \( x_1(t) \) is a periodic square wave with a period \( T = 10 \). It has a value of 1 from \( t = 0 \) to \( t = 1 \) and 0 from \( t = 1 \) to \( t = 10 \).

The formula for \( a_0 \) is:

\[
a_0 = \frac{1}{T} \int_0^T x_1(t) \, dt
\]

Substituting the given values:

\[
a_0 = \frac{1}{10} \left( \int_0^1 1 \, dt + \int_1^{10} 0 \, dt \right)
\]

\[
a_0 = \frac{1}{10} \left( [t]_0^1 + 0 \right)
\]

\[
a_0 = \frac{1}{10} (1 - 0) = 0.1
\]
Thus, the Fourier series coefficient \( a_0 \) is \( 0.1 \).
\\
\textbf{Response w/o spurious caption:}
To determine the Fourier series coefficient \( a_0 \) for the periodic rectangular pulse signal \( x_1(t) \), we need to calculate the average value of the signal over one period.

The formula for \( a_0 \) is:

\[
a_0 = \frac{1}{T} \int_0^T x_1(t) \, dt
\]

where \( T \) is the period of the signal. For this signal, \( T = 20 \).

The signal \( x_1(t) \) has an amplitude of 2 for a duration of 2 units of time in each period and is zero otherwise. Therefore, the integral over one period is:

{\color{red}\[
\int_0^{20} x_1(t) \, dt = \int_0^2 2 \, dt + \int_5^7 2 \, dt + \int_{10}^{12} 2 \, dt + \int_{15}^{17} 2 \, dt
\]}

Calculating each integral:

\[
\int_0^2 2 \, dt = 2 \times 2 = 4, \ 
\int_5^7 2 \, dt = 2 \times 2 = 4
\]
\[
\int_{10}^{12} 2 \, dt = 2 \times 2 = 4, \
\int_{15}^{17} 2 \, dt = 2 \times 2 = 4
\]

Adding these, the total integral over one period is:
4 + 4 + 4 + 4 = 16.

Now, calculate \( a_0 \):

\[
a_0 = \frac{1}{20} \times 16 = 0.8
\]

{\color{red}Thus, the Fourier series coefficient \( a_0 \) is \( 0.8 \).}
\vspace{0.4em}
\end{minipage}
\hrule
\vspace{0.4em}
\caption{Examples of LMM laziness in EEE-Bench. We mark the \textit{spurious text} in the caption and error in response with {\color{red}red} color}
\label{lazy_3}
\end{figure}

\begin{figure}[th]
\hrule
\centering
\begin{minipage}[c]{0.45\textwidth}
\includegraphics[width=\textwidth]{       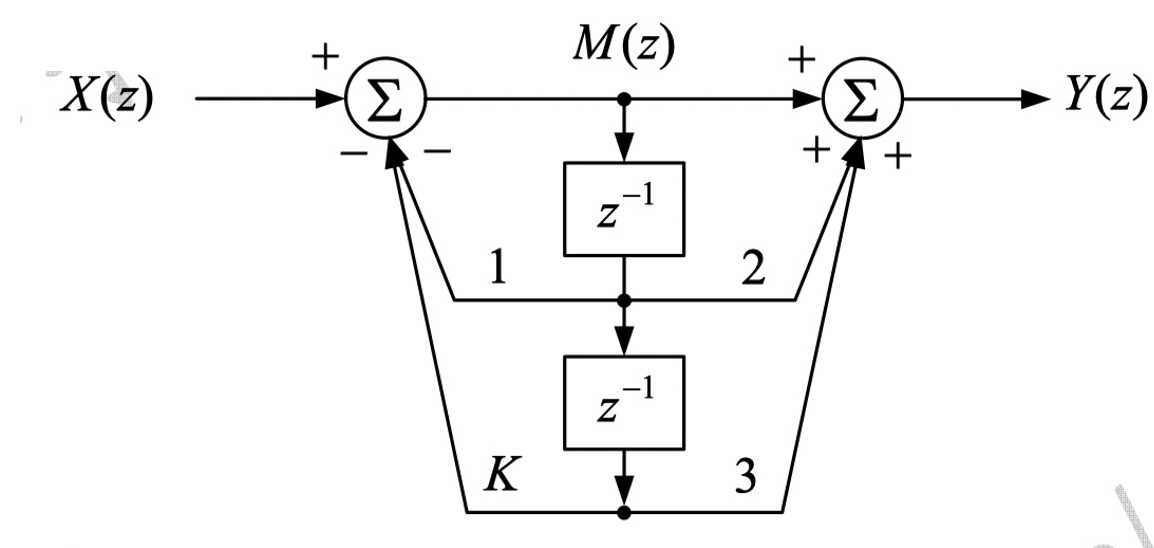}
\end{minipage}
\hfill
\begin{minipage}[c]{0.98\textwidth}
\small
\textbf{Spurious context:} Image caption context:

The diagram is a block diagram representing a {\color{red}continuous-time} signal processing system using the {\color{red}Laplace transform}. Here is a detailed description:

1. Inputs and Outputs:
   - The input to the system is denoted as \( X({\color{red}s}) \).
   - The output of the system is denoted as \( Y({\color{red}s}) \).

2. Summing Junctions:
   - There are {\color{red}three} summing junctions in the diagram.
   - The first summing junction has {\color{red}three inputs}: \( X({\color{red}s}) \) with a {\color{red}negative} sign and a feedback path with a {\color{red}positive} sign.
   - The second summing junction has {\color{red}two inputs}, all with {\color{red}negative} signs.

3. Paths and Connections:
   - The output of the first summing junction is connected to a path labeled \( {\color{red}N(s)} \).
   - This path leads to the second summing junction.

4. Delay Elements:
   - There is {\color{red}one} delay element in the diagram, represented by a block labeled \( {\color{red}s^{-1}} \).
   - The delay element is connected in {\color{red}parallel} with the output of the first summing junction.

5. Feedback and Feedforward Paths:
   - The output of the delay element is fed back to the first summing junction with a gain of \( K \).
   - The output of the delay element is connected to the second summing junction with a gain of {\color{red}3}.
   - The output of the second summing junction is also connected to the second summing junction with a gain of {\color{red}2}.

6. Gains:
   - The feedback path from the output of the delay element to the first summing junction has a gain of \( K \).
   - The feedforward path from the output of the delay element to the second summing junction has a gain of {\color{red}3}.
   - The feedforward path from the output of the {\color{red}second} summing junction to the {\color{red}second summing junction} has a gain of {\color{red}2}.

This block diagram represents a {\color{red}continuous-time} system with feedback and feedforward paths, incorporating a delay element and gains, which can be analyzed using {\color{red}Laplace transform} techniques to determine the system's behavior and response.
\\
\textbf{Question: }If the system is stable when \( a < K < b \), then \( a + b = \) ( ).
\\
\textbf{Response w/o spurious caption:}To determine the stability of the system, we need to find the characteristic equation of the closed-loop transfer function. The system has two delay elements, each represented by \( z^{-1} \).

The transfer function \( M(z) \) can be derived from the block diagram:

1. The forward path gain is \( 1 \).
2. The feedback path includes two delay elements and a gain \( K \).

The characteristic equation for the closed-loop system is:
\[ 1 + K z^{-2} = 0 \]
Solving for \( z \), we get:
\[ K z^{-2} = -1, z^2 = -K \]
For stability, the poles of the system must lie inside the unit circle in the z-plane. This means the magnitude of \( z \) must be less than 1:
$|z| < 1 $.
Substituting \( z^2 = -K \), we have:
\[ |-K| < 1,  K < 1 \]
Additionally, since \( z^2 = -K \), \( K \) must be positive for the system to have real poles: $K > 0$
Thus, the system is stable when \( 0 < K < 1 \). Therefore, \( a = 0 \) and \( b = 1 \), so:
\[ a + b = 0 + 1 = 1.0 \]
The final value is \( 1.0 \).
\\
\textbf{Response w/ spurious caption:} To determine the stability of the system, we need to analyze the characteristic equation derived from the block diagram. The system is stable if all poles of the transfer function have negative real parts.

Given the block diagram, the characteristic equation can be derived from the feedback loop. The transfer function \( H(z) \) can be expressed as:

\[ H(z) = \frac{Y(z)}{X(z)} = \frac{M(z)}{1 + M(z) \cdot G(z)} \]

where \( M(z) \) is the forward path transfer function and \( G(z) \) is the feedback path transfer function.

From the diagram:
- The forward path \( M(z) \) includes the two delay elements, so {\color{red}\( M(z) = \frac{1}{z^2} \)}.
- The feedback path \( G(z) \) includes the gain \( K \) and the delay elements, so {\color{red} \( G(z) = K \cdot \frac{1}{z^2} \)}.
The characteristic equation is:
{\color{red}\[ 1 + M(z) \cdot G(z) = 1 + \frac{1}{z^2} \cdot K \cdot \frac{1}{z^2} = 1 + \frac{K}{z^4} = 0 \]}
This simplifies to:
$ z^4 + K = 0 $.
For stability, the roots of this equation must lie inside the unit circle in the z-plane. The roots are:
\[ z = \sqrt[4]{-K} \]
For the system to be stable, the magnitude of \( z \) must be less than 1:
\( |\sqrt[4]{-K}| < 1 \)
This implies:
\( |K| < 1 \)
Thus, the system is stable when \( -1 < K < 1 \).
Therefore, \( a = -1 \) and \( b = 1 \), so \( a + b = 0 \).
The final value is \( 0.0 \).
\vspace{0.4em}
\end{minipage}
\hrule
\vspace{0.4em}
\caption{Examples of LMM laziness in EEE-Bench. We mark the \textit{spurious text} in the caption and error in response with {\color{red}red} color}
\label{lazy_4}
\end{figure}

\begin{figure}[th]
\hrule
\centering
\begin{minipage}[c]{0.25\textwidth}
\includegraphics[width=\textwidth]{       }
\end{minipage}
\hfill
\begin{minipage}[c]{0.98\textwidth}
\small
\textbf{Spurious context:} Image caption context:

The diagram represents a {\color{red}sinusoidal} function \( f(t) \) plotted against time \( t \). Here is a detailed description:

1. Axes:
   - The horizontal axis is labeled \( t \), representing time.
   - The vertical axis is labeled \( f(t) \), representing the function value.

2. Function Description:
   - The function \( f(t) \) is defined over the interval from \( t = -2 \) to \( t = 2 \).
   - The function is zero for \( t < -2 \) and \( t > 2 \).

3. Piecewise Segments:
   - **From \( t = -2 \) to \( t = -1 \)**:
     - The function {\color{red}linearly decreases from 1 to 0}.
   - **From \( t = -1 \) to \( t = 1 \)**:
     - The function remains constant at {\color{red}0}.
   - **From \( t = 1 \) to \( t = 2 \)**:
     - The function {\color{red}linearly increases from 0 to 1}.

4. Key Points:
   - At \( t = -2 \), \( f(t) = {\color{red}1} \).
   - At \( t = -1 \), \( f(t) = {\color{red}0} \).
   - At \( t = 0 \), \( f(t) = 0 \).
   - At \( t = 1 \), \( f(t) = {\color{red}0} \).
   - At \( t = 2 \), \( f(t) = {\color{red}1} \).

This function is a {\color{red}sinusoidal} waveform, often used in signal processing and control systems to represent a pulse with linear rise and fall times.

\textbf{Question:}
The signal \( f(t) \) is as shown in the figure, and the frequency spectrum function \( F(j\omega) \) is equal to:

A. \( \text{Sa}^2\left(\frac{\omega}{2}\right)[1 + 2\cos\omega] \)  
B. \( \text{Sa}^2\left(\frac{\omega}{2}\right)[1 + 2\sin\omega] \)  
C. \( \text{Sa}^2\left(\frac{\omega}{2}\right)[1 + \cos\omega] \)  
D. \( \text{Sa}^2\left(\frac{\omega}{2}\right)\left[1 + 2\cos\frac{\omega}{2}\right] \)

\textbf{Response w/o spurious caption:}
To find the frequency spectrum \( F(j\omega) \) of the given signal \( f(t) \), we need to analyze the shape of the signal. The signal is a trapezoidal pulse, which can be decomposed into a rectangular pulse and two triangular pulses.

1. **Rectangular Pulse**: The central part from \(-1\) to \(1\) is a rectangular pulse of width 2 and height 1. The Fourier transform of a rectangular pulse of width \( T \) and height \( A \) is \( A \cdot T \cdot \text{Sa}(\omega T/2) \).

2. **Triangular Pulses**: The two triangular parts on either side can be considered as the difference between two shifted rectangular pulses. The Fourier transform of a triangular pulse is the square of the sinc function.

The overall Fourier transform will be the sum of these components. The sinc function is defined as \( \text{Sa}(x) = \frac{\sin(\pi x)}{\pi x} \).

Given the symmetry and the shape of the signal, the Fourier transform will involve the square of the sinc function and a cosine term due to the symmetry about the origin.

The correct choice is:
A. \( \text{Sa}^2\left(\frac{\omega}{2}\right)[1 + 2\cos\omega] \)

\textbf{Response w/ spurious caption: }{\color{red}The function \( f(t) \) is a triangular pulse}, which can be represented as the convolution of two rectangular pulses. The Fourier transform of a triangular pulse is the square of the sinc function, often denoted as \(\text{Sa}^2\left(\frac{\omega}{2}\right)\).

Given the symmetry and the linear rise and fall, the frequency spectrum function \( F(j\omega) \) is:

{\color{red}\[ F(j\omega) = \text{Sa}^2\left(\frac{\omega}{2}\right)\left[1 + 2\cos\frac{\omega}{2}\right] \]}
Thus, the correct choice is:
{\color{red}**D.** \( \text{Sa}^2\left(\frac{\omega}{2}\right)\left[1 + 2\cos\frac{\omega}{2}\right] \)
\vspace{0.2em}}
\end{minipage}
\hrule
\begin{minipage}[c]{0.15\textwidth}
\includegraphics[width=\textwidth]{       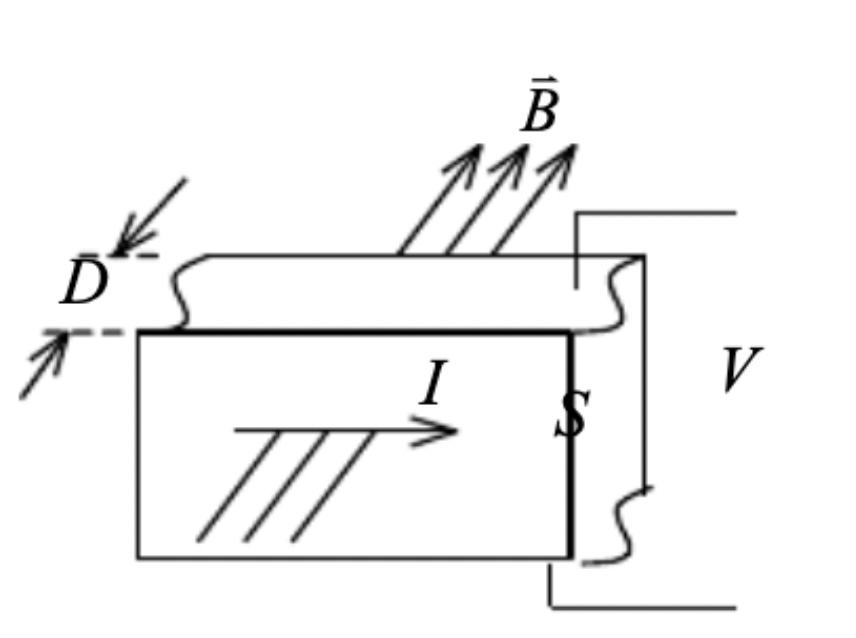}
\end{minipage}
\hfill
\begin{minipage}[c]{0.98\textwidth}
\small
\textbf{Spurious context: }Image caption context:

The diagram appears to represent a section of a {\color{red}simple electrical circuit involving basic resistive components}. Here is a detailed description of the components and parameters:

1. Rectangular Block: 
   - The main component is a rectangular block, which could represent a {\color{red}resistor or a non-conductive barrier}.
2. Magnetic Flux Density (\( \mathbf{B} \)):
   - Arrows pointing {\color{red}into the bottom} surface of the block indicate the direction of the magnetic flux density (\( \mathbf{B} \)), suggesting that the block is {\color{red}outside of any magnetic field influence}.
3. Electric Current (\( I \)):
   - An arrow labeled \( I \) is shown {\color{red}outside} the block, indicating the direction of electric current flow {\color{red}around the block, not through it}.
4. Electric Displacement Field (\( \mathbf{D} \)):
   - Arrows labeled \( \mathbf{D} \) are shown {\color{red}exiting the right} side of the block, indicating the direction of the electric displacement field. This suggests the {\color{red}absence of an electric field within} the block.
5. Voltage (\( V \)):
   - A curly bracket on the {\color{red}left} side of the block is labeled \( V \), indicating the potential difference or voltage {\color{red}along the side of the block}.
6. Surface (\( S \)):
   - The {\color{red}bottom} surface of the block is labeled \( S \), which could represent the area {\color{red}unrelated to the magnetic flux density}.

This diagram likely represents a scenario involving {\color{red}basic electrical principles, such as a simple resistor circuit}, where electric fields and currents are {\color{red}minimally interacting with magnetic fields}. The parameters \( \mathbf{B} \), \( \mathbf{D} \), \( I \), \( V \), and \( S \) are used to analyze the {\color{red}basic electrical behavior} of the system.

\textbf{Question: } 
A conductor carrying a current \( I \) has a thickness \( D \) and a cross-sectional area \( S \). It is placed in a uniform magnetic field with a magnetic flux density \( B \), where the magnetic field direction is perpendicular to the side surface of the conductor, as shown in the figure. The potential difference between the top and bottom surfaces of the conductor is measured to be \( V \). The Hall coefficient of this conductor is:

Choices:  
(A) \( \frac{V D S}{I B} \)  
(B) \( \frac{I B V}{D S} \)  
(C) \( \frac{V S}{I B D} \)  
(D) \( \frac{I V S}{B D} \)  
(E) \( \frac{V D}{I B} \)

\textbf{Response w/o spurious caption: }
The Hall coefficient \( R_H \) is given by the formula:
\(
R_H = \frac{V D}{I B}
\)
Thus, the correct option is:
(E) \( \frac{V D}{I B} \)

\textbf{Response w/ spurious caption: }
The Hall coefficient (\( R_H \)) is given by the formula:
{\color{red}\(
R_H = \frac{V}{I B D}
\)}
This formula relates the Hall voltage (\( V \)), current (\( I \)), magnetic flux density (\( B \)), and thickness (\( D \)) of the conductor.

Therefore, the correct choice is:
{\color{red}(C) \( \frac{V S}{I B D} \)}
\vspace{0.2em}
\end{minipage}
\hrule
\vspace{0.2em}
\caption{Examples of LMM laziness in EEE-Bench. We mark the \textit{spurious text} in the caption and error in response with {\color{red}red} color}
\label{lazy_5}
\end{figure}

\begin{figure}[th]
\hrule
\centering
\begin{minipage}[c]{0.23\textwidth}
\includegraphics[width=\textwidth]{       }
\end{minipage}
\hfill
\begin{minipage}[c]{0.98\textwidth}
\small
\textbf{Spurious context: }Image caption context:

The diagram depicts a transformer with a voltmeter connected to it. Here is a detailed description:

1. Transformer Core: 
   - The transformer has a {\color{red}circular core}, typically made of {\color{red}solid steel or another non-magnetic material}.

2. Primary Winding:
   - The primary winding is on the {\color{red}right} side of the core.
   - It is connected to a {\color{red}DC voltage source}, indicated by a {\color{red}plus and minus symbol}.
   - The number of turns in the primary winding is denoted as \( n_{\color{red}2} \).

3. Secondary Winding:
   - The secondary winding is on the {\color{red}left} side of the core.
   - It has {\color{red}two terminals labeled A and B}.
   - Terminal {\color{red}C} is a tap on the winding, indicating that the secondary winding is tapped to provide different {\color{red}current} levels.

4. Connections:
   - The primary winding is connected to a {\color{red}DC} source.
   - The secondary winding has {\color{red}two} output terminals (A, B) for different {\color{red}current} outputs.

5. Voltmeter:
   - A voltmeter is connected across the {\color{red}primary} winding.
   - The voltmeter is shown in a {\color{red}rectangular} symbol with the letter {\color{red}"A"} inside.

This setup is typical for a {\color{red}step-up transformer} with a tapped {\color{red}primary} winding, allowing for multiple {\color{red}current} outputs. The voltmeter measures the {\color{red}current} across the selected terminals of the {\color{red}primary} winding.
\\
\textbf{Question: } 
The transformer shown in the figure has an input voltage of 220 V and can output voltages of 12 V, 18 V, and 30 V. The primary coil with \( n_1 \) turns has a voltage varying with time as \( \mu = U_{\mathrm{m}} \cos (100 \pi t) \). A single-turn coil wound around the iron core is connected to an AC voltmeter, which shows a reading of 0.1 V. When a resistor \( R \) with a resistance of 12 \(\Omega\) is connected across points \( B \) and \( C \), the power is 12 W. Which of the following statements is correct?
  
A. \( n_1 \) is 1100 turns, and \( U_{\mathrm{m}} \) is 220 V  \\
B. The number of turns in the coil between \( B \) and \( C \) is 120, and the current through \( R \) is 1.4 A  \\
C. If \( R \) is connected across \( A \) and \( B \), the voltage across \( R \) is 18 V, with a frequency of 100 Hz \\ 
D. If \( R \) is connected across \( A \) and \( C \), the current through \( R \) is 2.5 A, with a period of 0.02 s

\textbf{Response w/o spurious caption: }
To solve this problem, we need to analyze the transformer and the given conditions.

1. **Single-turn coil reading**: The voltmeter connected to a single-turn coil reads 0.1 V. This means the induced EMF in one turn is 0.1 V.

2. **Power across \( R \) between \( B \) and \( C \)**: The power is given as 12 W, and the resistance \( R \) is 12 \(\Omega\). Using the power formula \( P = I^2 R \), we can find the current \( I \):
   \(
   12 = I^2 \times 12 \implies I = 1 \, \text{A}
   \).
   The voltage across \( R \) is \( V = I \times R = 1 \times 12 = 12 \, \text{V} \).

3. **Number of turns between \( B \) and \( C \)**: Since the voltage across \( B \) and \( C \) is 12 V and the voltage per turn is 0.1 V, the number of turns is:
   \(
   \frac{12}{0.1} = 120 \, \text{turns}
   \).
4. **Frequency and period**: The input voltage varies as \( \mu = U_{\mathrm{m}} \cos (100 \pi t) \), which indicates a frequency of \( f = \frac{100 \pi}{2\pi} = 50 \, \text{Hz} \). The period \( T \) is \( \frac{1}{f} = 0.02 \, \text{s} \).

Now, let's evaluate the options:

- **A**: \( n_1 \) is 1100 turns, and \( U_{\mathrm{m}} \) is 220 V.  
  This option does not provide enough information to verify.

- **B**: The number of turns in the coil between \( B \) and \( C \) is 120, and the current through \( R \) is 1.4 A.  
  The number of turns is correct, but the current is 1 A, not 1.4 A.

- **C**: If \( R \) is connected across \( A \) and \( B \), the voltage across \( R \) is 18 V, with a frequency of 100 Hz.  
  The frequency is 50 Hz, not 100 Hz.

- **D**: If \( R \) is connected across \( A \) and \( C \), the current through \( R \) is 2.5 A, with a period of 0.02 s.  
  The period is correct, but we need to verify the current.

Since none of the options are entirely correct, let's verify the current for option D:

- Voltage across \( A \) and \( C \) is 30 V (given).
- Current \( I = \frac{V}{R} = \frac{30}{12} = 2.5 \, \text{A} \).

Thus, option D is correct regarding the current and period.

**Correct Answer: D**

\textbf{Response w/ spurious caption: }
To solve this problem, we need to analyze the given information and apply transformer principles.

1. **Transformer Basics**:
   {\color{red}- The voltage ratio in a transformer is given by \( \frac{V_1}{V_2} = \frac{n_1}{n_2} \).
   - Power in a resistor is given by \( P = I^2 R \).}

2. **Given Information**:
   - Input voltage \( V_1 = 220 \, \text{V} \).
   - Output voltages: 12 V, 18 V, 30 V.
   - Voltmeter reading: 0.1 V.
   - Resistor \( R = 12 \, \Omega \).
   - Power across \( B \) and \( C \) is 12 W.

3. **Analysis**:
   - **Power Calculation**: 
     \[
     P = I^2 R \implies 12 = I^2 \times 12 \implies I = 1 \, \text{A}
     \]
   - **Voltage across \( B \) and \( C \)**:
     \[
     V = I \times R = 1 \times 12 = 12 \, \text{V}
     \]
   - **Turns Ratio**:
     - For 12 V output, the turns ratio \( \frac{n_1}{n_2} = \frac{220}{12} \).

{\color{red}4. **Checking Options**:
   - **A**: \( n_1 \) is 1100 turns, and \( U_{\mathrm{m}} \) is 220 V.
     - This option does not provide enough information to verify without additional calculations.
   - **B**: The number of turns in the coil between \( B \) and \( C \) is 120, and the current through \( R \) is 1.4 A.
     - Current through \( R \) is 1 A, not 1.4 A.
   - **C**: If \( R \) is connected across \( A \) and \( B \), the voltage across \( R \) is 18 V, with a frequency of 100 Hz.
     - The frequency is correct (100 Hz), but the voltage needs verification.
   - **D**: If \( R \) is connected across \( A \) and \( C \), the current through \( R \) is 2.5 A, with a period of 0.02 s.
     - The period of 0.02 s corresponds to 50 Hz, not 100 Hz.
**Correct Option**: C}
\vspace{0.4em}
\end{minipage}
\hrule
\vspace{0.1em}
\caption{Examples of LMM laziness in EEE-Bench. We mark the \textit{spurious text} in the caption and error in response with {\color{red}red} color}
\label{lazy_6}
\end{figure}

\begin{figure}[th]
\hrule
\centering
\begin{minipage}[c]{0.35\textwidth}
\includegraphics[width=\textwidth]{       }
\end{minipage}
\hfill
\begin{minipage}[c]{0.98\textwidth}
\small
\textbf{Spurious context: } Image caption context:

The diagram represents an electrical circuit with the following components and connections:

1. Ammeter (A):
   - Positioned at the {\color{red}end} of the circuit.
   - Measures the {\color{red}voltage} across the circuit.

2. Transformer:
   - Consists of two coils (primary and secondary) {\color{red}without a core symbol between them}.
   - The primary coil is connected {\color{red}in parallel} with the ammeter.
   - The secondary coil is connected to the rest of the circuit.

3. Voltmeter (V):
   - Connected {\color{red}in series} with the secondary coil of the transformer.
   - Measures the {\color{red}current} through the secondary coil.

4. Resistor \( R_1 \):
   - Connected {\color{red}in series} with the voltmeter.
   - Part of the load on the {\color{red}primary} side of the transformer.

5. Switch (S):
   - Connected {\color{red}in parallel} with resistor \( R_2 \).
   - Controls the {\color{red}disconnection} of \( R_2 \) from the circuit.

6. Resistor \( R_2 \):
   - Connected {\color{red}in parallel} with the switch.
   - When the switch is closed, \( R_2 \) is {\color{red}removed} from the circuit.

**Connections**:
- The primary side of the transformer is connected {\color{red}in parallel} with the ammeter.
- The secondary side of the transformer is connected {\color{red}in series} to the voltmeter and resistor \( R_1 \).
- The switch \( S \) and resistor \( R_2 \) are connected {\color{red}in parallel}, forming a {\color{red}series} branch with \( R_1 \) when the switch is closed.

This setup can be used to measure the {\color{red}voltage and current} in a transformer circuit, and to analyze the effects of {\color{red}removing or adding a series load} (resistor \( R_2 \)) on the circuit.

\textbf{Question: } In the circuit shown in figure, an ideal transformer has its primary coil connected to a high voltage and its secondary coil connected to a load, with the line resistance being negligible. When \( S \) is initially closed and \( R_{1} = R_{2} \), and then \( S \) is opened, the AC voltage reading \( U \), AC current reading \( I \), and power \( P_{1} \) on the load \( R_{1} \) will 

A. \( U \) increases, \( I \) increases, \( P_{1} \) increases 

B. \( U \) remains unchanged, \( I \) decreases, \( P_{1} \) increases  

C. \( U \) decreases, \( I \) decreases, \( P_{1} \) decreases  

D. \( U \) remains unchanged, \( I \) decreases, \( P_{1} \) remains unchanged

\textbf{Response w/o spurious caption: }When the switch \( S \) is initially closed, the total resistance in the secondary circuit is \( R_1 + R_2 \). When \( S \) is opened, the resistance becomes just \( R_1 \).

Since \( R_1 = R_2 \), opening the switch doubles the resistance in the secondary circuit. 

1. **Voltage \( U \):** The voltage across the load \( R_1 \) remains unchanged because the transformer maintains the same voltage ratio regardless of the load.

2. **Current \( I \):** The current decreases because the total resistance in the circuit increases when \( S \) is opened.

3. **Power \( P_1 \):** The power \( P_1 = U^2 / R_1 \) remains unchanged because the voltage across \( R_1 \) remains the same and \( R_1 \) itself does not change.

Therefore, the correct answer is:

D. \( U \) remains unchanged, \( I \) decreases, \( P_{1} \) remains unchanged

\textbf{Response w/ spurious caption: }
In the given circuit, when the switch \( S \) is initially closed, resistors \( R_1 \) and \( R_2 \) are in parallel, effectively reducing the total resistance on the secondary side. When \( S \) is opened, \( R_2 \) is removed, leaving only \( R_1 \) in the circuit, which increases the total resistance.

1. {\color{red}**Voltage \( U \):** The voltage across the secondary coil remains unchanged because the transformer maintains a constant voltage ratio between the primary and secondary coils.}

2. **Current \( I \):** With the increase in resistance (since \( R_2 \) is removed), the current through the circuit decreases.

3. {\color{red}**Power \( P_1 \):** Power is given by \( P = I^2R \). Since the current decreases and the resistance remains constant at \( R_1 \), the power \( P_1 \) decreases.}

Therefore, the correct choice is:

{\color{red}C. \( U \) decreases, \( I \) decreases, \( P_{1} \) decreases}

\vspace{0.4em}
\end{minipage}
\hrule
\vspace{0.1em}
\caption{Examples of LMM laziness in EEE-Bench. We mark the \textit{spurious text} in the caption and error in response with {\color{red}red} color}
\label{lazy_7}
\end{figure}

\begin{figure}[th]
\hrule
\centering
\begin{minipage}[c]{0.4\textwidth}
\includegraphics[width=\textwidth]{       }
\end{minipage}
\hfill
\begin{minipage}[c]{0.98\textwidth}
\small
\textbf{Spurious context: }Image caption context:

The diagram depicts a {\color{red}sinusoidal} waveform centered on the time axis, which is commonly used in signal processing and control systems. Here is a detailed description:

1. Axes:
   - The horizontal axis represents time, labeled as \( t \).
   - The vertical axis represents the amplitude of the signal, labeled as \( x(t) \).

2. Waveform Shape:
   - The waveform is a symmetric {\color{red}sine wave} centered at the origin.

3. Key Points:
   - The {\color{red}sine wave} peaks at \( t = 0 \) with an amplitude of 1.
   - The {\color{red}base of the sine wave} extends from \( t = -1 \) to \( t = 1 \).

4. Slopes:
   - From \( t = -1 \) to \( t = 0 \), the waveform {\color{red}smoothly} increases from 0 to 1.
   - From \( t = 0 \) to \( t = 1 \), the waveform {\color{red}smoothly} decreases from 1 to 0.

5. Mathematical Representation:
   - For \( -1 \leq t < 0 \), \( x(t) = {\color{red}\sin(\pi t + \pi)} \).
   - For \( 0 \leq t \leq 1 \), \( x(t) = {\color{red}\sin(\pi t)} \).

This {\color{red}sinusoidal} waveform is often used in applications involving signal modulation, filtering, and analysis in electrical and electronics engineering.

\textbf{Question: } Given that the Fourier transform of \( x(t) \) is \( X(j \omega) \), and the waveform of \( x(t) \) is as shown in the figure, determine \( X(0) = \) ( ).

\textbf{Response w/o spurious caption: }
To find \( X(0) \), which is the Fourier transform of \( x(t) \) evaluated at \( \omega = 0 \), we need to calculate the area under the curve of \( x(t) \). This is because \( X(0) \) represents the integral of \( x(t) \) over all time:

\[
X(0) = \int_{-\infty}^{\infty} x(t) \, dt
\]

The waveform shown is a triangle with a base from \(-1\) to \(1\) and a height of \(1\). The area \( A \) of a triangle is given by:

\[
A = \frac{1}{2} \times \text{base} \times \text{height}
\]

For this triangle:

- Base = \(1 - (-1) = 2\)
- Height = \(1\)

Thus, the area is:

\[
A = \frac{1}{2} \times 2 \times 1 = 1
\]

Therefore, \( X(0) = 1.0 \).

\textbf{Response w/ spurious caption: }
To find \( X(0) \), which is the Fourier transform of \( x(t) \) evaluated at zero frequency, we need to calculate the area under the curve of \( x(t) \) over its period. This is equivalent to finding the integral of \( x(t) \) over its domain.

Given the piecewise function:
- {\color{red}For \( -1 \leq t < 0 \), \( x(t) = \sin(\pi t + \pi) = -\sin(\pi t) \).
- For \( 0 \leq t \leq 1 \), \( x(t) = \sin(\pi t) \).}

The integral of \( x(t) \) from \(-1\) to \(1\) is:

\[
X(0) = \int_{-1}^{0} -\sin(\pi t) \, dt + \int_{0}^{1} \sin(\pi t) \, dt
\]

Calculating each integral:

1. \(\int_{-1}^{0} -\sin(\pi t) \, dt = \left[ \frac{\cos(\pi t)}{\pi} \right]_{-1}^{0} = \frac{\cos(0) - \cos(-\pi)}{\pi} = \frac{1 - (-1)}{\pi} = \frac{2}{\pi}\)

2. \(\int_{0}^{1} \sin(\pi t) \, dt = \left[ -\frac{\cos(\pi t)}{\pi} \right]_{0}^{1} = \frac{-\cos(\pi) + \cos(0)}{\pi} = \frac{-(-1) + 1}{\pi} = \frac{2}{\pi}\)

Adding these results:

\[
X(0) = \frac{2}{\pi} + \frac{2}{\pi} = \frac{4}{\pi}
\]

{\color{red}Thus, the value of \( X(0) \) is approximately \( 1.3 \).}
\vspace{0.4em}
\end{minipage}
\hrule
\caption{Examples of LMM laziness in EEE-Bench. We mark the \textit{spurious text} in the caption and error in response with {\color{red}red} color}
\label{lazy_8}
\end{figure}

\begin{figure}[th]
\hrule
\centering
\begin{minipage}[c]{0.4\textwidth}
\includegraphics[width=\textwidth]{       }
\end{minipage}
\hfill
\begin{minipage}[c]{0.98\textwidth}
\small
\textbf{Spurious context: }Image caption context:

The diagram is a table listing different types of beads along with their prices per kilogram. The table has two columns: the first column lists the bead shapes, and the second column lists the corresponding prices. The table is organized as follows:

1. Oval beads - priced at {\color{red}\$3} per kilogram.
2. Rectangular beads - priced at {\color{red}\$2} per kilogram.
3. Star-shaped beads - priced at {\color{red}\$3} per kilogram.
4. Spherical beads - priced at {\color{red}\$3} per kilogram.
5. Heart-shaped beads - priced at {\color{red}\$2} per kilogram.
6. Square beads - priced at {\color{red}\$3} per kilogram.
7. Flower-shaped beads - priced at {\color{red}\$3} per kilogram.

The table is presented with a {\color{red}light blue} background, and each row alternates in a slightly darker shade of {\color{red}blue} for clarity. This table can be used for calculations involving the cost of purchasing different types of beads by weight.

\textbf{Question: } Chase wants to buy 4 kilograms of oval beads and 5 kilograms of star-shaped beads. How much will he spend? (Unit: \$)

\textbf{Response w/o spurious caption: }
Chase wants to buy 4 kilograms of oval beads and 5 kilograms of star-shaped beads.

- Cost of oval beads: 4 kg × \$2/kg = \$8
- Cost of star-shaped beads: 5 kg × \$2/kg = \$10

Total cost: \$8 + \$10 = \$18

18

\textbf{Response w/ spurious caption: }
To calculate the total cost:

1. **Oval beads**: 4 kg at {\color{red} \$3 per kg = 4 * 3 =  \$12}
2. **Star-shaped beads**: 5 kg at  {\color{red}\$3 per kg = 5 * 3 =  \$15}

{\color{red}Total cost =  \$12 +  \$15 = \$27

27}
\vspace{0.4em}
\end{minipage}
\hrule
\begin{minipage}[c]{0.4\textwidth}
\includegraphics[width=\textwidth]{       }
\end{minipage}
\hfill
\begin{minipage}[c]{0.98\textwidth}
\small
\textbf{Spurious context: }Image caption context:

The diagram is a stem-and-leaf plot titled "Houses per block." It is used to display quantitative data in a compact form, showing the distribution of values.

- Title: Houses per block

- Structure: The plot is divided into two columns: "Stem" and "Leaf."

- Stem Column: 
  - The stem represents the {\color{red}hundreds} digit of the data values.
  - Values in the stem column are: 0, 1, 2, 3.

- Leaf Column:
  - The leaf represents the {\color{red}tens} digit of the data values.
  - For each stem value, the corresponding leaf values are listed in {\color{red}descending} order.

- Data:
  - Stem 0: Leaves are {\color{red}7, 5}. This represents the numbers {\color{red}70 and 50}.
  - Stem 1: Leaves are {\color{red}5, 4, 0}. This represents the numbers {\color{red}150, 140, and 100}.
  - Stem 2: Leaves are {\color{red}9, 8, 7, 6, 4, 3, 3, 2, 1, 0, 0}. This represents the numbers {\color{red}290, 280, 270, 260, 240, 230, 230, 220, 210, 200, and 200}.
  - Stem 3: Leaves are {\color{red}8, 6, 4, 2, 1, 0}. This represents the numbers {\color{red}380, 360, 340, 320, 310, and 300}.

This stem-and-leaf plot provides a clear visual representation of the distribution of the number of houses per block, allowing for easy identification of patterns, such as clustering and spread of the data.

\textbf{Question: } A real estate agent drove around the neighborhood and counted the number of houses on each block. How many blocks have exactly 36 houses? (Unit: blocks)

\textbf{Response w/o spurious caption: }
To find the number of blocks with exactly 36 houses, look at the stem "3" and the leaf "6" in the stem-and-leaf plot.

There is 1 block with exactly 36 houses.

Final value: 1

\textbf{Response w/ spurious caption: }
{\color{red}The stem-and-leaf plot does not show any blocks with exactly 36 houses. The closest values in the plot are 30 and 40, but not 36.

Therefore, the number of blocks with exactly 36 houses is 0.}
\vspace{0.4em}
\end{minipage}
\hrule
\caption{Examples of LMM laziness in MathVista. We mark the \textit{spurious text} in the caption and error in response with {\color{red}red} color}
\label{lazy_9}
\end{figure}

\begin{figure}[th]
\hrule
\centering
\begin{minipage}[c]{0.4\textwidth}
\includegraphics[width=\textwidth]{       }
\end{minipage}
\hfill
\begin{minipage}[c]{0.98\textwidth}
\small
\textbf{Spurious context: }Image caption context:

The diagram is a bar chart titled "Accuracy of different algorithms." It displays the accuracy of two algorithms, labeled "icy" and "irony," on the x-axis. The y-axis is labeled "Accuracy" and ranges from 0 to {\color{red}100}, with grid lines at intervals of {\color{red}20} units.

- The bar for "icy" is light blue and reaches an accuracy level of {\color{red}20}.
- The bar for "irony" is also light blue and reaches an accuracy level of {\color{red}80}.

The chart is designed to compare the performance of the two algorithms based on their accuracy.

\textbf{Question: } How many algorithms have accuracies higher than 9?

\textbf{Response w/o spurious caption: }
The number of algorithms with accuracies higher than 9 is 0.

\textbf{Response w/ spurious caption: }
{\color{red}2}
\vspace{0.4em}
\end{minipage}
\hrule
\begin{minipage}[c]{0.3\textwidth}
\includegraphics[width=\textwidth]{       }
\end{minipage}
\hfill
\begin{minipage}[c]{0.98\textwidth}
\small
\textbf{Spurious context: }Image caption context:

The diagram is a graph of a mathematical function plotted on a Cartesian coordinate system. Here is a detailed description:

1. Axes:
   - The graph has a horizontal x-axis and a vertical y-axis, both intersecting at the origin (0,0).
   - The axes are labeled with integer values at regular intervals: {\color{red}-3}, 0, and {\color{red}3} on both axes.

2. Grid:
   - The background consists of a grid with squares, each representing {\color{red}two} units on both the x and y axes.

3. Function:
   - The graph depicts an S-shaped curve, characteristic of a hyperbolic {\color{red}sine} function or a logistic function.
   - The curve passes through the origin (0,0), indicating it is symmetric about this point.

4. Behavior:
   - As x approaches negative infinity, the curve approaches y = {\color{red}-3}, suggesting a horizontal asymptote.
   - As x approaches positive infinity, the curve approaches y = {\color{red}3}, indicating another horizontal asymptote.
   - The curve is steepest around the origin, where it transitions from negative to positive values.

5. Quadrants:
   - The curve starts in the {\color{red}second} quadrant, passes through the origin, and continues into the {\color{red}fourth} quadrant.

This graph is likely representing a function with a sigmoid shape, commonly used in statistics, neural networks, and logistic growth models.

\textbf{Question: } What is the degree of this function?

\textbf{Response w/o spurious caption: }
The graph shown is that of a cubic function, which typically has the form \( f(x) = ax^3 + bx^2 + cx + d \). The characteristic S-shape indicates that the degree of the function is 3.

3

\textbf{Response w/ spurious caption: }
{\color{red}The graph represents a sigmoid function}, which is not a polynomial function. Therefore, it does not have a degree in the traditional sense used for polynomials. {\color{red}The degree of this function is 0.}
\vspace{0.4em}
\end{minipage}
\hrule
\vspace{0.2em}
\caption{Examples of LMM laziness in MathVista. We mark the \textit{spurious text} in the caption and error in response with {\color{red}red} color}
\label{lazy_10}
\end{figure}

\begin{figure}[th]
\hrule
\centering
\begin{minipage}{0.2\textwidth}
    \centering
    \includegraphics[width=\textwidth]{   } 
\end{minipage}
\hfill
\begin{minipage}{0.7\textwidth}
\small
\textbf{Question: } In the circuit shown in figure, the corresponding logical expression is ( ).
\\
A. \( F = A + B + C \)\\
B. \( F = A \cdot B \cdot C \)\\
C. \( F = \overline{A + B + C} \)\\
D. \( F = \overline{A \cdot B \cdot C } \)
\\
\textbf{Answer: } D
\end{minipage}
\begin{minipage}[c]{0.98\textwidth}
\textbf{GPT4-o: } The truth table shows that the output \( F \) is 1 for all combinations of \( A, B, C \) except when all are 1. This corresponds to the logical expression for a NOR gate, which is the negation of the AND operation.

The correct logical expression is:
\[ F = \overline{A \cdot B \cdot C} \]

So, the correct option is:
D. \( F = \overline{A \cdot B \cdot C} \)\\

\textbf{Gemini-Pro 1.5: }The truth table shows that F is 0 only when A, B, and C are all 1. This corresponds to a NAND gate. Therefore, the correct logical expression is:

**D.  \( F = \overline{A \cdot B \cdot C } \)** 
\\
\textbf{Qwen-VL-Max:} {\color{red}The logical expression for the circuit shown in the image is \(F=A+B+C\), which represents the sum of three inputs A, B, and C. Therefore, the correct choice is:

A. \( F = A + B + C \)}
\\
\textbf{InternVL2-26B: }{\color{red}The given circuit is a combination of AND gates. The output F is 1 only when all the inputs A, B, and C are 1.} Therefore, the logical expression for F is:
{\color{red}
\[ F = A \cdot B \cdot C \]

So, the correct option is B.}
\\
\textbf{MiniCPM-V-2-6: } Answer: D\\
\\
\textbf{LLaVA-v1.6-7b: }{\color{red}The correct option letter is C. The logical expression for the circuit shown in the figure is the complement of the sum of A, B, and C.}

\vspace{0.4em}
\end{minipage}
\hrule

\begin{minipage}{0.25\textwidth}
    \centering
    \includegraphics[width=\textwidth]{   } 
\end{minipage}
\hfill
\begin{minipage}{0.7\textwidth}
\small
\textbf{Question: } The logical circuit shown in figure, when A = "0", after the arrival of C pulse, the D flip-flop ( )\\
A. Has counter function
B. Set to "0"\\
C. Set to "1"
D. Uncertain
Answer is B
\\
\textbf{Answer: } B
\end{minipage}
\begin{minipage}[c]{0.98\textwidth}
\textbf{GPT4-o: } When A = "0", the output of the AND gate will be "0" regardless of the other input. Therefore, the D input of the flip-flop will be "0". After the arrival of the C pulse (clock pulse), the D flip-flop will be set to "0".

Answer: B
\\

\textbf{Gemini-Pro 1.5: }When A = "0", the output of the AND gate will be "0" regardless of the other input. Therefore, the D input of the flip-flop will be "0". After the arrival of the C pulse (clock pulse), the D flip-flop will be set to "0".

Answer: B
\\
\textbf{Qwen-VL-Max:} When A = "0", the AND gate output will be "0" regardless of the value of B. When the C pulse arrives, it will pass through the OR gate and trigger the D flip-flop. Since the input to the D flip-flop is "0" due to the AND gate output, the D flip-flop will set to "0". Therefore, the correct answer is:

B. Set to "0"
\\
\textbf{InternVL2-26B: }B. Set to "0"
\\
\textbf{MiniCPM-V-2-6: } {\color{red}Answer: C}\\
\\
\textbf{LLaVA-v1.6-7b: }The correct answer is B. The D flip-flop is set to "0" when A = "0" and after the arrival of C pulse.
\vspace{0.4em}
\end{minipage}
\hrule
\vspace{0.1em}
\caption{Examples of responses among 6 baselines in EEE-Bench. The error part os each response is marked as {\color{red} red}.}
\label{all_comparison_1}
\end{figure}

\begin{figure}[th]
\hrule
\centering
\begin{minipage}{0.2\textwidth}
    \centering
    \includegraphics[width=\textwidth]{   } 
\end{minipage}
\hfill
\begin{minipage}{0.7\textwidth}
\small
\textbf{Question: }In the circuit shown in figure, it is known that \( u_{i1} = 2V \), \( u_{i2} = -2V \). The output voltage \( u_o \) is ( )V.
\\
\textbf{Answer: } -8.
\end{minipage}
\begin{minipage}[c]{0.98\textwidth}
\textbf{GPT4-o: } To solve this problem, we need to analyze the circuit, which consists of two operational amplifiers (op-amps) in an inverting configuration.

 First Op-Amp:
- The first op-amp is in an inverting configuration.
- The input voltage \( u_{i1} = 2V \).
-{\color{red} The resistors \( R_1 \) and \( R \) form the feedback network.

The voltage at the output of the first op-amp, \( u_{o1} \), is given by the inverting amplifier formula:
\[
u_{o1} = -\left(\frac{R}{R_1}\right) u_{i1}
\]
Assuming \( R = R_1 \), we have:
\(
u_{o1} = -2V
\)}

Second Op-Amp:
- The second op-amp is also in an inverting configuration.
- The input voltage \( u_{i2} = -2V \).
- The resistors \( R_2 \) and \( R \) form the feedback network.

The voltage at the output of the second op-amp, \( u_o \), is given by:
{\color{red}\[
u_o = -\left(\frac{R}{R_2}\right) (u_{o1} + u_{i2})
\]
Assuming \( R = R_2 \), we have:
\[
u_o = -(u_{o1} + u_{i2}) = -(-2V - 2V) = 4V
\]
Therefore, the output voltage \( u_o \) is \( 4.0 \) volts.}
\\
\textbf{Gemini-Pro 1.5: }The circuit shown is a two-stage inverting amplifier. The gain of each stage is determined by the ratio of the feedback resistor to the input resistor. Since the open-loop gain of the op-amps is infinite, the voltage at the inverting input of each op-amp is equal to the voltage at the non-inverting input, which is 0V.

{\color{red}For the first stage, the output voltage \(u_{o1}\) can be calculated as:

\(u_{o1} = - \frac{R}{R}u_{i1} = -u_{i1} = -2V\)

For the second stage, the output voltage \(u_o\) can be calculated as:

\(u_o = - \frac{R}{R}u_{o1} = -u_{o1} = 2V\)

Therefore, the output voltage \(u_o\) is **2.0**. }
\\
\textbf{Qwen-VL-Max:} The circuit shown in the image is a simple voltage divider circuit. To find the output voltage \(u_o\), we need to calculate the voltage across the resistor \(R_2\) using Ohm's law.

Ohm's law states that \( V = IR \), where \(V\) is the voltage, \(I\) is the current, and \(R\) is the resistance.

{\color{red}In this case, the voltage across \(R_2\) can be calculated as follows:
\(V_{R2} = I * R_2\)
First, let's find the total resistance of the circuit (\(R_T\)):
\(R_T = R + R_1 + R_2\)
Then, we can find the current through the circuit (\(I\)) by dividing the input voltage (\(u_{i1}\)) by the total resistance:
\(I = \frac{u_{i1}}{R_T}\)
Finally, we can calculate the voltage across \(R_2\) (\(V_{R2}\)):

\(V_{R2} = I * R_2\)

Since \(u_i1 = 2V\) and \(u_i2 = -2V\), the net input voltage is zero. Therefore, the output voltage \(u_o\) will also be zero.

So, the output voltage \(u_o\) is 0.0 V.}
\\
\textbf{InternVL2-26B: }To determine the output voltage \( u_o \) in the given circuit, we need to analyze the circuit and apply the appropriate rules for resistors and operational amplifiers.

1. **Identify the Operational Amplifiers (Op-Amps):**
   - The circuit has two operational amplifiers, each with an input resistance of \( R \) and a feedback resistance of \( R_1 \) and \( R_2 \) respectively.

2. {\color{red}**Determine the Input Voltage for Each Op-Amp:**
   - For the left Op-Amp:
     - \( u_{i1} = 2V \)
     - \( u_{i2} = -2V \)
   - For the right Op-Amp:
     - \( u_{i1} = u_o \)
     - \( u_{i2} = u_o \)

3. **Apply the Op-Amp Voltage Rule:**
   - The voltage difference between the inverting and non-inverting inputs of each Op-Amp must be zero for proper operation.
   - For the left Op-Amp:
     \[
     u_{i1} - u_{i2} = 2V - (-2V) = 4V
     \]
     This indicates that the left Op-Amp is not functioning correctly, as the voltage difference is not zero.

4. **Analyze the Circuit Configuration:**
   - The circuit configuration suggests that the resistors \( R \) and \( R_1 \) are connected in series with the left Op-Amp, and \( R \) and \( R_2 \) are connected in series with the right Op-Amp.
   - The output voltage \( u_o \) is connected to both Op-Amps, indicating that the circuit is likely designed to sum the input voltages.

5. **Summing the Input Voltages:**
   - Since the circuit is designed to sum the input voltages, we can write the equation for \( u_o \):
     \(
     u_o = u_{i1} + u_{i2}
     \)
   - Substituting the given values:
     \[
     u_o = 2V + (-2V) = 0V
     \]

Therefore, the output voltage \( u_o \) is \( 0V \).}

\vspace{0.4em}
\end{minipage}
\hrule

\vspace{0.3em}
\caption{Examples of responses among 4 baselines in EEE-Bench, where all baselines fail to reason correctly for this problem.}
\label{all_comparison_2}
\end{figure}
\end{document}